%% file: main.tex
\documentclass[journal]{IEEEtran}

\usepackage{graphicx}
\usepackage{array}
\usepackage{makecell}
\usepackage{multirow}
\usepackage{amsmath,amsfonts}
\usepackage[normalem]{ulem}
\usepackage{tcolorbox}
\usepackage{colortbl}
\usepackage{enumitem}
\usepackage{url}
\usepackage{enumerate}
\usepackage{booktabs}
\usepackage{xspace}

\usepackage{color}
\usepackage{threeparttable}
\usepackage{ulem}
\usepackage{amssymb}

\usepackage{float}
\usepackage[ruled,linesnumbered]{algorithm2e}
\usepackage{balance}
\usepackage[caption=false,font=normalsize,labelfont=sf,textfont=sf]{subfig}
\usepackage{textcomp}
\usepackage{stfloats}
\usepackage{verbatim}
\usepackage{balance}
\usepackage{hyperref}
\hypersetup{
    colorlinks=true,        
    linkcolor=blue,         
    citecolor=green,        
    urlcolor=cyan,          
    pdfborder={0 0 0},      
}

\usepackage{algorithmic}

\newcommand{\myformer}{TimeControl }

\newcommand{\Rmnum}[1]{\expandafter\@slowromancap\romannumeral #1@}

\def\BibTeX{{\rm B\kern-.05em{\sc i\kern-.025em b}\kern-.08em
    T\kern-.1667em\lower.7ex\hbox{E}\kern-.125emX}}

\definecolor{green_bg}{RGB}{180, 215, 183}
\definecolor{green_hyp}{RGB}{160, 196, 143}

\begin{document}

\title{Domain Fusion Controllable Generalization for Cross-Domain Time Series Forecasting from Multi-Domain Integrated Distribution}

\author{
Xiangkai Ma, Xiaobin Hong, Mingkai Lin, Han Zhang, Wenzhong Li, Sanglu Lu
\vspace{-20pt}
\thanks{
Xiangkai Ma, Xiaobin Hong, Mingkai Lin, Han Zhang, Wenzhong Li, Sanglu Lu are with the State Key Laboratory for Novel Software Technology, Nanjing University, Nanjing 210023, China.
(e-mail: \{xiangkai.ma, xiaobinhong, zhanh\}@smail.nju.edu.cn, \{mingkai, lwz, sanglu\}@nju.edu.cn)
}
\thanks{Corresponding Authour: Wenzhong Li (lwz@nju.edu.cn).}
}

\markboth{IEEE TRANSACTIONS ON KNOWLEDGE AND DATA ENGINEERING}
{}

\maketitle
\begin{abstract}
Conventional deep models have achieved unprecedented success in time series forecasting. However, facing the challenge of cross-domain generalization, existing studies utilize statistical prior as prompt engineering fails under the huge distribution shift among various domains. 
In this paper, a novel time series generalization diffusion model (TimeControl) that pioneers the Domain-Fusion paradigm, systematically integrating information from multiple time series domains into a unified generative process via diffusion models. 
Unlike the autoregressive models that capture the conditional probabilities of the prediction horizon to the historical sequence, we use the diffusion denoising process to model the mixed distribution of the cross-domain data and generate the prediction sequence for the target domain directly utilizing conditional sampling. 
The proposed TimeControl contains three pivotal designs: 
\begin{small}$(1)$\end{small} The condition network captures the multi-scale fluctuation patterns from the observation sequence, which are utilized as context representations to guide the denoising network to generate the prediction sequence; 
\begin{small}$(2)$\end{small} Adapter-based fine-tuning strategy, the multi-domain universal representation learned in the pretraining stage is utilized for downstream tasks in target domains; 
\begin{small}$(3)$\end{small} A novel hybrid architecture is designed to align the observation and prediction spaces, enabling TimeControl to generate prediction sequences of arbitrary lengths with flexibility. 
We conduct extensive experiments on mainstream 49 benchmarks and 30 baselines, and the TimeControl outperforms existing baselines on all data domains, exhibiting superior zero-shot generalization ability.
\end{abstract}
\begin{IEEEkeywords}
Domain Fusion, Time Series Forecasting, Cross-domain Generalization, Diffusion Model
\end{IEEEkeywords}

\input{segments/2-Introduction.tex}
\input{segments/3-RelatedWork.tex}

\input{segments/4-Methodology.tex}

\input{segments/5-Experiment.tex}
\input{segments/6-Conclusion.tex}

\bibliographystyle{IEEEtrans}
\normalem
\bibliography{references_master}

\vspace{-1.0cm}
\begin{IEEEbiography}[{\includegraphics[width=1in,height=1.25in,clip,keepaspectratio]{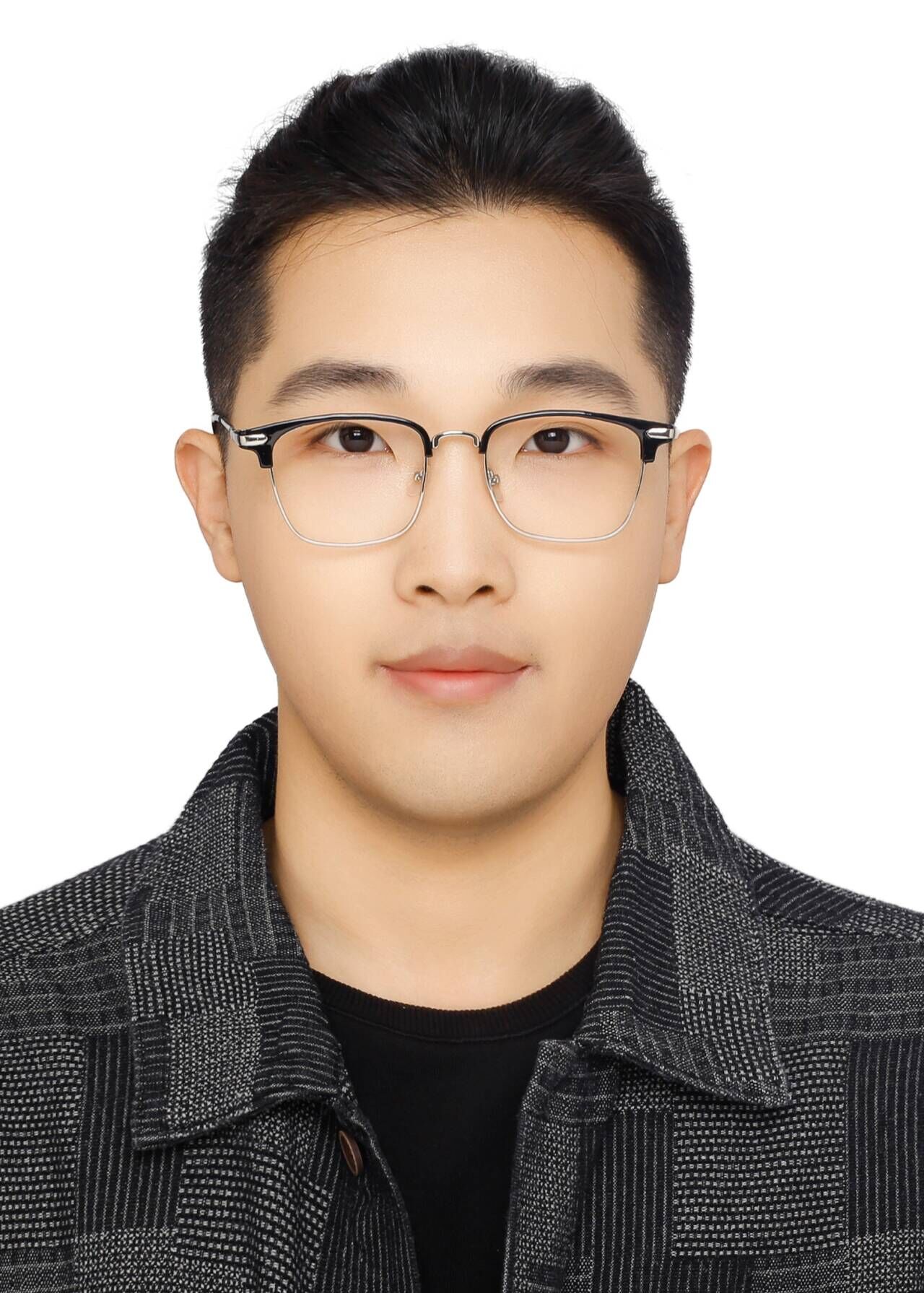}}]
  {Xiangkai Ma} is currently pursuing his Ph.D. at the School of Computer Science, Nanjing University. He earned his bachelor’s degree from the University of Electronic Science and Technology of China in 2022. His research interests include time series analysis, spatiotemporal data mining, large multimodal models, and embodied intelligence. He has authored multiple papers as the first author in top-tier journals and conferences such as ICDE and TKDD.
\end{IEEEbiography}
\vspace{-1.0cm}
\begin{IEEEbiography}[{\includegraphics[width=1in,height=1.25in,clip,keepaspectratio]{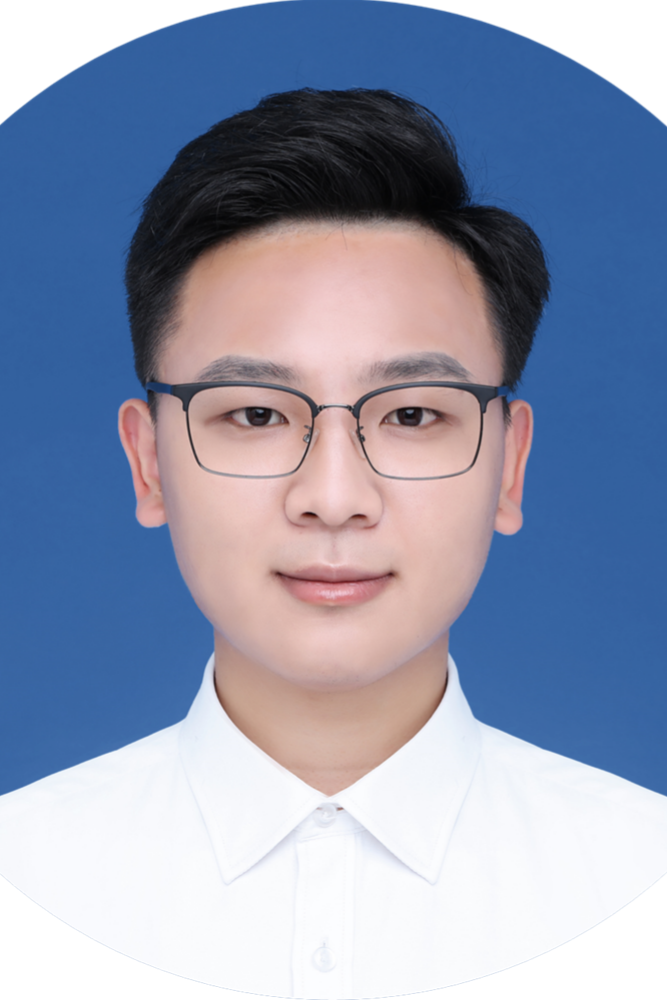}}]
  {Xiaobin Hong} received the MSc degree from Nanjing University of Science and Technology (211), Nanjing, China, and he is a current Ph.D student at the Department of Computer Science at Nanjing University (985/211) from 2022. He was the recipient of the Outstanding Master's Thesis award of the Jiangsu Computer Society. He has published about 20+ journal and conference papers, including JAS, Information Fusion, TKDD, AAAI, ICDE, ECCV, ICLR, ICRA, etc. He has served on the review committee of several journals and conferences, including TKDE, TOMM, AAAI, ICLR, CVPR, etc. His current research interests include Data Mining, Graph Learning, Time Series Analysis, LLM Reasoning, and AI4Science.
\end{IEEEbiography}
\vspace{-1.0cm}
\begin{IEEEbiography}[{\includegraphics[width=1in,height=1.25in,clip,keepaspectratio]{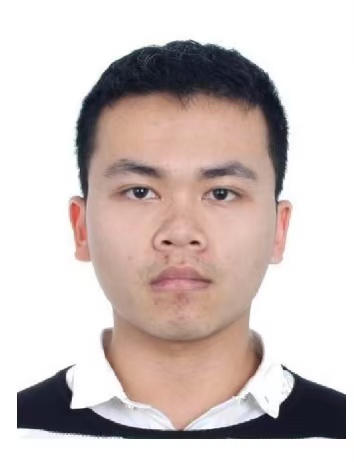}}]
  {Mingkai Lin} received the Ph.D degree from Nanjing University (985/211). He is a postdoctoral researcher at the School of Computer Science, Nanjing University.  His primary research focuses on graph learning. He has conducted systematic and in-depth research on key scientific issues such as large-scale social network sampling, graph neural network training optimization, and cross-domain generalization. His related research findings have been published in international journals such as TKDD and JAIR, as well as international conferences including WWW, AAAI, and ACM Multimedia. He was awarded the Best Paper Runner-Up at KSEM 2023.
\end{IEEEbiography}
\vspace{-1.0cm}
\begin{IEEEbiography}[{\includegraphics[width=1in,height=1.25in,clip,keepaspectratio]{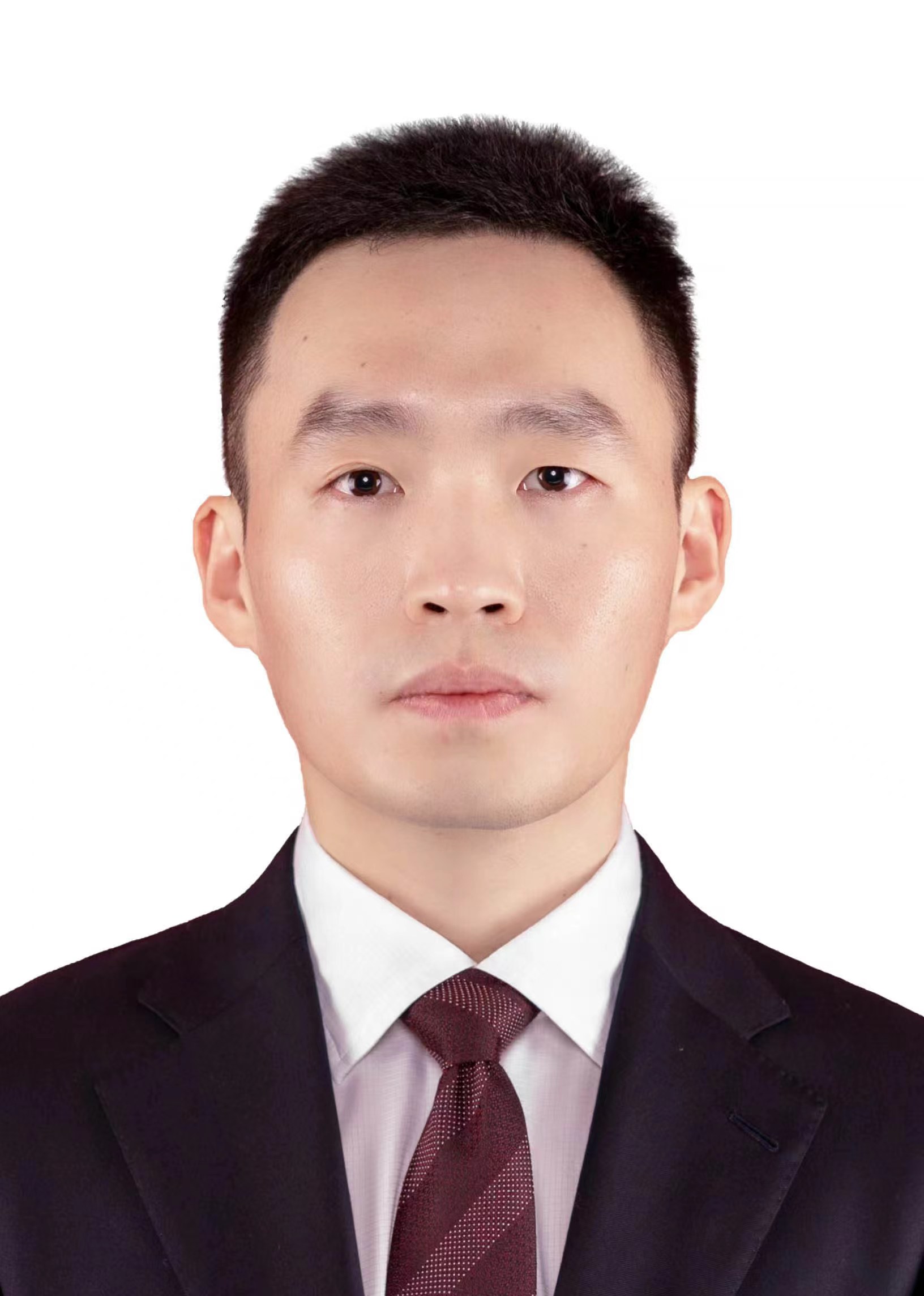}}]
  {Han Zhang} received his B.S. and M.S. degrees in Computer Science from Nanjing University, China. After graduating, he worked as a software engineer on the AI Platform team at Microsoft for four years. He is currently a Ph.D. candidate in the School of Computer Science at Nanjing University, with research interests in computer networking, deep learning, and embodied intelligence.
He has published multiple papers in venues such as INFOCOM and IEEE JSAC, with over 200 citations to date.
\end{IEEEbiography}
\vspace{-1.0cm}
\begin{IEEEbiography}[{\includegraphics[width=1in,height=1.25in,clip,keepaspectratio]{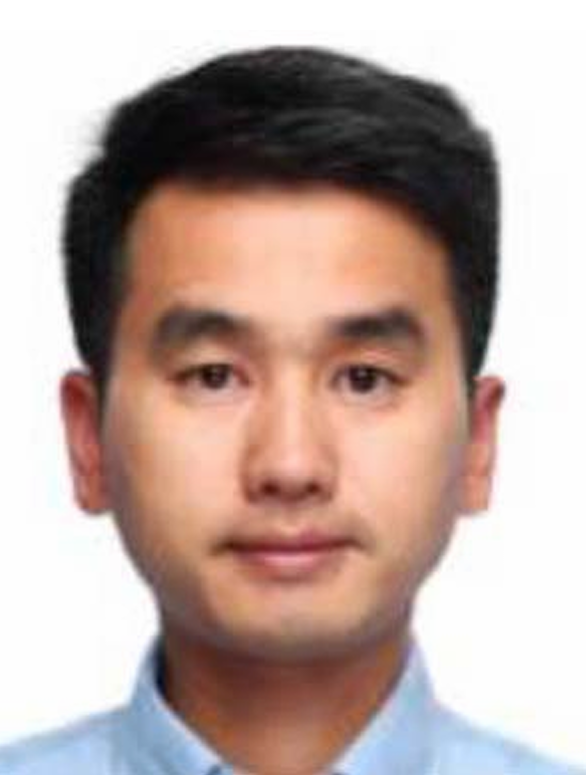}}]
  {Wenzhong Li} (Member, IEEE) receives the BS and PhD degree in computer science from Nanjing University, China. He was an Alexander von Humboldt Scholar fellow with the University of Goettingen, Germany. He is now a professor with the Department of Computer Science, Nanjing University. His research interests include distributed computing, Big Data mining and social networks. He has published more than 150 peer-review papers at international conferences and journals, which include INFOCOM, UBICOMP, AAAI, IJCAI, ACM Multimedia, CVPR, IEEE Communications Magazine, IEEE/ACM ToN, IEEE JSAC, IEEE TKDE, IEEE TPDS, etc. He served as program co-chair of MobiArch 2013 and Registration chair of ICNP 2013. He was the TPC member of several international conferences and the reviewer of many journals. He is the principle investigator of four fundings from NSFC, and the co-principle investigator of a China-Europe international research staff exchange program. He is a member of ACM, and China Computer Federation (CCF). He was featured on Elsevier’s Most Cited Chinese Researchers in 2022-2023. 
\end{IEEEbiography}
\vspace{-1.0cm}
\begin{IEEEbiography}[{\includegraphics[width=1in,height=1.25in,clip,keepaspectratio]{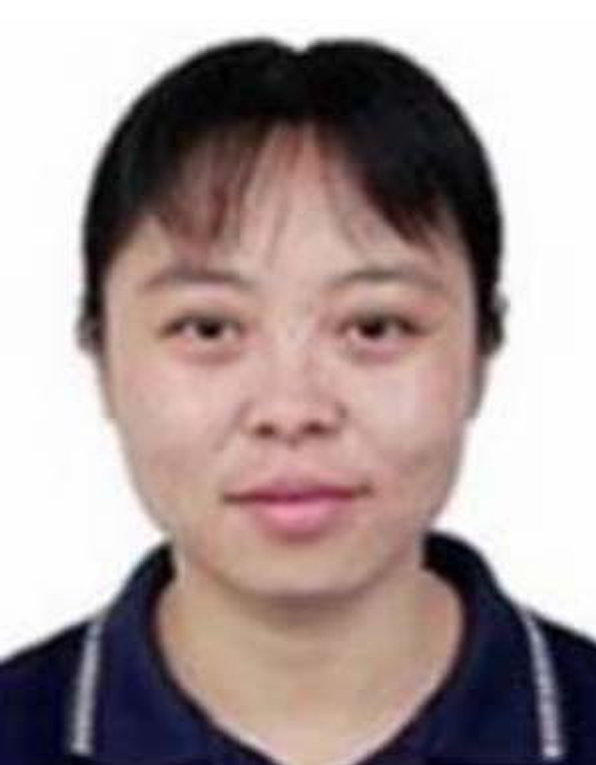}}]
  {Sanglu Lu} (Member, IEEE) received the BS, MS, and PhD degrees in computer science from Nanjing University in 1992, 1995, and 1997, respectively. She is currently a professor with the Department of Computer Science and Technology and the deputy Director of State Key Laboratory for Novel Software Technology.Her research interests include distributed computing, pervasive computing, and wireless networks. She has published more than 100 papers in referred journals and conferences in the above areas. She is a member of ACM.
\end{IEEEbiography}
\vspace{-1.0cm}

\clearpage
\input{segments/Appendix}

\vfill

\end{document}

%% file: segments/2-Introduction.tex
\section{Introduction}
Time Series (TS) data widely exist in many real-world fields~\cite{Fan2023DishTSAG,qiu2025duet}, such as power~\cite{Wang2022FewShotFA}, transportation~\cite{godahewa2021monash}, finance~\cite{Chi2022TimeSF}, etc. The wide application of time series analysis makes it of vital research significance in many practical fields. Empirical practice shows that time series data from different domains exhibit shifted statistical properties~\cite{Yuan2024DiffusionTSID}, such as period, frequency, data distribution, number of features, and fluctuation patterns, which poses a critical challenge to the generalizability and robustness of time series analysis.

A paramount challenge in modern time series analysis is temporal pattern fusion~\cite{fan2025medvia,wan2025csfformer,zhang2025new}, how to effectively integrate knowledge from these diverse sources to build models with profound generalization capabilities. Conventional models~\cite{Das2023TimesFM,shi2024timemoe,ansari2024chronos} often trained on a single domain, fail under substantial distribution shifts. This paper introduces Domain-Fusion, a novel paradigm that moves beyond simple multi-domain training. Domain-Fusion aims to create a cohesive and unified representation space by systematically blending the underlying temporal dynamics and statistical properties from multiple domains, thereby enabling models to perform zero-shot and few-shot inference on unseen domains.

\begin{figure*}[t]
\begin{center}
\vspace{-10pt}
\centerline{\includegraphics[width=2.0\columnwidth]{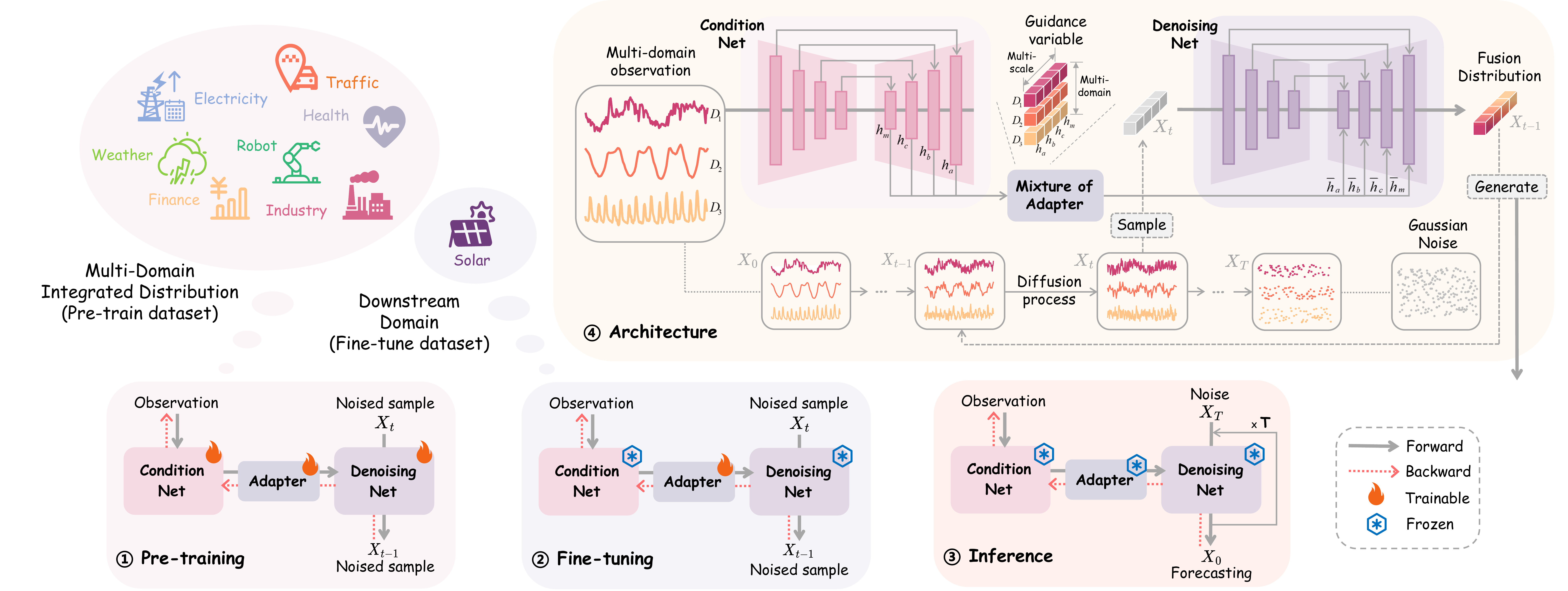}}
\caption{
Illustration of the paradigm in the domain-fused pre-training, fine-tuning and inference stages of the proposed \myformer architecture.
\textcircled{1} In the domain-fused training stage, all modules of \myformer are trained end-to-end on the fusion dataset with the forecasting task as the metric. 
\textcircled{2} In the fine-tuning stage, only the adapter module is allowed to continue training on a specific dataset. 
\textcircled{3} Finally, all the weights were frozen, and the prediction sequence was generated after \begin{small}$T$\end{small} rounds of iterative denoising.
Furthermore, \textcircled{4} presents the TimeControl's architecture.
\textbf{(a)} In the diffusion process, the original input sequence $X_0$ is progressively noised until degenerating into the Gaussian noise $X_T$. 
\textbf{(b)} In the context learning phase, mixed different domain sequences serve as inputs. The condition net captures cross-domain temporal fluctuation patterns as conditional variables to guide the generation process.
\textbf{(c)} In the denoising phase, model accepts representations from multiple domains to reconstruct the fusion distribution. Forecasting by iterative denoising process.
}\label{fig:intro_1}
\end{center}
\vspace{-20pt}
\end{figure*}

With the continuous development of deep learning, models based on DNN~\cite{zeng2023dlinear} and Transformer~\cite{Nie2023PatchTST}, remarkable achievements were made in time series analysis. 
Recently, with the success of diffusion models in the vision domain~\cite{zhang2023adding}, diffusion-based time series forecasting has also shown promising results. 
Early efforts like TimeGrad~\cite{Rasul2021timegrad} and CSDI~\cite{Tashiro2022csdi} grounded in diffusion models, progressively captures the dependencies in temporal patterns via markov chains.
Furthermore, Diffusion-TS~\cite{Yuan2024DiffusionTSID} and TimeDiff~\cite{shen2024timediff} employed diffusion model to generate all time points simultaneously.
However, existing approaches often focus on training domain-specific models tailored to individual datasets, limiting their ability to generalize well to unseen domains.

The success of Large Language Models (LLMs) \cite{Touvron2023Llama2O} has inspired the development of unified time series model. The unified model aims to achieve strong generalization capabilities and robustness to deliver satisfactory zero-shot inference performance on previously unseen domains. 
In the early explorations of unified time series models, LLM-based approaches~\cite{Chang2023LLM4TSAP,Jin2023TimeLLMTS,Zhou2023OneFA} leverage the alignment between text and time series modalities, utilizing pre-trained LLM, potentially with further fine-tuning. However, the inherent disparities between text and time series modalities may lead to concept drift and misalignment of representation dimensions. 
Recent attempts of establishing unified time series models~\cite{goswami2024moment,liu2024unitime,woo2024moirai} propose training a widely applicable model from scratch using data from multiple domains. 
Nevertheless, Different domains exhibit significant difference in data characteristics~\cite{Wang2025TimeMixerAG}.
Prior unified models often treat multi-domain data as a mere collection, lacking a principled methodology to fuse the domain-specific information into a coherent whole. 
These methods~\cite{woo2024moirai,goswami2024moment,liu2024unitime} struggle to construct a shared latent space that respects both the commonalities and peculiarities of each domain, leading to domain confusion rather than domain fusion~\cite{alsingscaling,shi2024scaling,shi2024timemoe}.

To address these challenges, this paper establishes a 
diffusion-based architecture explicitly designed for domain fusion in time series named \myformer for the first time.
The generative nature of diffusion models offers an ideal substrate for realizing the  domain fusion paradigm. Their exceptional capability to model complex, high-dimensional probability distributions, which ensures that they can learn and integrate temporal patterns from multiple domains.
Through simultaneously learning reverse denoising processes across multiple time-series domains, \myformer constructs a common representation space and creats a generative prior that is inherently cross-domain, enabling cross-domain generalization.
Our key contribution lies in tailoring the diffusion model explicitly for domain transfer by introducing multi-scale condition-denoising (MCD) and generation-style guidance (GSG) mechanisms. 
By explicitly modeling domain-invariant representations while dynamically controlling denoising trajectories via conditional information, TimeControl achieves superior predictive stability and cross-domain extrapolation.

Figure~\ref{fig:intro_1} shows the framework of \myformer, which contains three pivotal novel designs: multi-scale condition denoising modules, a hybrid architecture that accommodates observation sequences of arbitrary length and enables flexible prediction of future sequences with unrestricted length, and a ``plug-and-play'' adapter designed to support efficient fine-tuning.
We elaborate on implementations of these designs in two stages. In the context learning stage, sequences from different domains are fused together as inputs, from which the condition net captures a multi-scale representation of fluctuation patterns as context to guide the conditional generation at different levels. For example, shallow variables will guide to generate the trend part, and deep representations will guide the multi-periodic patterns. At each feature scale, global dependencies are captured by condition net, with the help of the adapter block, passes the fluctuation patterns at that scale to the denoising net. In the denoising stage, the contextual information is utilized to guide the process of reconstructing the forecast results from Gaussian noise, enhancing stability and accuracy. These contexts ensures that the diffusion model can model the inductive bias from different domain conditional probability distributions.
Notably, the hybrid architecture supports naturally efficient fine-tuning through the adapter. 
Specifically, during pre-training, a condition-denoising network with a large number of weights is fully frozen, while only a small number of weights in the adapter component need to be optimized. The condition net captures generic fluctuation patterns from observed sequences as conditional information, whereas the denoising net reconstructs sequences in the target domain based on specific fluctuation patterns. The adapter, enabled for fine-tuning, bridges the unified representation space with the proprietary representation space. The key contributions of our work are as follows.

\begin{itemize}
    \item 
    We propose TimeControl, the first work to systematically address domain fusion and generalization in time series utilizing the diffusion-based framework. TimeControl simultaneously models reverse denoising processes across multiple domains, constructing a common representation space via fused probability distribution.
    \item TimeControl introduces novel designs including multi-scale condition-denoising and generation-style guidance. By explicitly controlling the reverse process within a common space, it establishes distinct domain boundaries during the generation phase, ensuring robust and high-quality cross-domain extrapolation.
    \item TimeControl achieves superior predictive stability and cross-domain extrapolation in all 4 forecasting scenarios across 49 benchmarks. With overall performance improvements of 19.6\% and 21.2\% compared to existing 6 foundation models and 24 proprietary baselines. 
\end{itemize}


%% file: segments/3-RelatedWork.tex
\section{Related Work}
\subsection{Cross-Domain Generalization in Time Series}
\subsubsection{LLM-Empowered Models}
The first attempt to establish a unified time series model is OneFitsAll~\cite{Zhou2023OneFA}. OneFitsAll uses GPT2~\cite{Radford2019LanguageMA} pretrained from billions of tokens as a backbone, where it freezes the self-attention and feedforward layers in the pretrained language model and evaluates by fine-tuning the output and normalization layers on the time series data domain. 
However, the semantic information in the pre-trained model is difficult to be directly used in temporal scene. 
TimeLLM~\cite{Jin2023TimeLLMTS} uses text prototypes to reprogram the input sequence data and then feed it into the frozen Llama-2~\cite{Touvron2023Llama2O} to align the two modalities of time series and natural language. 
Recently, TimeVLM~\cite{zhong2025timevlm} proposes a novel multimodal framework that leverages pre-trained vision-language models to bridge temporal, visual, and textual modalities for enhanced forecasting. 
Although LLM-based models show good zero-shot inference ability, these models still face cross-modal challenges, and it is still urgent to establish a time series foundation model trained from scratch.

\subsubsection{Foundation Time Series Models}
Different from NLP and CV, the background knowledge and statistical characteristics of time series data from different domains often vary greatly~\cite{woo2024moirai,Ma2024TS3Net}, so it is challenging to train a unified time series model by utilizing multiple data domains. 
The first attempt to cross-domain training is UniTime~\cite{liu2024unitime}, which uses domain instructions and a Language-TS transfer module to provide recognition information to distinguish time series data from different domains, and uses masking technology to alleviate the problem of unbalanced domain convergence speed. 
On this basis, many time series foundation models have emerged. The first is the Moment~\cite{goswami2024moment} of encoder-only attention architecture with input patching.
TimesFM~\cite{Das2023TimesFM} proposes a decoder style attention model with input patching, using a large time-series corpus comprising both real-world and synthetic datasets.
Then, in order to overcome the differences between data domains, Moirai~\cite{woo2024moirai} based on mask encoder architecture includes multiple input-output projection layers to deal with different patterns of frequency-varying time series, and a spatio-temporal shared attention mechanism is designed. 
Chronos~\cite{ansari2024chronos} introduces a probabilistic time series models, which tokenizes time series values using scaling and quantization into a fixed vocabulary and trains existing models on these tokenized time series via the cross-entropy loss.

Previous approaches of cross-domain generalization on time series~\cite{Chang2023LLM4TSAP,Jin2023TimeLLMTS,woo2024moirai,Zhou2023OneFA} aim to train a broadly applicable model from multiple domains, and focus on designing a generic architectures that are generic for diverse temporal domain distributions. Nevertheless, varying statistical characteristics ~\cite{goswami2024moment} across domains make it challenging to design a common pipeline that effectively handles time series with distinct semantics. This often leads to domain confusion, significantly limiting the zero-shot and cross-domain generalization capabilities of such models.

Unlike prior approaches, by leveraging the diffusion model's excellent capability to model probability distributions, our approach directly produces diverse and high-quality forecasts by modelling a fusion probability distributions over multiple time series domains without establishing any inter-series projections. Concretely, we trained \myformer from scratch in integrated distribution. By modeling the multi-domain denoising processes, building a shared representation space for cross-domain generalization. It also forms implicit boundaries for domain-specific patterns and uses multi-scale context and generation-style guidance to align with input distributions (e.g., seasonality). The latent probability representation from different domains allows the model to generate high-quality time series samples in challenging scenarios.



\subsection{Diffusion Models in Time Series}
The generative prediction paradigm based on diffusion has become a promising methodology for time series forecasting. 
TimeGrad~\cite{Rasul2021timegrad} is the first time to use diffusion model to model the probability distribution and predict future series in an autoregressive way. 
Subsequently, CSDI~\cite{Tashiro2022csdi} and SSSD~\cite{alcaraz2022sssd} designed a diffusion model with the observed data as the context condition, and filled the missing part by introducing the noise of the diffusion process, and then gradually denoising at each step.
To improve the learning ability to model long-term dependencies in time series data, TimeDiff~\cite{shen2024timediff} introduces autoregressive initialization mechanisms to predict all timepoints of future sequence. 
DiffusionTS~\cite{Yuan2024DiffusionTSID} utilizes the transformer architecture to model the seasonal-trend components separately, and the Fourier-based losses are designed to reconstruct the sequence sample directly rather than the noise at each diffusion step. 
Inspired by LDM, recent LDT~\cite{li2024ldt} utilizes a transformer-based autoencoder to learn latent representations from raw observation sequence and subsequently predicts future sequence in a non-autoregressive manner in the latent space.
Although there have been a lot of methodologies~\cite{liu2024diffusion,shen2024timediff,Tashiro2022csdi} on diffusion to time series, to the best of our knowledge, there are few researches about multi-domain distribution fusion and generalization. 
The establishment of \myformer faces the challenge of capturing distributions from multiple domains, and limited inductive biases may not be sufficient for models to capture distribution characteristics of different domains~\cite{shen2024mrdiffusion}. 
Therefore, we strongly recommend the independent and complex component ensures that the diffusion model can capture multi-scale domain specified patterns on the denoising process. 

Compared with existing approaches~\cite{shen2024timediff,Tashiro2022csdi} that focus on prioritize denoising generation mechanics, our method adopts a two-stage diffusion paradigm that integrates conditional learning and denoising with effective representation understanding and generation modules that clarify domain boundaries. 
The proposed adapter design supports efficient fine-tuning, preserving pre-trained temporal details and aligning source and target domains for high-quality cross-domain generalization. 
To the best of our knowledge, this is the first diffusion architecture designed to support domain-fused training and fine-tuning, systematically addressing the challenges of domain generalization.

%% file: segments/4-Methodology.tex
\section{Preliminaries} \label{section:preliminaries}
In the long-term forecasting task, $X_{-L+1:0}^0\in\mathbb{R}^{d\times L}$ and $X_{1:H}^0\in\mathbb{R}^{d\times H}$ are utilized to represent the observed series and the future series, respectively, where $d$ denotes the number of channels of the multivariate time series, $L$ and $H$ denote the lookback window and forecast horizon.

\begin{figure*}[ht]
\begin{center}
\vspace{-10pt}
\centerline{\includegraphics[width=1.9\columnwidth]{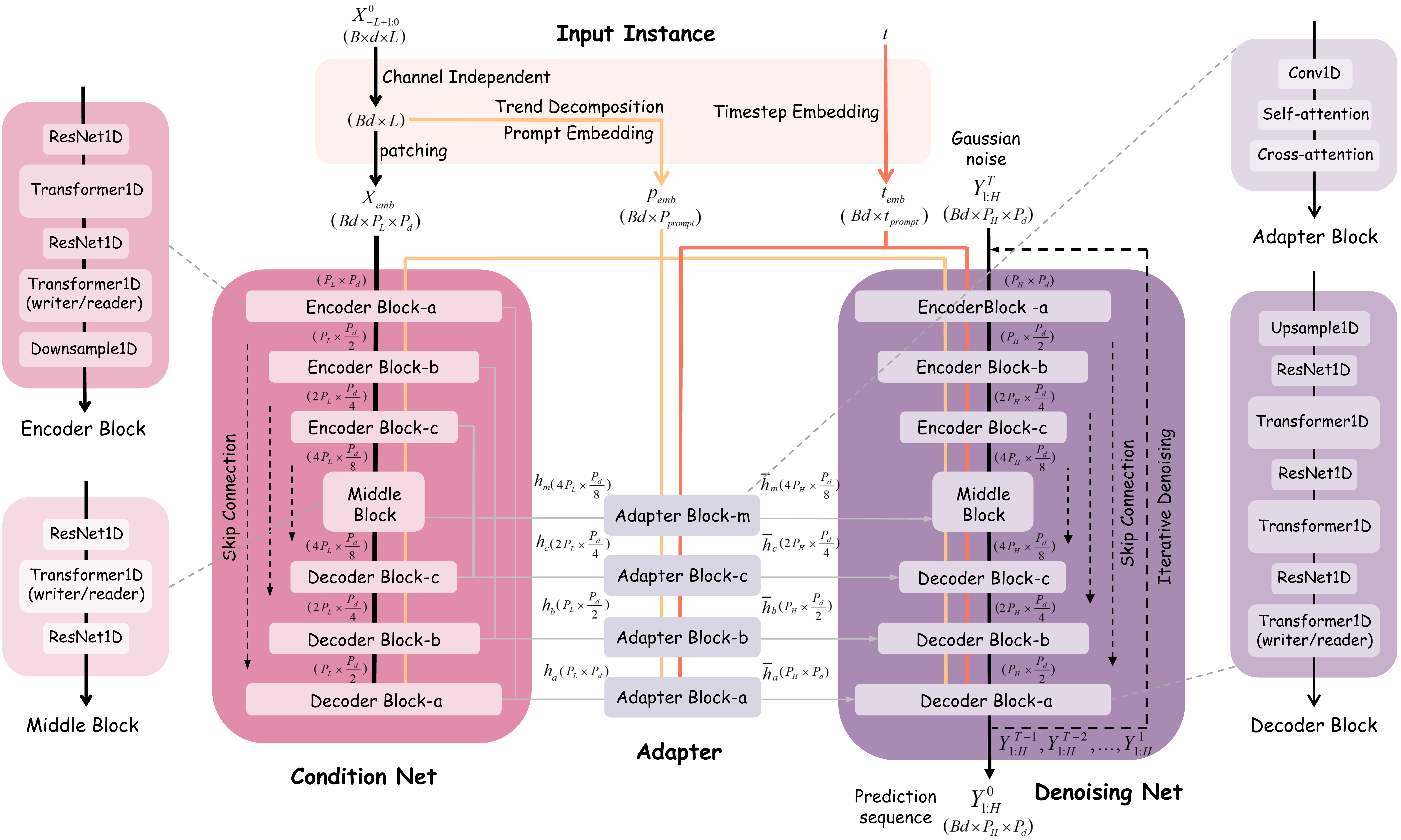}}
\vspace{-10pt}
\caption{
The overall framework of \myformer. Specifically, the observation sequence \begin{small}$X_{-L+1:0}^{0}$\end{small} and the diffusion timestep $t$ are processed by the input instance module to obtain the lookback embedding, trend-prompt embedding, and time embedding, which serve as inputs to the condition-denoising net and the adapter. The condition net captures the multi-scale representations \begin{small}$h_{a,b,c,m}$\end{small} and the adapter transforms those into the context variables \begin{small}$\overline{h}_{a,b,c,m}$\end{small} which utilized to guide conditional generation process for the prediction task.
}\label{fig:method_1}
\end{center}
\vspace{-20pt}
\end{figure*}

\subsection{Diffusion Model}
The Denoising Diffusion Probabilistic Models (DDPM)~\cite{Ho2020DenoisingDP} consists of two processes, the forward diffusion and the reverse denoising processes, see appendix for details.

\subsubsection{Forward Process} In the forward process, A set of real time series samples \begin{small}$X_{1:H}^{0}\sim q(X)$\end{small} are gradually noised until they degenerate into gaussian distribution \begin{small}$X_{1:H}^T\sim N(0,\mathrm{I})$\end{small}.
The complete diffusion process is regarded as the Markov chain, and the diffusion process at timestep \begin{small}$t\in[1,T]$\end{small} is represented:
\begin{equation}
  \begin{split}
    q\left(X_{1:H}^t\mid X_{1:H}^{t-1}\right)\textit{=}N\left(X_{1:H}^t;\sqrt{1-\beta_t}X_{1:H}^{t-1},\beta_t\mathbf{I}\right),
  \end{split}
\end{equation}
where \begin{small}$\beta_{t}\in(0,1)$\end{small} is the diffusion coefficient and $T$ is the length of the Markov chain.
The diffusion result \begin{small}$X_{1:H}^{t}$\end{small} corresponding to any number of steps \begin{small}$t$\end{small} can be computed from \begin{small}$X_{1:H}^{0}$\end{small} via:
\begin{equation}
  \begin{split}
    X_{1:H}^t\text{=}\sqrt{\prod_{i=1}^t\left(1\text{-}\beta_i\right)}\cdot X_{1:H}^0\text{+}\sqrt{1\textit{-}\prod_{i=1}^t\left(1\text{-}\beta_i\right)}\cdot\varepsilon.
  \end{split}
\end{equation}

\subsubsection{Reverse Process} In the reverse process, the deep model is utilized to progressively denoise from gaussian distribution.
The denoising process at timestep \begin{small}$t$\end{small} is represented as \begin{small}$p_\theta\left(X_{1:H}^{t-1}\mid X_{1:H}^t,c\right)$\end{small}, where \begin{small}$c$\end{small} represents the condition variable calculated from the observation sequence, \begin{small}$\mu_\theta(X_{1:H}^t,t)$\end{small} represents the denoising model established at the diffusion timestep \begin{small}$t$\end{small}, \begin{small}$\theta$\end{small} represents parameters, and \begin{small}$\sigma_{t}$\end{small} serves as a hyperparameter.
Following DDPMs, we calculate the MSE between the \begin{small}$\mu_{\theta}(X_{1:H}^{t},t)$\end{small} and the mean \begin{small}$\mu\Big(X_{1:H}^{t},X_{1:H}^{0}\Big)$\end{small} of the posterior distribution \begin{small}$q\left(X_{1:H}^{t-1}\mid X_{1:H}^0,X_{1:H}^t\right)$\end{small} as the loss function:
\begin{equation}
  \begin{split}
    L\text{=}\sum_{t=1}^TE_{q\left(X_{1:H}^t|X_{1:H}^0\right)}\|\mu\left(X_{1:H}^t,X_{1:H}^0\right)\text{-}\mu_\theta\left(X_{1:H}^t,t\right)\|^2.
  \end{split}
\end{equation}

\subsection{Generation-Style Guidance}
TimeControl proposes a conditional generative paradigm time-series forecasting. 
In addition, conditional diffusion models face a critical challenge in forecasting tasks where observed sequences must guide denoising process to ensure causal consistency between generated future series and input history series. 
To enhance control over conditional denoising processes, TimeControl employs generation-style guidance, a technique that simultaneously generates both conditional and unconditional denoising outputs during inference. 
By linearly interpolating these two outputs with a control coefficient $\lambda$, our approach dynamically balances free exploration diversity and target domain style control by input conditions, thereby enhancing generalization.
The context representation captured by the condition net from the observation sequence is denoted as the condition variable \begin{small}$c$\end{small}, so that the goal of the reverse denoising process can be described as:
\begin{equation}
  \begin{split}
    p_\theta\left(X_{1:H}^{0:T}\mid c\right)\textit{=}p_\theta\left(X_{1:H}^T\right)\Pi_{t=1}^Tp_\theta\left(X_{1:H}^{t-1}\mid X_{1:H}^t,c\right),
  \end{split}
\end{equation}
where \begin{small}$X_{1:H}^T\sim N(0,\mathrm{I})$\end{small} represents the initial state obtained by sampling from gaussian distribution. Furthermore, according to the Bayesian formula, we have:
\begin{equation}
  \begin{split}
    p_\theta\left(X_{1:H}^{t-1}\mid X_{1:H}^t,c\right)\text{=}\frac{p_\theta\left(X_{1:H}^{t-1}\mid X_{1:H}^t\right)\cdot p_\theta\left(c\mid X_{1:H}^{t-1},X_{1:H}^t\right)}{p_\theta\left(c\mid X_{1:H}^t\right)}.
  \end{split}
\end{equation}
To obtain the probability score function to control the condition generation process, we run gradient update on \begin{small}$X_{1:H}^{t-1}$\end{small} and the score function as follows:
\begin{equation}
  \begin{split}
    \nabla_{X}\log p_\theta\left(X_{1:H}^{t-1}\mid X_{1:H}^t,c\right)\textit{=}&\nabla_{X}\log p_\theta\left(X_{1:H}^{t-1}\mid X_{1:H}^t\right) \\
    \text{+}&\nabla_{X}\log p_\theta\left(c\mid X_{1:H}^{t-1},X_{1:H}^t\right).
  \end{split}
\end{equation}
Where \begin{small}$\log p_\theta\left(X_{1:H}^{t-1}\mid X_{1:H}^t\right)$\end{small} and \begin{small}$\log p_\theta\left(X_{1:H}^{t-1}\mid X_{1:H}^t\right)$\end{small} represent the denoising model and classifier, respectively, which are used to approximate the sampling result \begin{small}$\log p_\theta\left(X_{1:H}^{t-1}\mid X_{1:H}^t,c\right)$\end{small} representing the posterior distribution.
During training and inference, we propose a high-quality conditional probabilistic diffusion paradigm GSG: 
First, based on the Bayesian formula, we derive the following procedure: 
\begin{equation}
  \begin{split}
    \log p_\theta\left(c\mid X_{1:H}^{t-1},X_{1:H}^t\right)\text{=}\log p_\theta\left(X_{1:H}^{t-1}\mid X_{1:H}^t,c\right) \\
    \text{+}\log p_\theta\left(c\mid X_{1:H}^t\right)\text{-}\log p_\theta\left(X_{1:H}^{t-1}\mid X_{1:H}^t\right).
  \end{split}
\end{equation}
Then we calculate the gradient with respect to the variable \begin{small}$X_{1:H}^{t-1}$\end{small}, thus obtaining formula:
\begin{small}
\begin{equation}
  \begin{split}
    \nabla_{X}\log p_\theta\left(c\mid X_{1:H}^{t-1},X_{1:H}^t\right)\text{=}&\nabla_{X}\log p_\theta\left(X_{1:H}^{t-1}\mid X_{1:H}^t,c\right) \\
    \text{-}&\nabla_{X}\log p_\theta\left(X_{1:H}^{t-1}\mid X_{1:H}^t\right).
  \end{split}
\end{equation}
\end{small}
Subsequently, this formula is substituted into the score function, and generation-style guidance is: 
\begin{small}
\begin{equation}
  \begin{split}
    \nabla_{X}\log p_{\theta,\lambda}\left(X_{1:H}^{t-1}\mid X_{1:H}^t,c\right)\text{=}&\lambda\nabla_{X}\log p_\theta\left(X_{1:H}^{t-1}\mid X_{1:H}^t,c\right) \\
    \text{+}(1\textit{-}&\lambda)\nabla_{X}\log p_\theta\left(X_{1:H}^{t-1}\mid X_{1:H}^t\right),
  \end{split}
\end{equation}
\end{small}
which $\lambda$ is the control parameter used to combine two formulas.
Where \begin{small}$\log p_{\theta,\lambda}(X_{1:H}^{t-1}\mid X_{1:H}^t,c)$\end{small}, \begin{small}$\log p_\theta(X_{1:H}^{t-1}\mid X_{1:H}^t)$\end{small}, and \begin{small}$\log p_\theta(X_{1:H}^{t-1}\mid X_{1:H}^t,c)$\end{small} represents the final output, unconditional output, conditional output of the denoising net, respectively. 
In the implementation, condition net first accepts the observation sequence as input and subsequently outputs the captured multi-scale representation as the observation sequence context. 
Then, two identical initial samples are sampled from the gaussian noise, and two output sequences are generated by iterative denoising with the observation sequence context and zero vector as conditional variables, respectively, which are conditional- and unconditional- output. Finally, we calculate the final output of the model based on the weight specified by user. We provide pseudocode in the Section~\ref{sec:algorithm} to illustrate how GSG is used for training and sampling.

\begin{figure*}
\begin{center}
\vspace{-10pt}
\centerline{\includegraphics[width=1.85\columnwidth]{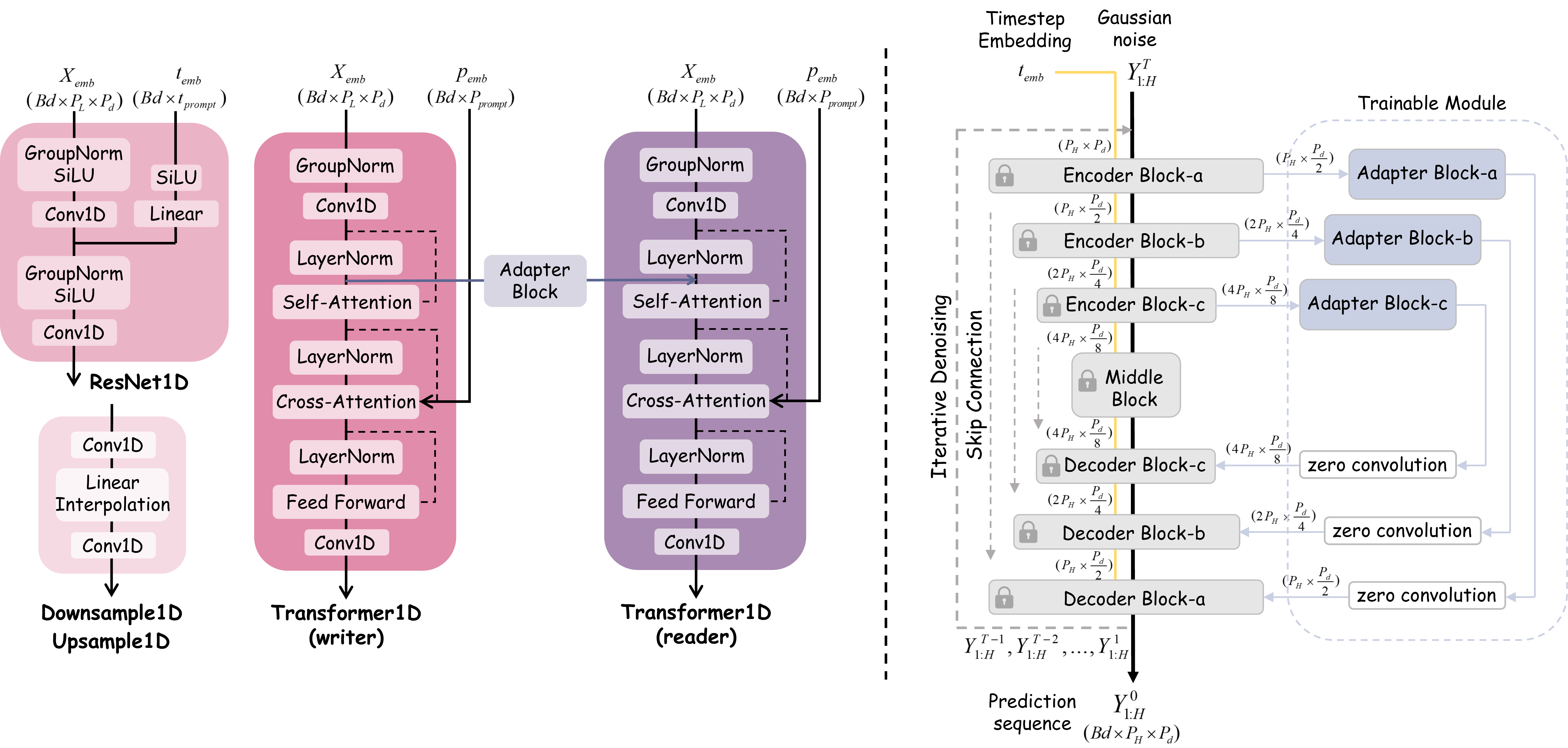}}
\vspace{-10pt}
\caption{
(a): Illustration of three sub-modules, where Transformer1D contains two versions that are utilized to establish condition-denoising net, respectively. Downsample1D and Upsample1D contain different linear interpolation layers.
(b): In the non-conditional generation paradigm, the \myformer architecture can be divided into there components, where the locked grey block shows the backbone of the pre-trained obtained \myformer. based on this, several trainable consecutive adapter blocks (blue blocks) are added with a set of zero convolution layer (white) to construct the fine-tuned network. Which is used to align the multi-domain uniform space with the specified domain representation space.
}\label{fig:method_2}
\end{center}
\vspace{-20pt}
\end{figure*}

\section{The Domain-Fusion Paradigm of \myformer} \label{section:unidiff}
The core idea of domain fusion is to learn a unified generative model from multiple source domains, such that the model captures a fused distribution that subsumes the essential temporal patterns from all domains. 
This fused knowledge serves as a strong prior, enabling the model to quickly adapt to or directly generate data for a novel target domain. TimeControl instantiates this paradigm through a conditional diffusion model, where the condition embodies the fused knowledge.
Our design insight is that simultaneously learning reverse denoising processes across multiple time series domains to construct a common representation space. 
Subsequently, TimeControl leverages historical sequences to guide the denoising process, ensuring alignment between output with input distributions (e.g., seasonality), to enhance cross-domain generalization. 
By conditioning on observation in the target domain, the model generates predictions that adapt to domain-specific characteristics, enabling robust generalization even under shifts such as scaling, noise, or partial feature alignment.

Another critical challenge lies in the fact that long-term forecasting requires models to extract sufficient temporal information to serve as conditional context, thereby generating forecast sequences that adhere to the distributional characteristics of the target domain. 
Nevertheless, existing condition diffusion models~\cite{Tashiro2022csdi,li2024ldt,Yuan2024DiffusionTSID} often use simple design to capture the conditional variable on single scale. 
However, sequences from several domains often have different multi-scale latent representations, such as sampling rate, periodic frequency characteristics, multi-periodic patterns~\cite{Ma2024TS3Net}, etc. 
Due to the difficulty for the model to learn enough fluctuation pattern information from the lookback window and the randomness of the initial state, diffusion model demonstrates low accuracy in the prediction task~\cite{shen2024mrdiffusion}.

To address these challenges, \myformer focus on tailoring the model explicitly for domain transfer by introducing multi-scale condition-denoising (MCD) and generation-style guidance (GSG) mechanisms. 
By explicitly modeling domain-invariant representations while dynamically controlling denoising trajectories via conditional information, TimeControl achieves superior predictive stability and cross-domain extrapolation compared to existing approaches.

\vspace{-10pt}
\subsection{Input Observation Instance}
Based on the channel independent design, observation sequence \begin{small}$X_{-L+1:0}^{0}\in\mathbb{R}^{B\times d\times L}$\end{small} is first processed as \begin{small}${X_{-L+1:0}^{0}}^{\prime}\in\mathbb{R}^{Bd\times L}$\end{small}, where \begin{small}$B$\end{small} denotes the batch size, and then following the patching instance strategy with no padding and no overlap, each sequence of length \begin{small}$L$\end{small} is represented as \begin{small}${X_{-L+1:0}^{0}}^{\prime\prime}\in\mathbb{R}^{Bd\times P_{L}\times P_{d}}$\end{small}, where \begin{small}$P_{d}$\end{small} and \begin{small}$P_{L}$\end{small} as the dimension and number of tokens, and \begin{small}$P_{d}\times P_{L}\textit{=}L$\end{small}. 
Subsequently, observation tokens \begin{small}$X_{_{emb}}\in\mathbb{R}^{Bd\times P_{L}\times P_{d}}$\end{small} are computed from \begin{small}${X_{-L+1:0}^{0}}^{\prime\prime}$\end{small} by the embedding layer. 
In addition to the observation tokens, \myformer accepts two input. Firstly, the historical trend is considered to be very critical information for the prediction, we proposes to utilize the embedding \begin{small}$p_{emb}$\end{small} of the trend segment as prompt vector. 
Besides, since the reverse denoising process requires the diffusion timestep \begin{small}$t$\end{small} as guidance information. 
This paper proposes to compute the embedding of \begin{small}$t$\end{small} via a time encoder. 

\vspace{-10pt}
\subsection{Condition-Denoising Structure}
Firstly, the architecture composition of the conditional learning and denoising generation module are introduced. For convenience, the term ``Block'' is utilized to refer to a set of consecutive neural network, which are reused as important components in building the \myformer structure.
In Figure~\ref{fig:method_1}, condition net and denoising net both follow the design of U-Net~\cite{2015Unet}, where the encoder and decoder are composed of Encoder Block$_{a,b,c}$ and Decoder Block$_{a,b,c}$. Besides, the Middle Block is designed to connect the encoder and decoder. Encoder Block$_{a}$ and Decoder Block$_{a}$ have the same feature dimension \begin{small}$P_{d}/2$\end{small} and skip connections are established between the two blocks, similarly, the other Blocks have dimensions \begin{small}$P_{d}/4$\end{small} and \begin{small}$P_{d}/8$\end{small}, respectively. We provide pseudocode in the Section~\ref{sec:algorithm} to illustrate \myformer's architecture.

\subsubsection{Condition Net for Domain-Fused Representation Learning}
The Condition Net is the cornerstone of our fusion process. During pre-training, it takes mixed-domain sequences as input and learns to extract domain-fused contexts. Its multi-scale architecture learns to identify and blend shared temporal motifs that are universally useful, while also preserving a capacity to represent domain-specific nuances.

Specifically, condition net accepts observation tokens \begin{small}$X_{emb}$\end{small} as input, and Decoder Block$_{a}$ at the end does not output any result. Specifically, the deep representation included in the observation sequence undergoes consecutive Blocks to obtain the multi-scale historical fluctuation pattern \begin{small}$\{h_{m,a,b,c}\}$\end{small}, as shown in Figure~\ref{fig:method_1}. This set of temporal representations is passed through Adapter Blocks with the same structure to obtain a set of conditional variables \begin{small}$\{\overline{h}_{m,a,b,c}\}$\end{small}. This is passed as a condition context to several Blocks in denoising net.

\subsubsection{Denoising Net for Guided Generation} 
In the denoising generation stage, \begin{small}$Y_{1:H}^T\in\mathbb{R}^{Bd\times P_H\times P_d}$\end{small} obtained by sampling from gaussian distribution \begin{small}$N(0,\mathrm{I})$\end{small} is utilized as the initial input, where \begin{small}$P_{H}=H/P_{d}$\end{small}. In each round of denoising iteration, the denoising net accepts the diffusion timestep \begin{small}$t$\end{small} and \begin{small}$Y_{1:H}^t$\end{small} as input, and its output is utilized to calculate the sample \begin{small}$Y_{1:H}^{t-1}$\end{small} for the next round of denoising iteration. After \begin{small}$T$\end{small} rounds of denoising process, \begin{small}$Y_{1:H}^0\in\mathbb{R}^{Bd\times P_H\times P_d}$\end{small} undergoes the flatten and channel independent to obtain the prediction result \begin{small}$Y_{1:H}^{0}\in\mathbb{R}^{B\times d\times H}$\end{small}. In particular, during inference, condition net only needs to learn a set \begin{small}$\{\overline{h}_{m,a,b,c}\}$\end{small} from observations, and in the subsequent all \begin{small}$T$\end{small} rounds of iterations, denoising net reuses historical patterns to predict at different timesteps \begin{small}$t$\end{small}.

\vspace{-10pt}
\subsection{Blocks Implementation}
All Blocks included in condition-denoising net are built from two smaller modules, ResNet1D and Transformer1D (writer/reader), as shown in Figure~\ref{fig:method_1}.
The ResNet1D accepts the embedding \begin{small}$t_{emb}$\end{small} of diffusion timesteps \begin{small}$t$\end{small} (only in the denoising generation stage) and the latent representation \begin{small}$X_{emb}$\end{small} as input, which contains two convolutions. After the first convolution layer, the latent representation \begin{small}$X_{emb}$\end{small} is added to the timestep embedding \begin{small}$t_{emb}$\end{small} through the linear layer, and the output is obtained by second convolution layer.
The Transformer1D has two versions, writer and reader, which constitute Blocks in condition and denoising net, respectively. 
Transformer1D-writer accepts the latent representation \begin{small}$X_{emb}\in\mathbb{R}^{Bd\times P_{L}\times P_{d}}$\end{small} as input, where self-attention mechanism is utilized to capture dependencies between global patches in the lookback window. 
Transformer1D-reader accepts the condition variable \begin{small}$\{\overline{h}_{m,a,b,c}\}$\end{small}, noised sample \begin{small}$Y_{1:H}^t\in\mathbb{R}^{Bd\times P_H\times P_d}$\end{small}, and the trend embedding \begin{small}$p_{emb}$\end{small} (as prompt) as input, where self-attention concatenates the context into the key-value vector, thereby using historical fluctuation patterns as contextual information to guide the denoising process. In the subsequent cross-attention, the historical trend contained in \begin{small}$p_{emb}$\end{small} is used to model the long-term trend of prediction. 
The proposed condition-denoising net establishes the direct connection between the generated sequence space and the observed sequence through Transformer1D-writer,reader, ensuring that \myformer has strong generalization capabilities to address the challenge of cross-domain probability distribution modeling. 
The implementation details are given in the Figure~\ref{fig:method_2}.

\vspace{-10pt}
\subsection{Adapter for Fusion-to-Target Alignment}
The universal temporal features learned by condition net in the unified data domain space may not be adapted to all downstream data domains. In order to deal with cross-domain challenges, it is necessary to design fine-tuning strategies that can adapt to the diffusion and denoising process. These reasons make it essential to establish a projection relationship between condition and denoising net to improve the generalization ability and robustness of the model. 

To explore the possibility of \myformer becoming a domain-fused training paradigm, we have designed a fine-tuning strategy based on a ‘plug-and-play’ adapter for \myformer, enabling the diffusion and denoising processes to be compatible with the pre-training fine-tuning paradigm. And in order to accommodate the multi-scale representation in the observation window, the proposed adapter structure contains seven identical adapter blocks and the feature dimensions correspond to the Unet Blocks, as shown in Figure \ref{fig:method_1}.

Specifically, domain-fused training to obtain a condition denoising net with a large number of weights (condition and denoising net) is completely frozen, and only a small number of weights (mixture of adapter) in the adapter component need to be optimised. This hybrid structure can avoid catastrophic forgetting and better cope with cross-domain challenges. Meanwhile, the condition and denoising net adopts encoder-decoder structure which is composed of complex stacked blocks. In contrast, the adapter is only composed of a single-layer convolution and a Transformer layer. This means that fine-tuning only the mixture of adapter structure has a negligible impact, and its time cost can be comparable to that of mainstream lightweight models (such as PatchTST). 

The effectiveness of the fine-tuning strategy can be intuitively explained by the fact that the pre-trained condition net is responsible for capturing generic fluctuation patterns from observed sequences, the denoising net is required to reconstruct sequence samples from noise in the target domain based on specific fluctuation patterns, and adapter is used to connect the unified representation space with the proprietary representation space. In additional, In adapter, the innovative \textbf{1 $\times$ 1 Conv1D} is designed to align the number of tokens in the observation space and the forecasting space. These hybrid structures ensure that \myformer has the flexibility to generate high-quality prediction sequences of arbitrary length. 

\vspace{-10pt}
\begin{footnotesize}
\begin{algorithm}[htbp]
   \caption{Training of \myformer}
   \label{alg:training}
\begin{algorithmic}[1]
   \STATE {\bfseries Input:} The observation series $X_{-L+1:0}^0$ and prediction series $X_{1:H}^0$, the iteration \begin{small}$N_{\mathrm{iter}}$\end{small}, the sequence of noise levels \begin{small}$\{\beta_t\mid t\in[1,T]\}$\end{small}, the diffusion steps \begin{small}$T$\end{small}.
   \STATE {\bfseries Output:} Optimized \textit{Condition Net}, \textit{MoA} and \textit{Denoising Net}.
   \FOR{ $i=1$ {\bfseries to} $N_{\mathrm{iter}}$}
   \STATE sample denoising steps: \begin{small}$t\sim \textit{Uniform}(\{1,\cdots,T\})$\end{small}, \\
   \STATE 
          \begin{small}{$X_{-L+1:0}^0\to X_{-L+1:0}^\textit{trend},X_{-L+1:0}^\textit{seasonal}$}\end{small};\\
          \begin{small}{$X_{-L+1:0}^{\textit{seasonal}}\in\mathbb{R}^{{Bd}\times L}\to X_{\textit{emb}}\in\mathbb{R}^{{Bd}\times P_L\times P_d}$}\end{small};\\
          \begin{small}{$X_{-L+1:0}^\textit{trend}\in\mathbb{R}^{Bd\times L}\to p_\textit{emb}\in\mathbb{R}^{Bd\times p_\textit{prompt}}$}\end{small}.
   \STATE 
          \begin{small}{$\begin{pmatrix}h_m,h_c,h_b,h_a\end{pmatrix}\textit{=ConditionNet}(X_{emb},p_{emb})$}\end{small}, \\
          \begin{small}{$\left(\overline{h}_m,\overline{h}_c,\overline{h}_b,\overline{h}_a\right)\textit{=MoA}\left(h_m,h_c,h_b,h_a\right)$}\end{small}.
   \STATE 
          \begin{small}{$t\to t_\textit{emb}\in\mathbb{R}^{Bd\times t_\textit{prompt}}$}\end{small}. \\
          \begin{small}{$X_{1:H}^t\textit{=}\sqrt{\prod_{i=1}^t\left(1\textit{-}\beta_i\right)} X_{emb}\textit{+}\sqrt{1\textit{-}\prod_{i=1}^t\left(1\textit{-}\beta_i\right)}\varepsilon,\varepsilon\sim N\left(0,\mathrm{I}\right)$}\end{small}, \\
          \begin{small}{$\tilde{Y}_{1:H}^{t-1}\textit{=DenoisingNet}\left(X_{1:H}^t,t_\textit{emb},p_\textit{emb},\left(\overline{h}_m,\overline{h}_c,\overline{h}_b,\overline{h}_a\right)\right)$}\end{small}, \\
          \begin{small}{$\hat{Y}_{1:H}^{t-1}\textit{=DenoisingNet}\left(X_{1:H}^t,t_\textit{emb}\right)$}\end{small}, \\
          \begin{small}{$Y_{1:H}^{t-1}\textit{=}\lambda\tilde{Y}_{1:H}^{t-1}+(1-\lambda)\hat{Y}_{1:H}^{t-1}$}\end{small}. \\
   \STATE \begin{small}{$\mathcal{L}_{mse}\textit{=MSE}\left(Y_{1:H}^{t-1},X_{1:H}^{t-1}\right)$}\end{small}. \\
          Update model parameters based on $\mathcal{L}_{mse}$ gradient.
   \ENDFOR
\end{algorithmic}
\end{algorithm}
\end{footnotesize}

\vspace{-20pt}
\subsection{Algorithm for training and sampling of \myformer} \label{sec:algorithm}
Motivated by the success of ControlNet~\cite{zhang2023adding} in the field of visual generation. To enhance control over conditional denoising processes, TimeControl employs generation-style guidance (GSG), a technique that simultaneously generates both conditional and unconditional denoising outputs during inference. By linearly interpolating these two outputs with a control coefficient $lambda$, our approach dynamically balances free exploration diversity and target domain style control by input conditions, thereby enhancing cross-domain generalization.
We provide the training procedure of \myformer in Algorithm~\ref{alg:training}.
Where $p_\textit{emb}$ and $\left(\overline{h}_m,\overline{h}_c,\overline{h}_b,\overline{h}_a\right)$ together serve as the condition variable $c$, 
$\tilde{Y}_{1:H}^{t-1}\in\mathbb{R}^{Bd\times P_H\times P_d}$ and $\hat{Y}_{1:H}^{t-1}\in\mathbb{R}^{Bd\times P_H\times P_d}$ represent the conditional and unconditional outputs of the denoising network, respectively. Furthermore, $Y_{1:H}^{t-1}$ as the final result, and $\lambda$ is a parameter that controls the proportion of conditional and unconditional generation in the final result ($\lambda \textit{=7.5}$ for our experiment)

\vspace{-10pt}
\subsection{Flexible-Length Sequence Forecasting}
The core capability of TimeControl lies in its flexibility to process observation sequences of arbitrary length and generate forecast sequences of any desired horizon.

First, \myformer processes the observation sequence $X_{-L+1:0}^0\in\mathbb{R}^{B\times d\times L}$ and forecast sequence $Y_{1:H}^0\in\mathbb{R}^{B\times d\times H}$ with the same channel-independence operation and patch-tokenizer operation to obtain $X_{emb}\in\mathbb{R}^{B\cdot d\times P_{L}\times P_{d}}$ and $Y_{emb}\in\mathbb{R}^{B\cdot d\times P_{H}\times P_{d}}$ where $P_d$ is the length of the patch, respectively, and we adopt a padding-free and overlap-free approach to perform the patch-tokenizer operation to ensure that $P_L=L/P_d$ and $P_H=H/P_d$. Specifically, we convert the observation sequence and forecast sequence into $P_L$ and $P_H$ tokens, respectively, by setting the hyperparameter $P_d$, where the dimension of the token is fixed to $P_d$.

Subsequently, the Condition Net accepts $X_{emb}\in\mathbb{R}^{B\cdot d\times P_{L}\times P_{d}}$ as input and outputs a set of multiscale condition variables $(h_m\in\mathbb{R}^{4\cdot P_L\times\frac{P_d}8},h_c\in\mathbb{R}^{2\cdot P_L\times\frac{P_d}4},h_b\in\mathbb{R}^{P_L\times\frac{P_d}2},h_a\in\mathbb{R}^{P_L\times P_d})$. Correspondingly, the Denoising Net accepts a set of multiscale condition variables $(\overline{h}_{m}\in\mathbb{R}^{4\cdot P_{H}\times\frac{P_{d}}{8}},\overline{h}_{c}\in\mathbb{R}^{2\cdot P_{H}\times\frac{P_{d}}{4}},\overline{h}_{b}\in\mathbb{R}^{P_{H}\times\frac{P_{d}}{2}},\overline{h}_{a}\in\mathbb{R}^{P_{H}\times P_{d}})$ as input and outputs $Y_{emb}\in\mathbb{R}^{B\cdot d\times P_{H}\times P_{d}}$, which is straightened directly as the prediction result $Y_{1:H}^{0}\in\mathbb{R}^{B\times d\times H}$.

The most innovative and critical design is that the innovative adapter module is designed to align the channel dimensions (the number of tokens $P_L$ and $P_H$) of the two tensors $(h_m,h_c,h_b,h_a)$ and $(\bar{h}_{m},\bar{h}_{c},\bar{h}_{b},\bar{h}_{a})$ in the observation space and the generation space. Specifically, the **1 $\times$ 1 Conv1D** in Adapter Block-a accepts $h_a\in\mathbb{R}^{P_L\times P_d}$ as input, keeping the feature dimension ($P_d$) unchanged, and converts the number of input channels from ($P_L$) to ($P_H$). This can be expressed as Equation $\overline{h}_a=AdapterBlock_a(h_a)$, where $h_a\in\mathbb{R}^{P_L\times P_d}$ and $\overline{h}_a\in\mathbb{R}^{P_H\times P_d}$.
The most critical challenge lies in allowing condition net and denoising net to be able to handle an arbitrary number of tokens (e.g. $P_L$ and $P_H$)? Indeed, this is achieved through the formula $X=(Conv1D(X^T))^T$, where $X^T\in\mathbb{R}^{P_d\times P_L}$ denotes the transpose of $X\in\mathbb{R}^{P_L\times P_d}$. Thus, it is only necessary to ensure that the number of input and output channels of Conv1D is fixed to $P_d$. On the other hand, attention is not sensitive to $P_L$ or $P_H$ when capturing global dependencies, and does not change the shape of the input tensor, so \myformer is able to accept any timesteps as an observation sequence, and can also output a prediction sequence with any timesteps.

\vspace{-10pt}
\subsection{Analysis of Computational Complexity}

Condition and denoising net adopts the flexible design that supports inputs and outputs of arbitrary timesteps, where the tensor is shaped as $(B,P_d,P_L)$ or $(B,P_d,P_H)$, where $B$ and $P_d$ denote the batch size and the dimensionality of the patch, and $P_L=\frac L{P_d}$ and $P_H=\frac H{P_d}$ denote the number of tokens contained in the observation and prediction sequences, respectively. 
Therefore, the convolutional complexity and the complexity of the attention mechanism are both $O(P_L\cdot(P_d)^2)=O(L\cdot P_d)$ in condition net, and similarly, $O(P_H\cdot(P_d)^2)=O(H\cdot P_d)$ in denoising net.

Adapter serves as a bridge between the observation and prediction space, which aligns the number of tokens ($P_L$ and $P_H$) between the condition net and the denoising net utilizing a special $1\times 1$ Conv1D through the transpose operation of the tensor. 
Thus the complexity of the attention mechanism are both $O(P_d\cdot(P_L)^2)=O(\frac {L^2}{P_d})$ in Adapter.


The computational complexity of \myformer is $O((\frac{(L^2+H^2)}{P_d}+k\cdot P_d\cdot(H+L))\cdot T)$, where $L$ and $H$ denote the length of the input and output sequences, $P_d$ is the dimension of the patch-wise tokenizer, $K$ denotes the depth of the encoder-decoder, and $T$ denotes the number of rounds of iterative denoising operation, respectively. 
In addition, we believe that the analysis on complexity comparison between \myformer and other diffuison models are important will help the reader to understand the comprehensive performance of the proposed model. 
Specifically, detailed theoretical comparisons and quantitative experiments can be found in Section~\ref{sec:computational} and Table~\ref{tab:computational}.

\input{tables_brief/benchmark_baseline.tex}

%% file: tables_brief/benchmark_baseline.tex
\begin{table*}[t!]
\centering
\vspace{-10pt}
\caption{
\textbf{Experimental Tasks.} We validate the proposed \myformer on four forecasting tasks.
To comprehensively evaluate the generalization of TimeControl across diverse time series scenarios, we designed four representative forecasting paradigms. 
First, we collected 28 datasets from a wide range of domains (including Nature, Transport, Energy, and Web) for joint pre-training.
We strictly divided the training set and the validation set, the pre-trained model was directly evaluated on target domains to assess its zero-shot performance as shown in (1). 
To verify that the representations learned in the integrated domain enhance the model on unseen datasets, we fine-tuned the pre-trained model on target domains before testing to evaluate its continual learning capability as indicated in (2). 
To validate the proposed method's ability to transfer learned knowledge to entirely new domains, we conducted cross-domain transfer learning studies among the popular ETT sub-datasets as described in (3). 
Finally, we compared the performance differences between TimeControl and existing generative models including Diffusion and GAN-based approaches as shown in (4). 
}
\vspace{-10pt}
\vskip 0.05in
\resizebox{0.95\linewidth}{!}{
\begin{tabular}{ccccccc}
\toprule
\textbf{Forecasting Scenarios} & \textbf{Descriptions} & \textbf{Baselines} & \textbf{Metrics} \\ 
\midrule

\begin{tabular}[c]{@{}c@{}}(1) Domain-fused pre-training \\ (zero-shot without fine-tuning)\end{tabular} & 
\begin{tabular}[c]{@{}c@{}}Model trained on multi-domain fusion dataset and \\ subsequently tested on each small-scale dataset\end{tabular} & 
\begin{tabular}[c]{@{}c@{}}Moirai~\cite{woo2024moirai}, UniTime~\cite{liu2024unitime}, GPT4TS~\cite{Zhou2023OneFA}\end{tabular} &
\begin{tabular}[c]{@{}c@{}}MSE, MAE\end{tabular} \\
\midrule

\begin{tabular}[c]{@{}c@{}}(2) Domain-fused pre-training \\ (fine-tuning on target dataset)\end{tabular} & 
\begin{tabular}[c]{@{}c@{}}Model trained on multi-domain fusion dataset and \\ fine-tuned on target dataset \\ subsequently tested on each small-scale dataset\end{tabular} & 
\begin{tabular}[c]{@{}c@{}}TimeLLM~\cite{Jin2023TimeLLMTS}, LLM4TS~\cite{Chang2023LLM4TSAP}, GPT4TS~\cite{Zhou2023OneFA} \\ DUET~\cite{qiu2025duet}, PDF~\cite{dai2024pdf}, Pathformer~\cite{chen2024pathformer}, TimeMixer~\cite{wang2024timemixer}\end{tabular} &
\begin{tabular}[c]{@{}c@{}}MSE, MAE\end{tabular} \\
\midrule

\begin{tabular}[c]{@{}c@{}}(3) Domain-transfer learning \\ (zero-shot)\end{tabular} & 
\begin{tabular}[c]{@{}c@{}}Model trained on dataset-A and \\ subsequently tested on other dataset-B\end{tabular} & 
\begin{tabular}[c]{@{}c@{}}TimeVLM~\cite{zhong2025timevlm}, TimeLLM~\cite{Jin2023TimeLLMTS}, LLMTime~\cite{llmtime}, GPT4TS~\cite{Zhou2023OneFA}, \\ DUET~\cite{qiu2025duet}, PDF~\cite{dai2024pdf}, Pathformer~\cite{chen2024pathformer}, \\ TimeMixer~\cite{wang2024timemixer}, PatchTST~\cite{Nie2023PatchTST}, DLinear~\cite{zeng2023dlinear}\end{tabular} &
\begin{tabular}[c]{@{}c@{}}MSE, MAE\end{tabular} \\
\midrule

\begin{tabular}[c]{@{}c@{}}(4) Time series generation \\ (through Diffusion/GAN)\end{tabular} & 
\begin{tabular}[c]{@{}c@{}}Generative model trained on each small-scale dataset \\ and demonstratd with repeated sampling on test set\end{tabular} & 
\begin{tabular}[c]{@{}c@{}}Diffusion-TS~\cite{Yuan2024DiffusionTSID}, LDT~\cite{li2024ldt}, mrDiffusion~\cite{shen2024mrdiffusion}, \\ TimeDiff~\cite{shen2024timediff}, CSDI~\cite{Tashiro2022csdi}, SSSD~\cite{alcaraz2022sssd}, TimeGAN~\cite{yoon2021timegan}, \\ TimeVAE~\cite{desai2023timevae}, TimeGrad~\cite{Rasul2021timegrad}, Cot-GAN~\cite{xu2022cotgan}\end{tabular} &
\begin{tabular}[c]{@{}c@{}}Context-FID \\ Correlational Score \\ Discriminative Score \\ Predictive Score\end{tabular} \\

\toprule
\end{tabular}}
\label{tab:experimental-benchmark}
\vspace{-10pt}
\end{table*}

%% file: segments/5-Experiment.tex
\section{Experiments}
For different forecasting scenarios, we design four task paradigms, which include (1) domain-fused pre-training (zero-shot testing on the target domain without any fine-tuning), (2) domain-fused pre-training (fine-tuning on the target dataset, and then testing), (3) domain-transfer learning and (4) time series generation (through Diffusion/GAN). 
To evaluate the performance of the proposed method, we extensively experiment with 49 datasets aggregated from multiple sources, including the Monash Time Series Forecasting Repository~\cite{godahewa2021monash}, M-competitions, public-domain datasets from Kaggle and TFB datasets~\cite{qiu2024tfb} for comprehensive training and evaluation of \myformer. 
All utilized 30 baselines include the following three components: (1) time series foundation models (pre-trained on text, visual and time series data, separately), (2) time series proprietary models (unique model architectures for each dataset are designed and trained from scratch) and (3) time series generation models (trained to progressively reconstruct time series from noise).
Details of baselines utilized in four task paradigms are presented in Table~\ref{tab:experimental-benchmark},


\vspace{-10pt}
\subsection{Implementation Details}
To ensure consistency with existing studies~\cite{qiu2025duet,woo2024moirai}, we set four prediction lengths for the ILI dataset: 24, 36, 48, 60; for the other nine datasets, we adopt four prediction lengths: 96, 192, 336, 720;
To keep consistent with previous works~\cite{qiu2025duet,li2024foundts}, we repeatedly conduct experiments across multiple horizons for all baselines and report the best results across all horizons. For ILI, the optional horizons are 36, 104, For other datasets, the optional horizons are 96, 336, 512. 
Using the TFB codebase~\cite{qiu2024tfb}, we evaluate \myformer three times and report the averaged results, while the baseline results are sourced from the DUET~\cite{qiu2025duet} papers.
We strictly adhere to the configuration of FoundTS~\cite{li2024foundts} and refrain from the ``Drop Last'' trick to ensure result fairness. 
All experiments are conducted via Python 3.10 and PyTorch~\cite{paszke2019pytorch} 2.6.0 on NVIDIA V100 GPUs. We utilize L2 loss function and Adam optimizer, with the batch size from 16 to 64.

In the specific implementation of \myformer, the proposed model includes three components. 
condition net and denoising net are composed of the same structure, which contains $L$ encoder blocks, $L$ decoder blocks and $1$ middle block. 
We allow consecutive $L$-layer block residues to be stacked together and used to form the condition net and denoising net. 
Furthermore, the number of input and output channels of the middle block is denoted as $D$, and accordingly the number of input and output channels in the three pairs of encoder-decoder blocks are $D/4$, $D/2$, and $D$, respectively. 
By default, the hyperparameter of \myformer is $L=3,D=256$, and all experimental results presented follow this setting. 
We adopted the setting of DDIM, where backward's iteration steps is set to $50$ for the inference process and $200$ for the training process.


\input{tables_brief/forecasting_full.tex}

\input{tables_brief/forecasting_zero.tex}
\vspace{-10pt}
\subsection{Full-Shot and Zero-Shot Time Series Forecasting}
\vspace{-5pt}
\textbf{Setup.}
To validate that the proposed TimeControl possesses cross-domain generalization capability, transferring knowledge from the multi-domain integrated datasets to entirely unseen domain, we designed the following experiments. 
Specifically, we first pre-trained the model on a mixture of 39 datasets from Monash for learning pivotal information from multiple domains, and subsequently evaluated the model's performance on 10 benchmarks from the TFB. 
To test the most challenging zero-shot capability, we utilized the pre-trained TimeControl for zero-shot time series forecasting across a wide range of downstream domains after freezing the parameters, evaluating its predictive performance on previously unseen domains. 
Furthermore, we designed experiments to evaluate the effectiveness of the adapter-based fine-tuning strategy. 
For each benchmark in the TFB protocol, we fine-tuned TimeControl using its training set and subsequently evaluated the model's performance on the validation set.

\textbf{Results.}
The first column of Table \ref{tab:forecasting_full} shows the complete zero-shot results of cross-domain pre-training, which validates the ability of the proposed method to model multi-domain probability distributions. 
Compared to SOTA time series foundation models, \myformer achieves an overall better performance than them. Specifically, the average MSE of the proposed \myformer is reduced by \textbf{14.2\%}, \textbf{20.1\%} and \textbf{27.6\%} compared to the existing Moirai, UniTime and GPT4TS, respectively, which demonstrates the potential of \myformer as the unified temporal spreading model.

The second column of Table \ref{tab:forecasting_full} demonstrates the results of the domain-fused model after fine-tuning on target dataset based on the adapter, the average MSE is reduced by \textbf{17.9\%}, \textbf{18.6\%} and \textbf{22.4\%} compared to the existing TimeLLM, LLM4TS and GPT4TS, which indicates that the effectiveness of the proposed fine-tuning strategy
The third column of Table \ref{tab:forecasting_full} demonstrates the overall performance of the existing proprietary method which trained from scratch on target dataset. Overall, the fine-tuned \myformer achieves comparable performance to SOTA proprietary model DUET, and excitingly, the fine-tuned \myformer achieves a performance that exceeds the performance of existing foundation and proprietary model on most datasets.
The experimental results illustrate that the adapter-based finetuning strategy fully utilise the potential representations learned from multiple data domains during the pre-training phase. Our finetuning strategy effectively activates performance of pre-trained models in downstream tasks, which provides implications for future work. 


\vspace{-10pt}
\subsection{Domain-Transfer Time Series Forecasting}
\vspace{-5pt}
\textbf{Setup.}
To validate the proposed TimeControl's capability of transferring knowledge from a source domain to a completely unseen target domain, we designed domain-transfer forecasting experiments. 
Specifically, the model was first trained on a given source dataset and subsequently evaluated on a different target dataset \textbf{without any fine-tuning}, assessing its zero-shot generalization under significant distribution shifts. 
We constructed multiple transfer tasks, such as ETT$_{h1\to m2}$, to comprehensively evaluate the cross-domain robustness. 
The proposed method was compared against two groups of baselines: (1) time series foundation models pre-trained on multi-modality data (e.g., TimeVLM, TimeLLM and GPT4TS), which were also continue trained on the source domain, and (2) time series proprietary models (e.g., DLinear, PatchTST and TimesNet) trained from scratch on the source domain. Performance was measured using MSE and MAE across forecasting lengths of 96, 192, 336 and 720.

\input{tables_brief/generation_brief.tex}
\textbf{Results.}
The complete experimental results for the zero-shot domain-transfer forecasting task are presented in Table~\ref{tab:brief_forecasting_zero}. 
An encouraging finding is that TimeControl demonstrates superior generalization ability in these challenging transfer scenarios. 
Specifically, TimeControl achieves the best overall performance, leading in 68 out of 80 comparative scenarios, significantly outperforming all foundation and proprietary models. 
For instance, in the ETT$_{h1\to h2}$ task, TimeControl attains an average MSE of 0.317, which is lower than TimeVLM (0.338) and TimeLLM (0.353). 
This remarkable performance can be attributed to two key designs: Firstly, the multi-scale condition-denoising mechanism effectively guide the generation of predictions for the unseen target domain. 
Secondly, the generation-style guidance ensures that the predictions align with the target domain's style while maintaining diversity and accuracy. 
These results conclusively verify that TimeControl successfully establishes a shared representation space through its diffusion-based architecture, enabling robust and high-quality cross-domain extrapolation without requiring access to the target domain's data during training.

\begin{figure*}
\begin{center}
\vspace{-10pt}
\centerline{\includegraphics[width=1.8\columnwidth]{images/show_ablation_ETTm.pdf}}
\vspace{-10pt}
\caption{
Visualisation on the validity of proposed condition net. Where the upper part shows the visualisations on ETTm1 and ETTm2 datasets.
}\label{fig:vis_brief_4}
\vspace{-10pt}
\end{center}
\end{figure*}

\begin{figure*}
\begin{center}
\vspace{-10pt}
\centerline{\includegraphics[width=1.9\columnwidth]{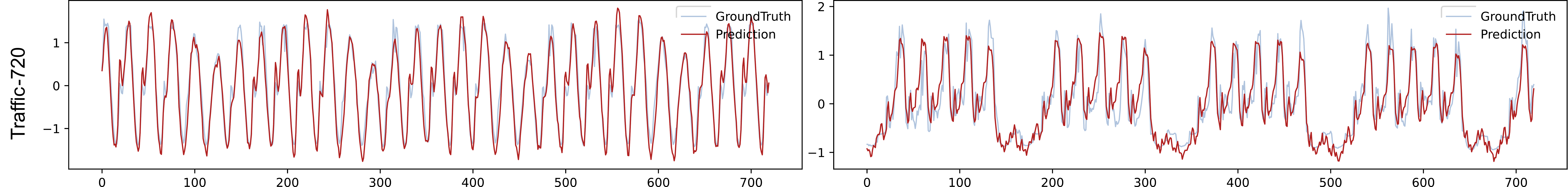}}
\centerline{\includegraphics[width=1.9\columnwidth]{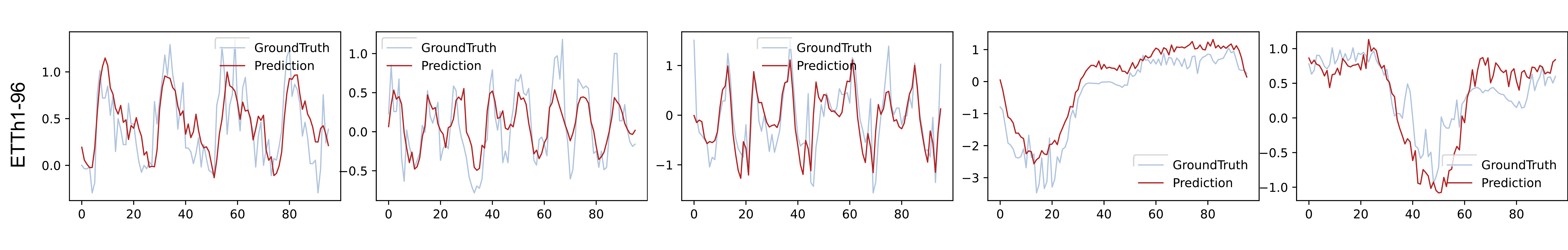}}
\vspace{-15pt}
\caption{
Demonstration of \myformer prediction results on real \textbf{long-term multi-periodic} sequences sampled from Traffic dataset.
Demonstration of \myformer prediction results on real \textbf{short-term non-periodic} sequences sampled from ETT dataset. The blue and red lines respectively represent the actual values and predicted values of the future time series.
}\label{fig:vis_brief_3}
\end{center}
\vspace{-10pt}
\end{figure*}


\input{tables_brief/solar_mujoco.tex}
\input{tables_brief/forecasting_tmdm.tex}
\vspace{-10pt}
\subsection{Time Series Generation}
\textbf{Setup.}
To comprehensively evaluate the generative capabilities of the proposed TimeControl, we conducted extensive experiments on the time series generation task. Following the evaluation protocol established in Diffusion-TS~\cite{Yuan2024DiffusionTSID}, we employed 7 key metrics to assess different aspects of generation quality: 
(1) Context-FID Score measures the distributional similarity between generated and real sequences using time-series-specific representations from TS2Vec; 
(2) Correlational Score evaluates the preservation of inter-variable relationships in multivariate time series; 
(3) Discriminative Score assesses how distinguishable generated sequences are from real ones using a GRU-based classifier; 
(4) Predictive Score measures the forecasting performance achieved by models trained on synthetic data; 
(5) QICE; 
(6) CRPS; and 
(7) PICP.
We compared TimeControl against state-of-the-art generative baselines including diffusion-based methods (Diffusion-TS, TimeGrad, TimeDiff) and GAN-based approaches (TimeGAN, Cot-GAN, TimeVAE). 
Experiments were conducted on datasets including ETTh, Electricity, Traffic and Energy, with sequence lengths of 64, 128, and 256 to evaluate scalability.

\textbf{Results.}
A comparison of the performance of \myformer and other diffusion baselines on the time series generation task is shown in the Table \ref{tab:generation_brief}. 
Specifically, the generation metrics Context-FID Score, Correlational Score, Discriminative Score and Predictive Score is reduced by \textbf{18.2\%}, \textbf{16.2\%}, \textbf{18.5\%} and \textbf{17.8\%} on ETTh dataset compared to the existing Diffusion-TS. 
Besides, compared with Diffusion-TS, the average improvement of all indicators of \myformer on ETTh and Energy datasets is \textbf{17.6\%} and \textbf{8.2\%}, respectively.
Furthermore, as shown in Table~\ref{tab:solar_mujoco}, TimeControl excels in imputation task, which consistently outperforms all baselines, and achieves an improvement of up to \textbf{24.5\%} in challenging imputation scenarios with 70\%, 80\%, and 90\% masking ratios. 
These significant improvements can be attributed to the synergistic effect of our multi-scale condition-denoising architecture and generation-style guidance mechanism, which enable the model to capture comprehensive temporal patterns at different resolutions while maintaining precise control over the generative process. 
The results in Table~\ref{tab:brief_forecasting_tmdm} validate that TimeControl can generate high-quality, diverse time series samples that closely match the distributional characteristics of real data across various domains and task scenarios.
Additionally, following the TMDM~\cite{li2024ldt} design, we evaluated TimeControl's performance on three popular datasets using three time-series probability prediction metrics (QICE, PICP, and CRPS), and tested it alongside several competitive benchmark models. 


\input{tables_brief/ablation_brief.tex}
\vspace{-10pt}
\subsection{Ablation Study}
\subsubsection{Quantitive Analysis of Proposed Component's Effectiveness}
\textbf{Setup.}
To ascertain the impact of designs within TimeControl, we perform ablations focusing on following components:
\noindent
(1) \textit{w/o ConditionNet:} Indicating the use of a simple single-layer MLP to replace the ConditionNet.
\noindent
(2) \textit{w/o Adapter}: Removing the apdater module. The output of the condition net is directly fed into the denoising net.
\noindent
(3) \textit{w/o Generation-style}: Utilizing the denoising results of the condition variables as the prediction sequence.

\textbf{Results.}
Table~\ref{tab:ablation_brief} illustrates the influence of each design. We have the following observations: 
(1) When the condition net is removed, the prototype waves showed severe degradation, which limited the model to extract key representations from long horizons. Especially on the dataset with significant temporal distribution changes, performance dropped sharply by 27.8\%. 
These results may come from that the proposed condition network will improve the the generalization capability of \myformer to learn multi-scale representations.
(2) When the adapter module was removed, the model showed a performance drop of more than 14.1\% and 10.9\% on Electricity and Weather dataset with strong channel correlations. This verifies that. the design of the adapter effectively aligns the observation space with the prediction space. 
(3) When using standard generation paradigm, the performance degraded significantly (from 13.6\% to 19.8\%) across all datasets, which confirms the importance of TimeControl in summarizing prediction-sensitive patterns from longer horizons.


\subsubsection{Qualitative Visualization of Proposed Component's Effectiveness}
To validate the effectiveness of the proposed model architecture, we visualised experiments against condition net ablation, as shown in Figure~\ref{fig:vis_brief_4}. For each dataset, the generative capacity of the full \myformer model is shown on the left, and the generative capacity of the model when the condition net structure is excluded is shown on the right, where the true values are in blue and the generated samples are in yellow. For each experiment, the model demonstrated the best generative performance when the generated samples and the true values completely overlapped together.


\begin{figure*}[ht]
\begin{center}
\vspace{-10pt}
\centerline{\includegraphics[width=1.8\columnwidth]{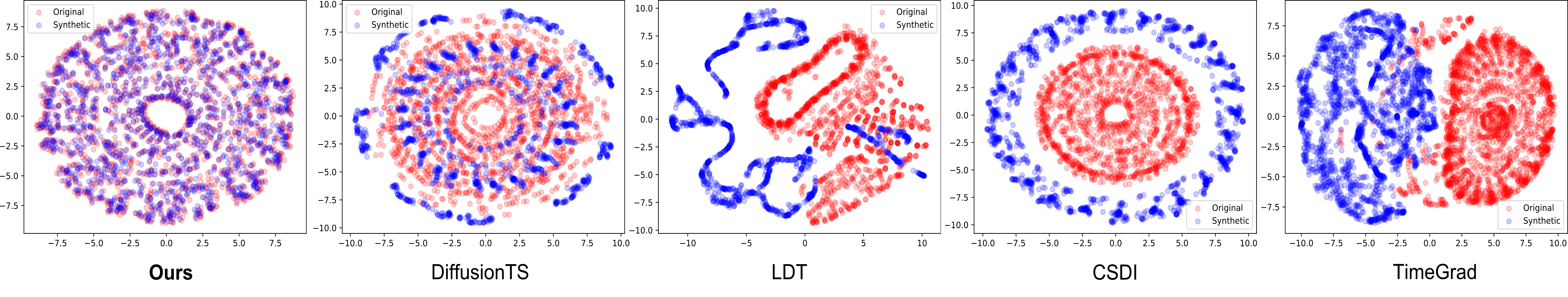}}
\centerline{\includegraphics[width=1.8\columnwidth]{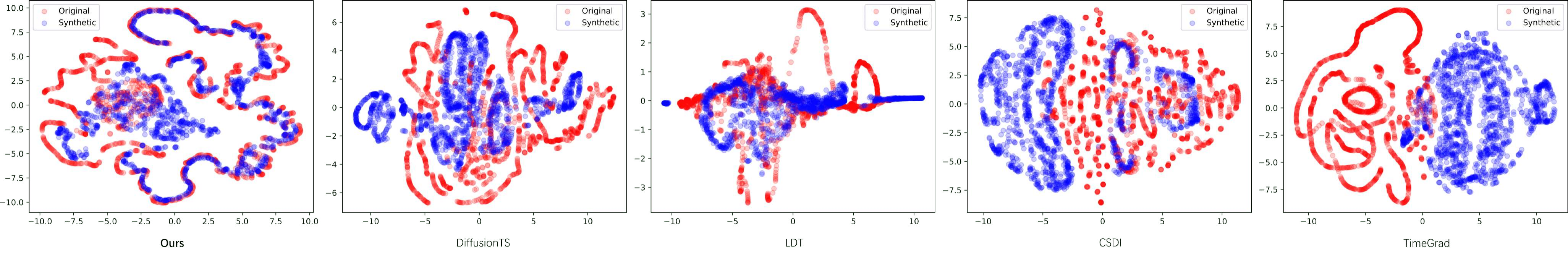}}
\vspace{-10pt}
\caption{
Visualization of comparisons between \myformer and baselines on ETTh1 dataset.
}\label{fig:vis_brief_2}
\vspace{-25pt}
\end{center}
\end{figure*}


\input{tables_brief/complexity.tex}
\vspace{-10pt}
\subsection{Complexity Comparison of TimeControl and Other Models}

\textbf{Computational Complexity.}\label{sec:computational}
The existing time series diffusion model (DiffusionTS) uses the stepwise iterative noise reduction algorithm DDPM (timesteps=1000), while we use the de-Markovised DDIM algorithm, which means that \myformer only requires a smaller number of iterations (timesteps=50) in the inference phases. In addition, traditional diffusion models first require a autoencoder before the noise reduction process can be iteratively trained in the hidden space. TimeControl does not require any prior stage, which greatly reduces the overall training overhead of \myformer. 
Table~\ref{tab:computational} presents the comparison of the total parameters, memory consumption, and inference speed between the proposed TimeControl model and existing baseline models.
Compared to the baseline model TimeLLM, the proposed TimeControl achieves 71.9\% and 48.3\% optimizations in memory overhead and inference speed, respectively. Compared to the specialized models CSDI and DiffusionTS, TimeControl incurs less than a 6\% increase in inference overhead.

\textbf{Advantages of Adapter-based Fine-tuning Strategy.}
An encouraging news is that \myformer has made great efforts to reduce the time overhead during training and inference. Firstly, the pre-train phase occurs entirely offline, and after cross-dataset pre-trained, the fine-tuned model weights can quickly converge to the desired state. In addition, \myformer's hybrid architecture supports naturally efficient fine-tuning through the ‘plug-and-play’ Adapter.


\input{tables_brief/sensitivity.tex}
\vspace{-10pt}
\subsection{Model Hyperparameter Analysis}
In order to verify the sensitivity of \myformer to hyperparameter selection, the Table~\ref{tab:exp_sensitivity} shows the performance of \myformer with different parameter scales on multiple benchmarks. We define the conditional net, denoising net and adapter as being composed of $L$-layer block residues to be stacked. The input and output channel numbers of the intermediate block are denoted as $D$. Correspondingly, the output and output channel numbers of three encoder-decoder blocks are D/4, D/2 and D respectively. The model parameter combinations include $(L,D)$ = $(2,128)~(4,256)$, with the metric of MSE.
For different hyperparameter combinations, the average performance fluctuation of the model on the same metric is below 10\%. For instance, on the ETTh1 and Weather datasets, the performance fluctuation is only 8.5\% and 7.9\%, respectively.


\vspace{-10pt}
\subsection{Visualization} 

\begin{figure*}
\begin{center}
\vspace{-20pt}
\centerline{\includegraphics[width=1.9\columnwidth]{images/show_truth_predition_tsne_ecl_1.pdf}}
\vspace{-10pt}
\caption{
Visualization of comparisons between \myformer and baselines on the Electricity dataset. The proposed TimeControl achieves the best visualization.
}\label{fig:vis_brief_1}
\vspace{-25pt}
\end{center}
\end{figure*}

\subsubsection{Comparison of Models in Time Series Forecasting}
We plotted the prediction intervals for the probabilistic baseline (50 repetitive samples for each model) as shown in the upper left of Figure \ref{fig:vis_brief_1}, where the light and dark green colors represent the prediction results for the 10-90\% and 25-75\% confidence intervals, and the blue and green curves indicate the ground truth and median prediction results, respectively. 
In addition, on the right part of Figure \ref{fig:vis_brief_1}, we fix the same observation sequence and make each probabilistic model perform inference repeatedly for 50 trials, then project all prediction results into a two-dimensional space through t-SNE, where the cross dashed intersection marks the ground truth. 
The bottom left of Figure \ref{fig:vis_brief_1} demonstrates the prediction results of the deep regression model (\myformer samples only once), where the blue and red curves indicate the ground truth and prediction results. 


\subsubsection{Comparison of Models in Time Series Generation}
To further demonstrate the distributions of the generated and truth series, Figure \ref{fig:vis_brief_2} shows the results of comparing several probabilistic baselines with \myformer. To validate the generative power of the diffusion-based probabilistic model, we visualise the generation results of \myformer and the pre-existing diffusion method on the same dataset using the t-SNE method. The red colour represents the real sequence samples and the blue colour represents the generated dataset, where the degree of aggregation of these two samples in two-dimensional space reflects the generative ability of the model. Specifically, when the projections of the two samples in 2D space are fully aggregated, the model exhibits excellent generative performance. The results show that \myformer learns comprehensive characterization and generates sequences that are more consistent with the actual distribution.


\subsubsection{Long-Term Multi-Periodic and Short-Term Non-Periodic Forecasting}
Regarding long-term multi-periodic series and short-term non-periodic series have been a great challenge for time series forecasting. 
Therefore it is considered important to introduce more visualisation results as shown in Figures \ref{fig:vis_brief_3}, which illustrates the real long-periodic series \textbf{(timesteps equals to 720)} sampled from ETT, ECL and Traffic datasets. 
And Figure \ref{fig:vis_brief_3} illustrates the real short-term non-periodic series \textbf{(timesteps equals to 96)} sampled from ETT, ECL and Traffic datasets. Specifically, \myformer demonstrates excellent performance in long-term series forecasting, which verifies that \myformer has the ability to capture long-term dependencies, which is crucial for practical applications. In addition, \myformer likewise exhibits satisfactory prediction results in short-term sequences, which demonstrates the ability of the diffusion-based forecasting model to generate high-quality time-series samples.

%% file: tables_brief/forecasting_full.tex
\begin{table*}[htbp]
\centering
\small
\vspace{-10pt}
\caption{
Comparison of the performance from diverse prediction lengths on \textbf{full-shot and zero-shot time series forecasting}.
Specifically, the left side of double-bar graph shows the zero-shot performance comparison between TimeControl and existing foundation models. The right side of double-bar graph shows the full-shot performance comparison between TimeControl and existing proprietary models. We boldface the best performance on two scenarios, respectively. Among them, the TimeControl demonstrated the best performance in most of the prediction scenarios.
}\label{tab:forecasting_full}
\vspace{-10pt}
\vskip 0.05in
\tabcolsep=0.20cm
\renewcommand\arraystretch{1.3}
\resizebox{0.95\textwidth}{!}{
\begin{tabular}{ccccccccc|cccccccc|cccccccc}
\toprule
\hline
\multicolumn{1}{c}{\multirow{5}{*}{Model}} & 
\multicolumn{8}{c|}{\textbf{\textit{Domain-fused pre-training}}} & 
\multicolumn{8}{c|}{\textbf{\textit{Domain-fused pre-training}}} & 
\multicolumn{8}{c}{\textbf{\textit{Models Trained From Scratch}}} \\ 

\multicolumn{1}{c}{} & 
\multicolumn{8}{c|}{\textbf{\textit{(zero-shot without fine-tuning)}}} & 
\multicolumn{8}{c|}{\textbf{\textit{(fine-tuning on target dataset)}}} & 
\multicolumn{8}{c}{\textbf{\textit{(full-shot)}}} \\ 

\cline{2-25} 

\multicolumn{1}{c}{} & 
\multicolumn{2}{c}{\textbf{\myformer}} & 
\multicolumn{2}{c}{Moirai} & 
\multicolumn{2}{c}{UniTime} & 
\multicolumn{2}{c|}{GPT4TS} & 
\multicolumn{2}{c}{\textbf{\myformer}} & 
\multicolumn{2}{c}{TimeLLM} & 
\multicolumn{2}{c}{LLM4TS} & 
\multicolumn{2}{c|}{GPT4TS} & 
\multicolumn{2}{c}{DUET} & 
\multicolumn{2}{c}{PDF} & 
\multicolumn{2}{c}{Pathformer} & 
\multicolumn{2}{c}{TimeMixer} \\ 

\multicolumn{1}{c}{} & 
\multicolumn{2}{c}{\textbf{(Ours)}} & 
\multicolumn{2}{c}{\cite{woo2024moirai}} & 
\multicolumn{2}{c}{\cite{liu2024unitime}} & 
\multicolumn{2}{c|}{\cite{Zhou2023OneFA}} & 
\multicolumn{2}{c}{\textbf{(Ours)}} & 
\multicolumn{2}{c}{\cite{Jin2023TimeLLMTS}} & 
\multicolumn{2}{c}{\cite{Chang2023LLM4TSAP}} & 
\multicolumn{2}{c|}{\cite{Zhou2023OneFA}} & 
\multicolumn{2}{c}{\cite{qiu2025duet}} & 
\multicolumn{2}{c}{\cite{dai2024pdf}} & 
\multicolumn{2}{c}{\cite{chen2024pathformer}} & 
\multicolumn{2}{c}{\cite{wang2024timemixer}} \\
    
    \cline{2-25} 
    
    \multicolumn{1}{c}{} & 
    MSE & MAE & MSE & MAE & MSE & MAE & MSE & MAE & 
    MSE & MAE & MSE & MAE & MSE & MAE & MSE & MAE & MSE & MAE & MSE & MAE & MSE & MAE & MSE & MAE \\
    
    \hline 
    
    \multicolumn{1}{c}{\multirow{1}{*}{\rotatebox{0}{ETTm1}}} & 
    \textbf{0.341} & \textbf{0.369} &
    0.448 & 0.410 &
    0.385 & 0.399 &
    0.551 & 0.483 &
    \textbf{0.288} & \textbf{0.336} &
    0.329 & 0.372 &
    0.343 & 0.378 &
    0.352 & 0.383 &
    0.338 & 0.369 &
    0.342 & 0.376 &
    0.357 & 0.374 &
    0.355 & 0.380 \\
    
    \multicolumn{1}{c}{\multirow{1}{*}{\rotatebox{0}{ETTm2}}} & 
    \textbf{0.263} & \textbf{0.329} &
    0.300 & 0.341 &
    0.293 & 0.334 &
    0.321 & 0.356 &
    \textbf{0.224} & \textbf{0.303} &
    0.251 & 0.313 &
    0.251 & 0.313 &
    0.267 & 0.326 &
    0.247 & 0.307 &
    0.250 & 0.313 &
    0.253 & 0.308 &
    0.257 & 0.318 \\
    
    \multicolumn{1}{c}{\multirow{1}{*}{\rotatebox{0}{ETTh1}}} & 
    \textbf{0.388} & \textbf{0.405} &
    0.399 & 0.424 &
    0.442 & 0.448 &
    0.502 & 0.461 &
    \textbf{0.334} & \textbf{0.383} &
    0.408 & 0.423 &
    0.404 & 0.418 &
    0.428 & 0.426 &
    0.398 & 0.418 &
    0.406 & 0.425 &
    0.417 & 0.426 &
    0.427 & 0.441 \\
    
    \multicolumn{1}{c}{\multirow{1}{*}{\rotatebox{0}{ETTh2}}} & 
    0.379 & 0.405 &
    \textbf{0.341} & \textbf{0.379} &
    0.378 & 0.403 &
    0.386 & 0.406 &
    \textbf{0.285} & \textbf{0.358} &
    0.334 & 0.383 &
    0.331 & 0.383 &
    0.355 & 0.395 &
    0.334 & 0.383 &
    0.347 & 0.391 &
    0.360 & 0.395 &
    0.349 & 0.397 \\
    
    \multicolumn{1}{c}{\multirow{1}{*}{\rotatebox{0}{Electricity}}} & 
    \textbf{0.183} & \textbf{0.289} &
    0.233 & 0.320 &
    0.216 & 0.305 &
    0.251 & 0.338 &
    \textbf{0.149} & \textbf{0.244} &
    0.158 & 0.252 &
    0.159 & 0.253 &
    0.167 & 0.263 &
    0.157 & 0.247 &
    0.160 & 0.253 &
    0.168 & 0.261 &
    0.184 & 0.284 \\
    
    \multicolumn{1}{c}{\multirow{1}{*}{\rotatebox{0}{Traffic}}} & 
    \textbf{0.342} & \textbf{0.239} &
    0.384 & 0.279 &
    0.356 & 0.266 &
    0.414 & 0.295 &
    \textbf{0.301} & \textbf{0.213} &
    0.388 & 0.264 &
    0.401 & 0.273 &
    0.414 & 0.295 &
    0.393 & 0.256 &
    0.395 & 0.270 &
    0.416 & 0.264 &
    0.409 & 0.279 \\
    
    \multicolumn{1}{c}{\multirow{1}{*}{\rotatebox{0}{Weather}}} & 
    \textbf{0.230} & \textbf{0.263} &
    0.242 & 0.267 &
    0.253 & 0.276 &
    0.293 & 0.309 &
    \textbf{0.202} & \textbf{0.246} &
    0.225 & 0.257 &
    0.223 & 0.260 &
    0.237 & 0.271 &
    0.218 & 0.252 &
    0.227 & 0.263 &
    0.225 & 0.258 &
    0.226 & 0.264 \\
    
    \multicolumn{1}{c}{\multirow{1}{*}{\rotatebox{0}{Exchange}}} & 
    0.391 & \textbf{0.397} &
    0.420 & 0.453 &
    \textbf{0.364} & 0.404 &
    0.421 & 0.446 &
    0.308 & \textbf{0.345} &
    0.381 & 0.429 &
    0.402 & 0.426 &
    0.423 & 0.464 &
    \textbf{0.280} & 0.364 &
    0.350 & 0.397 &
    0.384 & 0.414 &
    0.381 & 0.416 \\
    
    \multicolumn{1}{c}{\multirow{1}{*}{\rotatebox{0}{ILI}}} & 
    \textbf{1.872} & \textbf{0.883} &
    2.055 & 0.907 &
    2.137 & 0.929 &
    3.678 & 1.372 &
    1.742 & 0.854 &
    \textbf{1.435} & 0.801 &
    1.757 & 0.863 &
    1.925 & 0.903 &
    1.616 & \textbf{0.795} &
    1.808 & 0.898 &
    1.995 & 0.909 &
    1.820 & 0.886 \\
    
    \multicolumn{1}{c}{\multirow{1}{*}{\rotatebox{0}{Solar}}} & 
    \textbf{0.173} & \textbf{0.233} &
    0.180 & 0.242 &
    0.250 & 0.243 &
    0.315 & 0.294 &
    \textbf{0.164} & \textbf{0.208} &
    0.193 & 0.260 &
    0.202 & 0.232 &
    0.205 & 0.224 &
    0.189 & 0.241 &
    0.200 & 0.263 &
    0.204 & 0.228 &
    0.193 & 0.252 \\
    
    \hline
    
    \multicolumn{1}{c}{$1^{\text{st}}$ Count} & 
    \multicolumn{2}{c}{\textbf{17}} & 
    \multicolumn{2}{c}{2} & 
    \multicolumn{2}{c}{1} & 
    \multicolumn{2}{c|}{0} & 
    \multicolumn{2}{c}{\textbf{17}} & 
    \multicolumn{2}{c}{1} & 
    \multicolumn{2}{c}{0} & 
    \multicolumn{2}{c|}{0} & 
    \multicolumn{2}{c}{2} & 
    \multicolumn{2}{c}{0} & 
    \multicolumn{2}{c}{0} & 
    \multicolumn{2}{c}{0} \\
    \hline
    \bottomrule
    \end{tabular}
    }
\vspace{-10pt}
\end{table*}

%% file: tables_brief/forecasting_zero.tex
\begin{table*}[htpb]
\caption{
Comparison of the performance on \textbf{domain-transfer forecasting} task. 
We utilize the notation ``source\small{$\rightarrow$}target'' to represent the scenario of domain transfer. Specifically, the left side of the double vertical line illustrates performance comparisons between TimeControl and existing foundational models. These methods are first pre-trained on textual (e.g., TimeLLM, LLMTime, GPT4TS), visual (TimeVLM), or time-series (TimeControl) domains, followed by continuing to train on the source domain, and ultimately evaluated on the target domain. The right side of the double vertical line demonstrates performance comparisons of existing proprietary models, which are trained from scratch on the source domain and subsequently evaluated on the target domain.
}\label{tab:brief_forecasting_zero}
\vspace{-10pt}
\vskip 0.05in
\centering
\resizebox{1.9\columnwidth}{!}{
\begin{small}
\renewcommand{\multirowsetup}{\centering}
\tabcolsep=0.20cm
\renewcommand\arraystretch{1.3}
\begin{tabular}{ccccccccccc|cccccccccccc}
\toprule
\hline

\multicolumn{1}{c}{\multirow{5}{*}{Model}} & 
\multicolumn{10}{c|}{\textbf{\textit{Time Series Foundation Models}}} & 
\multicolumn{12}{c}{\textbf{\textit{Time Series Proprietary Models}}} \\ 

\multicolumn{1}{c}{} & 
\multicolumn{10}{c|}{\textbf{\textit{(Pre-trained on multi-domain, continue trained on the source domain)}}} & 
\multicolumn{12}{c}{\textbf{\textit{(Trained from scratch on the source domain)}}} \\ 

\cline{2-23} 

 &
\multicolumn{2}{c}{\textbf{\myformer}} & 
\multicolumn{2}{c}{TimeVLM} &
\multicolumn{2}{c}{TimeLLM} &
\multicolumn{2}{c}{LLMTime} &
\multicolumn{2}{c|}{GPT4TS} &
\multicolumn{2}{c}{DUET} &
\multicolumn{2}{c}{PDF} &
\multicolumn{2}{c}{Pathformer} &
\multicolumn{2}{c}{TimeMixer} &
\multicolumn{2}{c}{PatchTST} &
\multicolumn{2}{c}{DLinear} \\

&
\multicolumn{2}{c}{\textbf{(Ours))}} & 
\multicolumn{2}{c}{\cite{zhong2025timevlm}} &
\multicolumn{2}{c}{\cite{Jin2023TimeLLMTS}} &
\multicolumn{2}{c}{\cite{llmtime}} &
\multicolumn{2}{c|}{\cite{Zhou2023OneFA}} &
\multicolumn{2}{c}{\cite{qiu2025duet}} &
\multicolumn{2}{c}{\cite{dai2024pdf}} &
\multicolumn{2}{c}{\cite{chen2024pathformer}} &
\multicolumn{2}{c}{\cite{wang2024timemixer}} &
\multicolumn{2}{c}{\cite{Nie2023PatchTST}} &
\multicolumn{2}{c}{\cite{zeng2023dlinear}} \\

\cline{2-23} &
MSE & MAE & MSE & MAE & MSE & MAE & MSE & MAE & MSE & MAE & 
MSE & MAE & MSE & MAE & MSE & MAE & MSE & MAE & MSE & MAE & MSE & MAE \\

\hline
\multirow{1}{*}{ETT$_{h1\to h2}$}
& \textbf{0.317} & \textbf{0.360} & 0.338 & 0.385 & 0.353 & 0.387 & 0.992 & 0.708 & 0.406 & 0.422 & 0.493 & 0.488 & 0.380 & 0.405 & 0.421 & 0.431 & 0.582 & 0.552 & 0.582 & 0.548 & 0.866 & 0.741 \\

\multirow{1}{*}{ETT$_{h1\to m2}$}
& 0.291 & 0.356 & 0.293 & 0.350 & \textbf{0.273} & \textbf{0.340} & 1.867 & 0.869 & 0.325 & 0.363 & 0.415 & 0.452 & 0.314 & 0.360 & 0.327 & 0.361 & 0.455 & 0.487 & 0.457 & 0.483 & 0.747 & 0.686 \\

\multirow{1}{*}{ETT$_{h2\to h1}$}
& \textbf{0.425} & \textbf{0.439} & 0.496 & 0.480 & 0.479 & 0.474 & 1.961 & 0.981 & 0.757 & 0.578 & 0.703 & 0.574 & 0.565 & 0.513 & 0.865 & 0.621 & 0.768 & 0.613 & 0.757 & 0.608 & 1.223 & 0.857 \\

\multirow{1}{*}{ETT$_{h2\to m2}$}
& \textbf{0.256} & \textbf{0.318} & 0.297 & 0.353 & 0.272 & 0.341 & 1.867 & 0.869 & 0.335 & 0.370 & 0.328 & 0.386 & 0.325 & 0.365 & 0.342 & 0.376 & 0.363 & 0.409 & 0.366 & 0.411 & 0.658 & 0.572 \\

\multirow{1}{*}{ETT$_{m1\to h2}$}
& \textbf{0.329} & \textbf{0.358} & 0.354 & 0.397 & 0.381 & 0.412 & 0.992 & 0.708 & 0.433 & 0.439 & 0.464 & 0.475 & 0.439 & 0.438 & 0.457 & 0.454 & 0.468 & 0.483 & 0.470 & 0.479 & 0.768 & 0.680 \\

\multirow{1}{*}{ETT$_{m1\to m2}$}
& \textbf{0.243} & \textbf{0.309} & 0.264 & 0.319 & 0.268 & 0.320 & 1.867 & 0.869 & 0.313 & 0.348 & 0.335 & 0.389 & 0.296 & 0.334 & 0.322 & 0.354 & 0.460 & 0.479 & 0.469 & 0.484 & 0.657 & 0.637 \\

\multirow{1}{*}{ETT$_{m2\to h2}$}
& \textbf{0.283} & \textbf{0.348} & 0.359 & 0.399 & 0.354 & 0.400 & 0.992 & 0.708 & 0.435 & 0.443 & 0.455 & 0.471 & 0.409 & 0.425 & 0.435 & 0.443 & 0.421 & 0.406 & 0.423 & 0.439 & 0.612 & 0.750 \\

\multirow{1}{*}{ETT$_{m2\to m1}$}
& \textbf{0.374} & \textbf{0.382} & 0.432 & 0.426 & 0.414 & 0.438 & 1.933 & 0.984 & 0.769 & 0.567 & 0.649 & 0.537 & 0.568 & 0.492 & 0.769 & 0.567 & 0.751 & 0.589 & 0.755 & 0.591 & 1.213 & 0.829 \\

\hline
\multicolumn{1}{c}{$1^{\text{st}}$ Count} & 
\multicolumn{2}{c}{\textbf{68}} & 
\multicolumn{2}{c}{2} & 
\multicolumn{2}{c}{10} & 
\multicolumn{2}{c}{0} & 
\multicolumn{2}{c|}{0} & 
\multicolumn{2}{c}{0} & 
\multicolumn{2}{c}{0} & 
\multicolumn{2}{c}{0} & 
\multicolumn{2}{c}{0} & 
\multicolumn{2}{c}{0} & 
\multicolumn{2}{c}{0} \\
\hline
\bottomrule
\end{tabular}
\end{small}
}
\vspace{-10pt}
\end{table*}

%% file: tables_brief/generation_brief.tex
\begin{table}[htbp]
\caption{
To further verify the comprehensive performance of the proposed \myformer in Long-term Time-series Generation, we introduce additional evaluation metrics: Context-FID Score, Correlational Score, Discriminative Score, Predictive Score, with length of 256. We boldface the best performance on all metrics and datasets, respectively. For the diffusion-based generative models, we conducted three repeated experiments and report means and variances.
}\label{tab:generation_brief}
\vspace{-10pt}
\vskip 0.05in
\centering
\resizebox{1.0\columnwidth}{!}{
\begin{threeparttable}
\begin{small}
\renewcommand{\multirowsetup}{\centering}
\tabcolsep=0.05cm
\renewcommand\arraystretch{1.3}
\begin{tabular}{cc|ccccccc}
\toprule
\hline

\multicolumn{2}{c|}{\multirow{2}{*}{\scalebox{1.0}{Dataset}}} &
\multicolumn{1}{c}{\textbf{\myformer}} & 
\multicolumn{1}{c}{Diffusion-TS} & 
\multicolumn{1}{c}{TimeGAN} & 
\multicolumn{1}{c}{TimeVAE} & 
\multicolumn{1}{c}{TimeGrad} & 
\multicolumn{1}{c}{TimeDiff} &
\multicolumn{1}{c}{Cot-GAN} \\

& &
\multicolumn{1}{c}{\textbf{(Ours)}} & 
\multicolumn{1}{c}{\cite{Yuan2024DiffusionTSID}} & 
\multicolumn{1}{c}{\cite{yoon2021timegan}} & 
\multicolumn{1}{c}{\cite{desai2023timevae}} & 
\multicolumn{1}{c}{\cite{Rasul2021timegrad}} & 
\multicolumn{1}{c}{\cite{shen2024timediff}} & 
\multicolumn{1}{c}{\cite{xu2022cotgan}} \\

\hline
\multirow{4}[0]{*}{ETTh} 

& Context-FID & 
\textbf{0.347±.010} & 0.423±.038 & 5.872±.208 & 0.826±.093 & 2.899±.289 & 3.524±.830 & 4.075±.894 \\

& Correlational & 
\textbf{0.054±.003} & 0.064±.007 & 0.522±.013 & 0.046±.007 & 0.199±.003 & 0.135±.006 & 0.222±.010 \\

& Discriminative & 
\textbf{0.048±.011} & 0.060±.030 & 0.442±.056 & 0.178±.076 & 0.304±.068 & 0.243±.005 & 0.461±.010 \\

& Predictive & 
\textbf{0.090±.006} & 0.109±.013 & 0.220±.008 & 0.110±.027 & 0.132±.001 & 0.118±.003 & 0.129±.000 \\

\hline
\multirow{4}[0]{*}{Energy} 

& Context-FID & 
\textbf{0.122±.019} & 0.126±.024 & 5.032±.831 & 3.768±.998 & 5.572±.584 & 4.735±.729 & 2.533±.467 \\

& Correlational & 
\textbf{0.341±.039} & 0.361±.092 & 4.487±.214 & 1.279±.114 & 5.690±.102 & 1.800±.138 & 3.739±.089 \\

& Discriminative & 
\textbf{0.252±.047} & 0.290±.123 & 0.499±.000 & 0.499±.000 & 0.499±.000 & 0.437±.095 & 0.498±.004 \\

& Predictive & 
\textbf{0.214±.001} & 0.245±.001 & 0.351±.004 & 0.353±.003 & 0.251±.000 & 0.251±.000 & 0.275±.004 \\

\hline
\bottomrule
\end{tabular}
\end{small}
\end{threeparttable}
}
\end{table}

%% file: tables_brief/solar_mujoco.tex
\begin{table}{}
\vspace{-5pt}
\caption{
The performance of \myformer and existing models on imputation. To apply generative paradigms to imputation tasks, we enable TimeControl to generate masked portions from Gaussian distributions using unmasked segments as conditional variables. Specifically, we collected sensor data from simulation environments within popular general-purpose physics engines MuJoCo and employed varying masking ratios to evaluate the interpolation performance of TimeControl.
}\label{tab:solar_mujoco}
\vspace{-10pt}
\vskip 0.05in
\centering
\resizebox{1.0\columnwidth}{!}{
\begin{small}
\renewcommand{\multirowsetup}{\centering}
\tabcolsep=0.05cm
\renewcommand\arraystretch{1.3}
\begin{tabular}{ccccccccccccccc}
    \toprule

    \multicolumn{1}{c}{\multirow{2}{*}{\scalebox{1.0}{MuJoCo Imputation}}} &
    \multicolumn{1}{c}{\textbf{\myformer}} & 
    \multicolumn{1}{c}{Diffusion-TS} & 
    \multicolumn{1}{c}{SSSD} & 
    \multicolumn{1}{c}{CSDI} & 
    \multicolumn{1}{c}{TimeGAN} & 
    \multicolumn{1}{c}{TimeVAE} & 
    \multicolumn{1}{c}{TimeGrad} & 
    \multicolumn{1}{c}{TimeDiff} \\

    & 
    \multicolumn{1}{c}{\textbf{(Ours)}} & 
    \multicolumn{1}{c}{\cite{Yuan2024DiffusionTSID}} & 
    \multicolumn{1}{c}{\cite{alcaraz2022sssd}} & 
    \multicolumn{1}{c}{\cite{Tashiro2022csdi}} & 
    \multicolumn{1}{c}{\cite{yoon2021timegan}} & 
    \multicolumn{1}{c}{\cite{desai2023timevae}} & 
    \multicolumn{1}{c}{\cite{Rasul2021timegrad}} & 
    \multicolumn{1}{c}{\cite{shen2024timediff}} \\
    
    

    
    
    \hline
    
    70\% Mask & 
    \textbf{0.211} & 
    0.373 & 
    0.596 & 
    0.242 &
    0.379 &
    0.561 &
    0.493 &
    0.663 \\
    
    80\% Mask & 
    \textbf{0.374} & 
    0.432 & 
    1.003 & 
    0.617 &
    0.745 &
    1.112 &
    0.825 &
    0.768 \\
    
    90\% Mask & 
    \textbf{0.485} & 
    0.731 & 
    1.902 & 
    4.843 &
    0.864 &
    0.798 &
    1.085 &
    1.142 \\

    \bottomrule
  \end{tabular}
  \end{small}
}
\vspace{-5pt}
\end{table}

%% file: tables_brief/forecasting_tmdm.tex
\begin{table}[htpb]
\vspace{-10pt}
\caption{
To comprehensively verify the performance of TimeControl on probabilistic generation tasks, we introduce new evaluation metrics (CRPS, QICE and PICP) to supplement the experiments. Specifically, TimeControl outperforms existing methods on most datasets and metrics.
}\label{tab:brief_forecasting_tmdm}
\vspace{-5pt}
\vskip 0.05in
\centering
\resizebox{1.0\columnwidth}{!}{
\begin{small}
\renewcommand{\multirowsetup}{\centering}
\tabcolsep=0.25cm
\renewcommand\arraystretch{1.3}
\begin{tabular}{cccc|ccc|ccc}
\toprule
\hline

\multicolumn{1}{c}{\multirow{2}{*}{Model}} &
\multicolumn{3}{c|}{ETTm2} & 
\multicolumn{3}{c|}{Electricity} &
\multicolumn{3}{c}{Traffic} \\

\cline{2-10} &
QICE & CRPS & PICP &
QICE & CRPS & PICP &
QICE & CRPS & PICP \\
\hline

SSSD~\cite{alcaraz2022sssd} &
4.88 & 0.57 & 72.58 &
5.26 & 0.46 & 79.34 &
3.88 & 0.41 & 84.30 \\
\hline

CSDI~\cite{Tashiro2022csdi} &
5.07 & 0.50 & 71.78 &
4.74 & 0.42 & 78.94 &
3.50 & 0.37 & 83.51 \\
\hline

TimeDiff~\cite{shen2024timediff} &
14.22 & 0.40 & 13.16 &
12.74 & 0.38 & 32.37 &
13.53 & 0.28 & 9.11 \\
\hline

mrDiffusion~\cite{shen2024mrdiffusion} &
5.37 & 0.54 & 71.62 &
5.34 & 0.40 & 75.93 &
3.80 & 0.39 & 82.28 \\
\hline

TMDM~\cite{li2024ldt} &
3.75 & 0.37 & 73.20 &
3.81 & 0.33 & 82.35 &
2.36 & 0.26 & 86.83 \\
\hline

\rowcolor{blue!10}
\textbf{TimeControl} &
\textbf{3.51} & \textbf{0.26} & \textbf{73.91} &
\textbf{3.26} & \textbf{0.28} & \textbf{83.78} &
\textbf{2.79} & \textbf{0.25} & \textbf{87.45} \\

\hline
\bottomrule
\end{tabular}
  \end{small}
}
\vspace{-5pt}
\end{table}

%% file: tables_brief/ablation_brief.tex
\begin{table}
\vspace{-10pt}
\caption{
We demonstrate the performance comparison of TimeContrl and its three variants across four standard benchmarks.
}
\vspace{-10pt}
\vskip 0.05in
\label{tab:ablation_brief}
\centering
\tabcolsep=0.25cm
\renewcommand\arraystretch{1.3}
\resizebox{1.0\columnwidth}{!}{
\begin{tabular}{lccccccc}
\toprule

&
ETTh1 &
ETTm1 &
Electricity &
Weather &
Average &
$\uparrow$ \\

\midrule
\rowcolor{blue!10}
\myformer &
\textbf{0.334} &
\textbf{0.288} &
\textbf{0.149} &
\textbf{0.202} &
\textbf{0.273} &
\textbf{-} \\

w/o ConditionNet &
0.412 &
0.361 &
0.219 &
0.256 &
0.349 &
27.8\% \\

w/o Adapter &
0.371 &
0.339 &
0.170 &
0.224 &
0.310 &
13.6\% \\

w/o Generation-style &
0.387 &
0.322 &
0.187 &
0.234 &
0.318 &
16.5\% \\

\bottomrule
\end{tabular}
}
\vspace{-15pt}
\end{table}

%% file: tables_brief/complexity.tex
\begin{table}[htpb]
\vspace{-10pt}
\caption{
We compared TimeControl with foundation and proprietary models in metrics such as memory and inference speed.
}\label{tab:computational}
\vspace{-10pt}
\vskip 0.05in
\centering
\resizebox{1.0\columnwidth}{!}{
\begin{small}
\renewcommand{\multirowsetup}{\centering}
\tabcolsep=0.30cm
\renewcommand\arraystretch{1.3}
\begin{tabular}{c|ccc}
\toprule
\hline

\textbf{Metrics(ETTh1-96)} &
\textbf{Total Param. (M)} &
\textbf{Mem. (MiB)} &
\textbf{Speed (s/iter)} \\

\hline

UniTime~\cite{liu2024unitime} & 439.52 & 2074 & 0.335 \\
TimeLLM~\cite{Jin2023TimeLLMTS} & 3623.71 & 4537 & 0.184 \\
\rowcolor{blue!10}
\myformer (Ours) & 256.84 & 1273 & 0.095 \\

\hline

\textbf{Metrics(Electricity)} &
\textbf{Total Param. (M)} &
\textbf{Mem. (MiB)} &
\textbf{Speed (s/iter)} \\

CSDI~\cite{Tashiro2022csdi} & 54.29 & 983 & 0.245 \\
DiffusionTS~\cite{Yuan2024DiffusionTSID} & 258.64 & 1539 & 0.119 \\
\rowcolor{blue!10}
\myformer (Ours) & 283.79 & 1609 & 0.270 \\

\hline
\bottomrule
\end{tabular}
\end{small}
}
\vspace{-10pt}
\end{table}

%% file: tables_brief/sensitivity.tex
\begin{table}[htpb]
\caption{
To verify the sensitivity of \myformer to hyperparameter selection, this table shows the performance of \myformer with different parameter scales on multiple benchmarks, with a performance metric of MSE. Where, $L$ and $D$ denote the number of stacked blocks and the number of output channels in the condition net and denoising net, respectively. e boldface the best performance on all datasets.
}\label{tab:exp_sensitivity}
\vspace{-10pt}
\vskip 0.05in
\centering
\resizebox{1.0\columnwidth}{!}{
\begin{small}
\renewcommand{\multirowsetup}{\centering}
\tabcolsep=0.25cm
\renewcommand\arraystretch{1.3}
\begin{tabular}{cccccccc}
\toprule

\multicolumn{1}{c}{\multirow{1}{*}{\scalebox{1.1}{Hyperparameter}}} &
\multicolumn{1}{c}{ETTh1} & 
\multicolumn{1}{c}{ETTh2} &
\multicolumn{1}{c}{ETTm1} &
\multicolumn{1}{c}{ETTm2} &
\multicolumn{1}{c}{Weather} &
\multicolumn{1}{c}{Electricity} &
\multicolumn{1}{c}{Traffic} \\

\hline
\multirow{1}{*}{\scalebox{1.0}{\shortstack{(L=2,D=128)}}} &
0.352 & 
0.307 & 
0.335 & 
0.244 & 
0.215 & 
0.153 & 
0.311 \\

\hline
\multirow{1}{*}{\scalebox{1.0}{\shortstack{(L=2,D=256)}}} &
0.344 & 
0.302 & 
0.319 & 
0.254 & 
0.211 & 
0.158 & 
0.325 \\

\hline
\multirow{1}{*}{\scalebox{1.0}{\shortstack{(L=3,D=128)}}} &
0.328 & 
\textbf{0.274} & 
0.321 & 
0.239 & 
0.203 & 
0.154 & 
\textbf{0.296} \\

\hline
\rowcolor{blue!10}
\multirow{1}{*}{\scalebox{1.0}{\shortstack{\textbf{(L=3,D=256)}}}} &
0.334 & 
0.285 & 
\textbf{0.288} & 
\textbf{0.224} & 
0.202 & 
0.149 & 
0.301 \\

\hline
\multirow{1}{*}{\scalebox{1.0}{\shortstack{(L=4,D=128)}}} &
\textbf{0.322} & 
0.280 & 
0.302 & 
0.235 & 
\textbf{0.198} & 
\textbf{0.145} & 
0.298 \\

\hline
\multirow{1}{*}{\scalebox{1.0}{\shortstack{(L=4,D=256)}}} &
0.340 & 
0.313 & 
0.319 & 
0.234 & 
0.214 & 
0.160 & 
0.306 \\

\bottomrule
\end{tabular}
\end{small}
}
\vspace{-10pt}
\end{table}

%% file: segments/6-Conclusion.tex
\section{Conclusion}
\vspace{-5pt}
We propose a novel domain-generalization time series diffusion (\myformer) model. Leveraging the superior probability distribution modeling capability of diffusion models, \myformer captures the joint distribution across multiple data domains. 
To enhance cross-domain generalization, condition net extracts common fluctuation patterns during pre-training, while denoising net utilizes multi-scale representations as context for reverse denoising. Additionally, the adapter module maps shared patterns to the target domain's latent space, enabling generation of time series samples that match the style of the specified domain.

\paragraph{Discussion} Diffusion models suffer from instability and slow convergence during training, and require multiple iterative denoising steps during inference, unlike traditional deep models. However, \myformer reduces computational overhead compared to existing time-series foundation models, thanks to three novel designs: a patch-wise attention mechanism, iterative denoising in the real sequence space, and the DDIM algorithm.

%% file: segments/Appendix.tex
\section{Diffusion Model}
\label{sec:diffusion}
The diffusion model is a popular generative model and has attracted significant attention in various domains, such as image, video, 3-D objective, etc. A well-known diffusion model is the denoising diffusion probabilistic model (DDPM). DDPM consists of a forward diffusion process and a backward denoising process. The diffusion process means gradually adding Gaussian noise to the real samples of the dataset, while the denoising process means gradually denoising the noisy data to restore the real data points.

\newcommand{\bfx}{\mathbf{x}}
Given a data point sampled from a real data distribution $\bfx_0 \sim q(\bfx)$, the forward diffusion process gradually adds Gaussian noise $T$ steps, producing a series of noisy samples $\{\bfx_1, \cdots, \bfx_T\}$. The step size are controlled by a variance schedule $\{\beta_T \in (0,1)\}_{t=1}^{T}$. The diffusion process can be formulated as:
\begin{align}
    q(\mathbf{x}_t|\mathbf{x}_{t-1}) = \mathcal{N}(\mathbf{x}_t;\sqrt{1-\beta_t}\mathbf{x}_{t-1}, \beta_t\mathbf{I}).
\end{align}
Which means $\bfx_t$ is sampled from $q(\bfx_t|\bfx_{t-1})$, satisfied the Gaussian distribution $\mathcal{N}(\mathbf{x}_t;\sqrt{1-\beta_t}\mathbf{x}_{t-1}, \beta_t\mathbf{I})$. The diffusion process follows a Markov process:
\begin{align}
    q(\bfx_{1:T}|\bfx_0)= \prod \limits_{t=1}^T q(\bfx_t|\bfx_{t-1})
\end{align}
$\bfx_t$ is defined by $\bfx_{t-1}$ and $\beta_{T}$, and can be directly calculated with given $\bfx_0$ and $\{\beta_1, \cdots, \beta_T\}$ step values. Let $\alpha_t=1-\beta_t$, and $\bar{\alpha}_t=\prod \limits_{t=1}^T \alpha_i$, with the re-parametric, we get:
\begin{align}
    \begin{split}
        \bfx_t &= \sqrt{\alpha_t}\bfx_{t-1} + \sqrt{1-\alpha_t}\mathbf{z}_{t-1} \\
        &= \sqrt{\alpha_t \alpha_{t-1}}\bfx_{t-2} + \sqrt{1-\alpha_t \alpha_{t-1}}\mathbf{z}_{t-2} \\
        &= \cdots \\
        &= \sqrt{\bar{\alpha}_t}\bfx_0+\sqrt{1-\bar{\alpha}_t}\mathbf{z}_0
    \end{split}
\end{align}
where $\mathbf{z}_0, \cdots, \mathbf{z}_{t-1} \sim \mathcal{N}(\mathbf{0}, \mathbf{I})$, and 
\begin{align}
    q(\bfx_t|\bfx_0)=\mathcal{N}(\bfx_t;\sqrt{\bar{\alpha}_t}\bfx_0, (1-\bar{\alpha}_t)\mathbf{I}).
\end{align}
This indicates that with $\bfx_0$ and a fixed-value sequence $\{\beta_T\in(0,1)\}_{t=1}^{T}$, and sample $\mathbf{z}$ from norm distribution $\mathcal{N}(\mathbf{0,I})$, $\bfx_t$ is defined. In general, we can afford a larger update step when the sample gets noisier, so $\beta_1<\beta_2<\cdots<\beta_T$, and therefore $\bar{\alpha}_1>\cdots>\bar{\alpha}_T$.

The reverse denoising process samples from $q(\bfx_{t-1}|\bfx_t)$, and we can reconstruct the real data point for a random Gaussian distribution. However, we need to find the data distribution from the whole dataset, and we cannot predict the conditional distribution $q(\bfx_{t-1}|\bfx_t)$ directly, so we need to learn a model $p_\theta$ to approximately simulate this conditional probability to run the inverse diffusion process.
\begin{align}
\begin{split}
    p_\theta(\bfx_{0:T}) &= p(\bfx_T)\prod \limits_{t=1}^T p_\theta(\bfx_{t-1}|\bfx_t)\\
    p_\theta(\bfx_{t-1}|\bfx_{t}) &= \mathcal{N}\left(\bfx_{t-1};\mu_\theta(\bfx_t,t),\Sigma_\theta(\bfx_t, t)\right).
\end{split}
\end{align}
Given $\bfx_t$ and $\bfx_0$ the posteriori diffusion conditional probability can be formulated as :
\begin{align}
    q(\bfx_{t-1}|\bfx_t, \bfx_0) = \mathcal{N}(\bfx_{t-1};\tilde{\mu}(\bfx_t, \bfx_0), \tilde{\beta}_t\mathbf{I}).
    \label{eqn:10}
\end{align}
Following the Bayes' rule:
\begin{align}
\begin{split}
    &q(\bfx_{t-1}|\bfx_t,\bfx_0) = q(\bfx_t|\bfx_{t-1},\bfx_0)\frac{q(\bfx_{t-1}|\bfx_0)}{q(\bfx_t|\bfx_0)} \\
    & \propto\exp\left(-\frac{1}{2}\left(\frac{\left(\bfx_{t}-\sqrt{\bar{\alpha}_{t}} \bfx_{t-1}\right)^{2}}{\beta_{t}}+\frac{\left(\bfx_{t-1}-\sqrt{\bar{\alpha}_{t-1}} \bfx_{0}\right)^{2}}{1-\bar{\alpha}_{t-1}}-\frac{\left(\bfx_{t}-\sqrt{\bar{\alpha}_{t}} \bfx_{0}\right)^{2}}{1-\bar{\alpha}_{t}}\right)\right) \\
    &= \exp\left(-\frac{1}{2}\left(\left(\frac{\alpha_{t}}{\beta_{t}}+\frac{1}{1-\bar{\alpha}_{t-1}}\right) \bfx_{t-1}^{2}-\left(\frac{2\sqrt{\alpha_{t}}}{\beta_{t}} \bfx_{t}+\frac{2\sqrt{\bar{\alpha}_{t-1}}}{1-\bar{\alpha}_{t-1}} \bfx_{0}\right) \bfx_{t-1}+C\left(\bfx_{t}, \bfx_{0}\right)\right)\right.
\end{split}
\end{align}
where $C\left(\bfx_{t}, \bfx_{0}\right)$ is a function contains $\bfx_t$ and $\bfx_0$, without $\bfx_{t-1}$. The mean and variance can be calculated by:
\begin{align}
\begin{split}
\tilde{\beta}_{t} &= 1/(\frac{\alpha_{t}}{\beta_{t}} + \frac{1}{1 - \bar{\alpha}_{t-1}}) = \frac{1 - \bar{\alpha}_{t-1}}{1 - \bar{\alpha}_{t}} \cdot \beta_{t} \\
\tilde{\mu}_{t}(\bfx_{t}, \bfx_{0}) &= (\frac{\sqrt{\alpha_{t}}}{\beta_{t}} \bfx_{t} + \frac{\sqrt{\bar{\alpha}_{t-1}}}{1 - \bar{\alpha}_{t-1}} \bfx_{0})/(\frac{\alpha_{t}}{\beta_{t}} + \frac{1}{1 - \bar{\alpha}_{t-1}}) \\
&= \frac{\sqrt{\alpha_{t}}(1 - \bar{\alpha}_{t-1})}{1 - \bar{\alpha}_{t}} \bfx_{t} + \frac{\sqrt{\bar{\alpha}_{t-1}} \beta_{t}}{1 - \bar{\alpha}_{t}} \bfx_{0}
\end{split}
\label{eqn:12}
\end{align} 
In the forward process, we have $\bfx_0=\frac{1}{\sqrt{\bar{\alpha}_t}}(\bfx_t=\sqrt{1-\bar{\alpha}_t}\mathbf{z}_t)$.
Taking into Eqn. \ref{eqn:12}, we have:
\begin{align}
\tilde{\mu}_{t} &= \frac{\sqrt{\alpha_{t}}(1-\bar{\alpha}_{t-1})}{1-\bar{\alpha}_{t}}\bfx_{t} + \frac{\sqrt{\bar{\alpha}_{t-1}}\beta_{t}}{1-\bar{\alpha}_{t}}\frac{1}{\sqrt{\bar{\alpha}_{t}}}(\bfx_{t}-\sqrt{1-\bar{\alpha}_{t}}z_{t}) \\
&= \frac{1}{\sqrt{\alpha_{t}}}(\bfx_{t}-\frac{\beta_{t}}{\sqrt{1-\bar{\alpha}_{t}}}z_{t})
\label{eqn:13}
\end{align}

To train the diffusion model, one uniformly samples $t$ form $\{1,2,\cdots,T\}$ and then minimizes the following KL-divergence:
\begin{align}
    \mathcal{L}_t=D_{KL}(q(\bfx_{t-1}|\bfx_{t})||p_\theta(\bfx_{t-1}|\bfx_t)).
\end{align}
Connecting Eqn. \ref{eqn:10}, \ref{eqn:12}, and \ref{eqn:13}, the training objective is transformed into:
\begin{align}
    \mathcal{L}_t=\frac{1}{2\sigma_t^2}||\tilde{\mu}_{t}(\bfx_t,\bfx_0, t)-\mu_\theta(\bfx_t,t)||^2.
\end{align}


\clearpage
\section{More Experimental Setup}
\subsection{Baseline Models}
\subsubsection{Time Series Forecasting Foundation Models}
We selected mainstream foundational models for time series analysis to conduct comparative evaluations, including: 
(1) unified models pre-trained on time series data, such as Moirai~\cite{woo2024moirai} and UniTime~\cite{liu2024unitime}; 
(2) models that adapt pre-trained large-scale vision or language models to time series forecasting scenarios, such as TimeVLM~\cite{zhong2025timevlm} and TimeLLM~\cite{Jin2023TimeLLMTS}. We elaborate on the implementation details and characteristics of the selected baseline models in the subsequent section.

\textbf{TimeVLM~\cite{zhong2025timevlm}} is a novel multimodal framework that leverages frozen pre-trained Vision-Language Models to bridge temporal, visual, and textual data for forecasting. It integrates components for temporal feature retrieval, vision-based series encoding, and contextual text generation. The model achieves superior performance in few-shot and zero-shot scenarios, setting a new direction for time series analysis.

\textbf{Moirai~\cite{woo2024moirai}} is a universal time series forecasting transformer that overcomes key challenges in cross-frequency learning, multivariate scalability, and distributional diversity through novel architectural enhancements, achieving superior zero-shot performance across domains. Pre-trained on the Large-scale Open Time Series Archive (LOTSA) with 27B+ observations, it introduces masked encoder-based adaptations to conventional transformers.

\textbf{UniTime~\cite{liu2024unitime}} is a unified model for cross-domain multivariate time series forecasting that addresses key learning challenges. It adapts to varying data characteristics, uses domain instructions and a Language-TS Transformer for modality alignment, and employs masking to balance domain convergence rates. The model advances excellent performance and demonstrates strong zero-shot transferability across diverse application domains.

\textbf{GPT4TS~\cite{Zhou2023OneFA}} is a foundation model that adapts frozen pre-trained language or vision models for general time series analysis. By leveraging knowledge from large-scale source domains, it overcomes the data scarcity challenge in time series pre-training. The model achieves outstanding performance across diverse tasks including classification, anomaly detection, and forecasting, demonstrating remarkable cross-domain transfer capability.

\textbf{TimeLLM~\cite{Jin2023TimeLLMTS}} is a novel reprogramming framework that effectively repurposes frozen large language models (LLMs) for general time series forecasting. It overcomes the modality alignment challenge by reprogramming time series data with text prototypes and augmenting LLMs' reasoning capabilities through Prompt-as-Prefix (PaP). This approach enables powerful cross-modal transfer, achieving remarkable performance that surpasses specialized forecasting models and excelling in both few-shot and zero-shot learning scenarios.

\textbf{LLM4TS~\cite{Chang2023LLM4TSAP}} is a novel framework that adapts pre-trained Large Language Models for time-series forecasting through a two-stage fine-tuning strategy. It overcomes key challenges of modality alignment and multi-scale processing by employing time-series alignment and a novel two-level aggregation method. The framework achieves superior performance across multiple datasets in both full-shot and few-shot scenarios, demonstrating the significant transferability of LLM representations for time-series analysis and establishing a new paradigm for cross-modal forecasting.

\subsubsection{Time Series Forecasting Proprietary Models}
To thoroughly validate that TimeControl achieves optimal performance across most data domains, we fine-tuned the pre-trained weights on specific domains. Correspondingly, we selected leading and most representative domain-specific models as benchmarks, including transformer-based approaches such as DUET~\cite{qiu2025duet}, PDF~\cite{dai2024pdf}, and Pathformer~\cite{chen2024pathformer}, and linear-based TimeMixer~\cite{wang2024timemixer}, among others. We have summarized the unique design characteristics and a concise overview.

\textbf{DUET~\cite{qiu2025duet}} is a dual-clustering framework for multivariate time series forecasting that addresses temporal heterogeneity and complex channel correlations through a temporal clustering module for fine-grained pattern extraction and a channel-soft-clustering module for frequency-domain channel relationship modeling.

\textbf{PDF~\cite{dai2024pdf}} is a novel time series forecasting approach that captures 2D temporal variations through multi-periodic decoupling and dual variation modeling, outperforming existing CNN- and Transformer-based methods in long-term forecasting accuracy and efficiency by explicitly disentangling intricate temporal patterns into diverse components.

\textbf{Pathformer~\cite{chen2024pathformer}} is a multi-scale Transformer with adaptive pathways that dynamically models time series across varying temporal resolutions and distances, employing dual attention mechanisms to capture both global correlations and local details while adaptively adjusting it to input, achieving superior generalization in transfer scenarios.

\textbf{TimeMixer~\cite{wang2024timemixer}} is a fully MLP-based time series forecasting architecture that leverages multiscale-mixing to disentangle intricate temporal variations, employing past-decomposable-mixing and future-multipredictor-mixing blocks to hierarchically aggregate microscopic seasonal and macroscopic trend information while harnessing complementary predictors, achieving excellent performance with efficient computational overhead.

\textbf{PatchTST~\cite{Nie2023PatchTST}} is a channel-independent transformer architecture for multivariate time series forecasting that achieves superior efficiency and accuracy through token patching, enabling local semantic retention, quadratic reduction in attention complexity. Unifying patch-based tokenization with weight-sharing across univariate channels.

\textbf{Dlinear~\cite{zeng2023dlinear}} is a one-layer linear model that exposes critical limitations of transformer architectures in temporal relation extraction, achieving outstanding performance through deliberate avoidance of self-attention's permutation invariance, thereby establishing new baselines and research directions for time series analysis.

\subsubsection{Time Series Generation Models}
To verify the effectiveness of the multi-scale conditional denoising and generation-style guidance designed by TimeControl, we selected the most widely used time series generation model as the baseline, which includes methods based on Diffusion, such as mrDiffusion~\cite{shen2024mrdiffusion} and LDT~\cite{li2024ldt}, as well as methods based on GAN, such as TimeGAN~\cite{yoon2021timegan} and CoT-GAN~\cite{xu2022cotgan}.

\textbf{Diffusion-TS~\cite{Yuan2024DiffusionTSID}} is a novel diffusion model for multivariate time series generation. It employs an encoder-decoder transformer with disentangled representations and a fourier loss, directly reconstructing samples to enhance semantic capture and sequential detail for high-quality, interpretable results.

\textbf{mrDiffusion~\cite{shen2024mrdiffusion}} introduces a multi-resolution diffusion model for time series that leverages seasonal-trend decomposition. It diffuses fine-to-coarse trends sequentially, then generates data in an easy-to-hard, non-autoregressive manner by progressively adding finer details using coarser predictions as conditions.

\textbf{TMDM~\cite{li2024ldt}} is a transformer-modulated diffusion model that unites transformers with a conditional diffusion process for multivariate time series forecasting. It uses transformers to extract historical insights as prior knowledge, capturing covariate-dependence to enable precise distribution forecasting.

\textbf{SSSD~\cite{alcaraz2022sssd}} establishes a diffusion-based imputation model for time series. It integrates structured state space models to capture long-term dependencies, enabling effective probabilistic imputation across diverse data scenarios.

\textbf{CSDI~\cite{Tashiro2022csdi}} proposes a conditional score-based diffusion model for time series imputation. It is explicitly trained to exploit correlations in observed data, enabling accurate probabilistic value estimation and can also be applied to interpolation and forecasting tasks.

\textbf{TimeGAN~\cite{yoon2021timegan}} is a generative framework for time-series data that merges unsupervised flexibility with supervised control. It uses a jointly learned embedding space to ensure generated sequences respect the original data's temporal dynamics and variable relationships.

\textbf{TimeVAE~\cite{desai2023timevae}} introduces a vae-based architecture for generating synthetic time-series data. It ensures interpretability, incorporates domain knowledge like seasonality, and offers reduced training times while accurately preserving temporal attributes.

\textbf{TimeGrad~\cite{Rasul2021timegrad}} is a versatile diffusion model for waveform generation. This non-autoregressive model converts white noise to structured audio through markov chain, enabling high-fidelity, fast synthesis for various tasks.

\textbf{TimeDiff~\cite{shen2024timediff}} is a non-autoregressive diffusion model for time series prediction. It introduces novel conditioning mechanisms like future mixup and autoregressive initialization to capture essential patterns and enhance forecasting.

\textbf{Cot-GAN ~\cite{xu2022cotgan}} is an adversarial framework for sequential data, using causal optimal Transport. It integrates temporal causality and an entropic penalty, enabling stable learning of time-dependent distributions via a robust, less biased sinkhorn divergence.

\input{tables_brief/dataset}
\subsection{Datasets and Benchmarks}
For the fair comparison of the zero-shot forecasting performance between \myformer and existing time series foundation models, we strictly adhered to the evaluation benchmarking protocol established by Chronos~\cite{ansari2024chronos}. 
Specifically, we employ 49 datasets aggregated from multiple sources, including the Monash Time Series Forecasting Repository~\cite{godahewa2021monash}, M-competitions, and public-domain datasets from Kaggle, for comprehensive training and evaluation of \myformer. 
These datasets encompass diverse application domains such as energy, transportation, healthcare, networking, meteorology, and finance, with sampling frequencies ranging from 5-minute intervals to annual recordings. 
This rich statistical diversity ensures \myformer develops robust prior knowledge across extensive temporal patterns. The detailed source attributions and specifications of benchmarks, is documented in Table~\ref{tab:dataset}.

To evaluate HaRP's zero-shot performance on the TFB forecasting benchmark~\cite{qiu2024tfb}, we implemented strict segregation between training and evaluation datasets to prevent data leakage. 
Specifically, we selected widely-used datasets from the TFB benchmark, which were not included in Chronos' pre-training corpus as our cross-domain generalization  evaluation set. 
Our comprehensive evaluation protocol consists of two components: (1) we pre-trained the TimeControl on the fusion domain form all Monash train datasets, firstly, and (2) zero-shot (without fine-tuning) versus full-shot (with fine-tuning) forecasting evaluation on TFB benchmarks. Below we detail the benchmark specifications:

\subsubsection{Monash Benchmark} Following~\cite{woo2024moirai}, we pre-trained the model on 39 Monash datasets~\cite{godahewa2021monash} using GluonTS~\cite{alexandrov2020gluonts}, including M1 Monthly, M3 Monthly, M3 Other, M4 Monthly, M4 Weekly, M4 Daily, M4 Hourly, Tourism Quarterly, Tourism Monthly, CIF 2016, Australian Electricity Demand, Bitcoin, Pedestrian Counts, Vehicle Trips, KDD Cup, Weather, NN5 Daily, NN5 Weekly, Carparts, FRED-MD, Traffic Hourly, Traffic Weekly, Rideshare, Hospital, COVID Deaths, Temperature Rain, Sunspot, Saugeen River Flow, and US Births.

\subsubsection{TFB Benchmark} We evaluate our model on 10 widely used TFB datasets, including ETTh1, ETTh2, ETTm1, ETTm2, Electricity, Traffic, Exchange, ICL, Solar and Weather. Performance is assessed using Mean Squared Error (MSE) and Mean Absolute Error (MAE), with lower values indicating better forecasting accuracy. 

\subsection{Evaluation Metrics}
\subsubsection{Discriminative and Predictive score}
The discriminative score is calculated as $\mid$accuracy $-$ 0.5$\mid$, while the predictive score is the mean absolute error (MAE) evaluated between the predicted values and the ground-truth values in test data. For a fair comparison, we reuse the experimental settings of TimeGAN \cite{yoon2021timegan} for the discriminative and predictive score. Both the classifier and sequence-prediction model use a 2-layer GRU-based neural network architecture. 

\subsubsection{Context-FID score}
A lower FID score means the synthetic sequences are distributed closer to the original data. we propose a Frechet Inception distance (FID)-like score, Context-FID (Context-Frechet Inception distance) score by replacing the Inception model of the original FID with a time series representation learning method called TS2Vec \cite{r58}. They have shown that the lowest scoring models correspond to the best-performing models in downstream tasks and that the Context-FID score correlates with the downstream forecasting performance of the generative model. Specifically, we first sample synthetic time series and real-time series respectively. Then we compute the FID score of the representation after encoding them with a pre-trained TS2Vec model.

\subsubsection{Correlational score}
We estimate the covariance of the $i^{th}$ and $j^{th}$ feature of time series as follows:
\begin{equation}
    {{\mathop{\rm cov}} _{i,j}} = \frac{1}{T}\sum\limits_{t = 1}^T {X_i^tX_i^t}  - \left( {\frac{1}{T}\sum\limits_{t = 1}^T {X_i^t} } \right)\left( {\frac{1}{T}\sum\limits_{t = 1}^T {X_{j}^t}} \right).
\end{equation}
Then the metric on the correlation between the real data and synthetic data is computed by
\begin{equation}
    \frac{1}{{10}}\sum\limits_{i,j}^d {\left| {\frac{{{\mathop{\rm cov}} _{i,j}^r}}{{\sqrt {{\mathop{\rm cov}} _{i,i}^r} {\mathop{\rm cov}} _{j,j}^r}} - \frac{{{\mathop{\rm cov}} _{i,j}^f}}{{\sqrt {{\mathop{\rm cov}} _{i,i}^f} {\mathop{\rm cov}} _{j,j}^f}}} \right|},
\end{equation}

\subsubsection{PICP}
Prediction Interval Coverage Probability (PICP) measures the empirical coverage probability of a specific prediction interval. It checks how often the true value falls within a single pre-defined prediction interval (e.g., between the 5th and 95th percentiles). A well-calibrated model should have a PICP close to the nominal coverage rate. The formulaic description of A is as follows:
\begin{equation}
    \mathrm{PICP}=\frac{1}{N}\sum_{n=1}^{N}\mathbb{I}\{y_n\geq\hat{y}_n^{\mathrm{low}}\}\cdot\mathbb{I}\{y_n\leq\hat{y}_n^{\mathrm{high}}\}.
\end{equation}
Where $N$ is the number of data points, $y_n$ is the true value for the $n$-th point, $\hat{y}_n^{\mathrm{low}}$ and $\hat{y}_n^{\mathrm{high}}$ are predicted low and high percentiles for the $n$-th point (e.g., 5th and 95th). $\mathbb{I}\{\cdot\}$ is the indicator function (1 if condition is true, 0 otherwise).

\subsubsection{QICE}
Quantile Interval Coverage Error (QICE) is a granular extension of PICP that evaluates the overall calibration of the predicted distribution. Instead of a single interval, it splits the predictive distribution into multiple adjacent quantile intervals (QIs) and checks if each QI contains the correct proportion of true values. It penalizes imbalances where some intervals are over-populated and others are under-populated. The formulaic description of A is as follows: 
\begin{small}
\begin{equation}
  \begin{split}
    \mathrm{QICE}=\frac{1}{M}\sum_{m=1}^{M}\left|r_m-\frac{1}{M}\right|~where~r_m= \\
    \frac{1}{N}\sum_{n=1}^{N}\mathbb{I}\{y_n\geq\hat{y}_n^{\mathrm{low},m}\}\cdot\mathbb{I}\{y_n\leq\hat{y}_n^{\mathrm{high},m}\},
  \end{split}
\end{equation}
\end{small}
Where $M$ is the number of quantile intervals (QI), $r_m$ is the empirical coverage rate for the $m$-th QI, $\hat{y}_n^{\mathrm{low},m}$ and $\hat{y}_n^{\mathrm{high},m}$ are predicted lower and upper bounds for the $m$-th QI and $n$-th point.

\subsubsection{CRPS}
Continuous Ranked Probability Score (CRPS) is a proper scoring rule that provides a comprehensive measure of predictive performance. It evaluates both calibration (reliability of uncertainty) and sharpness (concentration of the forecast). A lower CRPS indicates a better model. It compares the entire predictive cumulative distribution function (CDF) to the true observation. The CRPS is the integral of the Brier score at all probability thresholds. For a predictive CDF $F(\cdot)$ and a true observation $y$, it is given by:
\begin{equation}
    \mathrm{CRPS}(F,y)=\int_{-\infty}^{\infty}(F(t)-\mathbb{I}\{t\geq y\})^2dt.
\end{equation}
In practice, this is often computed using the quantile representation or approximated from ensemble forecasts.


\section{Further Discussion}
\subsection{Potential information loss during reverse denoising process} 
In the denoising process, models learn conditional probability distributions, but at high noise levels, local features can be at risk of being destroyed. To address this, TimeContrl introduces two key innovations. (1) Multi-Scale Condition-Denoising: Stacked blocks in the conditioning and denoising networks capture temporal features at different scales. Shallow blocks focus on long-term dependencies (e.g., trends) with large receptive fields, while deeper blocks reconstruct local fluctuation patterns, ensuring accurate sequence reconstruction across scales. (2) generation-style Guidance (GSG): GSG blends conditional and unconditional predictions, aligning the generated sequence with input patterns while balancing creative exploration and temporal detail preservation. These mechanisms ensure critical features are retained during denoising, enabling robust generalization across domains.

\subsection{High arithmetic overhead and time cost} 
As mentioned earlier, compared to traditional deep models, diffusion models have the defect of unstable and slow convergence during training and require multiple rounds of iterative denoising processes during inference. These drawbacks lead to the fact that diffusion-based time-series prediction models require high arithmetic overhead and time cost, both in the training and inference phases. However, \myformer exhibits lower computational overheads compared to existing time series foundation models, which due to three innovative designs that, patch-wise attention mechanism, iterative denoising in real sequence space and de-Markovised DDIM algorithm.

\section{More experimental results}
\input{tables_brief/forecasting_full_all}

\subsection{Full-shot Forecasting}
In Table~\ref{tab:forecasting_deep_all}, we present the results of the full-shot time series forecasting.

\input{tables_brief/forecasting_zero_all}
\subsection{Zero-shot Forecasting}
In Table~\ref{tab:brief_forecasting_zero_all}, we present the results of the zero-shot time series forecasting.

\input{tables_brief/generation_full}
\subsection{Generation}
In Table~\ref{tab:generation_full}, we present the results of the time series generation.

\input{tables_brief/imputation_full}
\subsection{Imputation}
The Table \ref{tab:imputation_full} demonstrates the imputation results of training from scratch on each particular dataset, the average MSE is reduced by \textbf{17.0\%}, \textbf{20.1\%} and \textbf{38.7\%} compared to the existing GPT4TS, TimesNet and PatchTST. Excitingly, the proposed \myformer achieves better overall results on the imputation task than existing multitasking foundation and specific models.
Surprisingly, the proposed \myformer shows better results on the dataset characterized by multi-periodic patterns, which conforms to the multi-scale representation mechanism designed in our Condition-Denoising component. Specifically,the average MSE is reduced by \textbf{41.1\%}, \textbf{42.3\%} and \textbf{26.3\%} compared to the existing GPT4TS, TimesNet and PatchTST on the ECL dataset.

\section{More visualizations}\label{sec:more_visualizations}
\begin{figure*}[t]
\begin{center}
\centerline{\includegraphics[width=2.0\columnwidth]{images/show_distribution_tsne_etth_1_720.png}}
\vspace{10pt}
\centerline{\includegraphics[width=2.0\columnwidth]{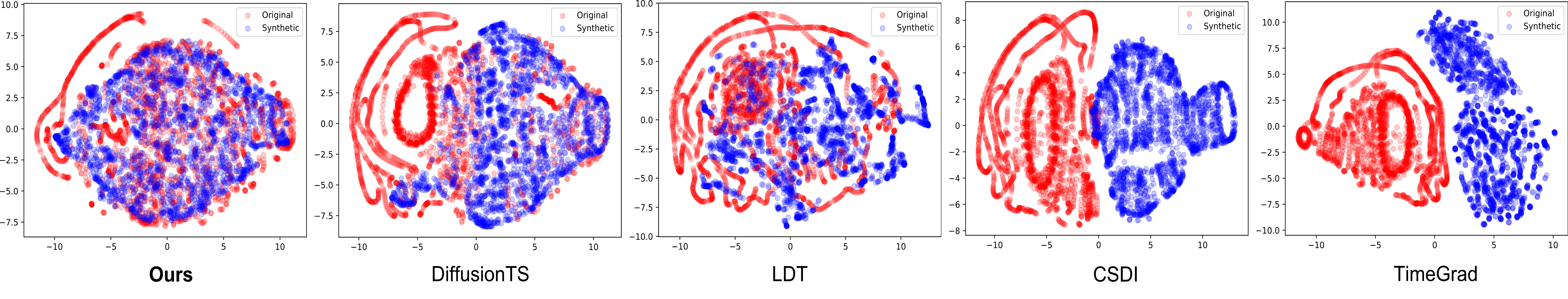}}
\vspace{10pt}
\centerline{\includegraphics[width=2.0\columnwidth]{images/show_distribution_tsne_etth_3.pdf}}
\vspace{10pt}
\centerline{\includegraphics[width=2.0\columnwidth]{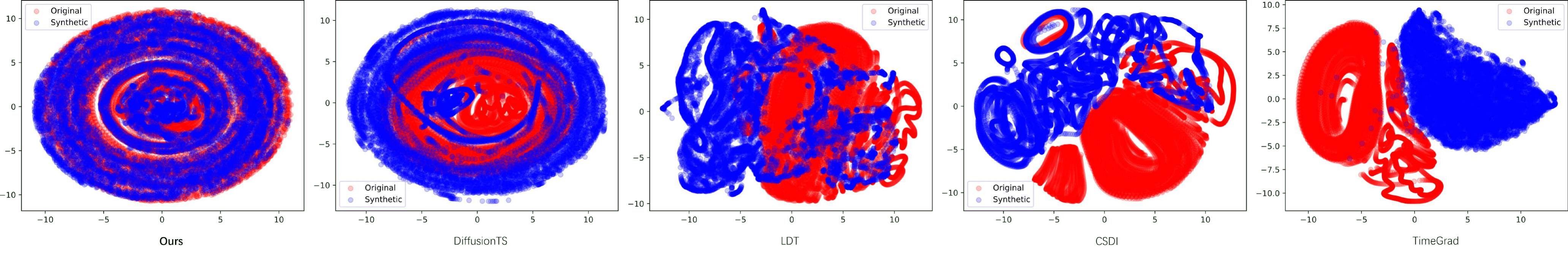}}
\vspace{10pt}
\centerline{\includegraphics[width=2.0\columnwidth]{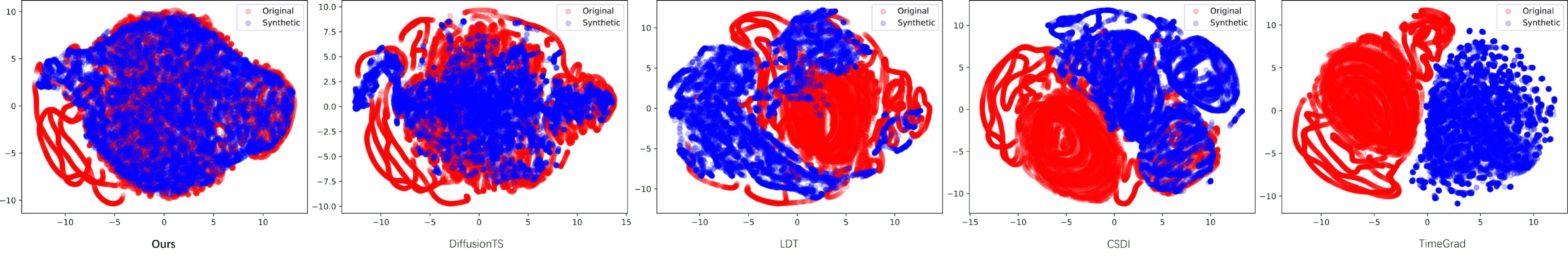}}
\caption{
Visualization of comparisons between \myformer and exsting probabilistic and deep model baselines on the ETTh (Upper) and ETTm (Bottom) dataset.
}\label{fig:vis_app_1}
\end{center}
\end{figure*}

In Figure \ref{fig:vis_app_1}, to validate the generative power of the diffusion-based probabilistic model, we visualise the generation results of \myformer and the pre-existing diffusion method on the same dataset using the t-SNE method. The red colour represents the real sequence samples and the blue colour represents the generated dataset, where the degree of aggregation of these two samples in two-dimensional space reflects the generative ability of the model. Specifically, when the projections of the two samples in 2D space are fully aggregated, the model exhibits excellent generative performance.

\begin{figure*}[t]
\begin{center}
	\centerline{\includegraphics[width=2.0\columnwidth]{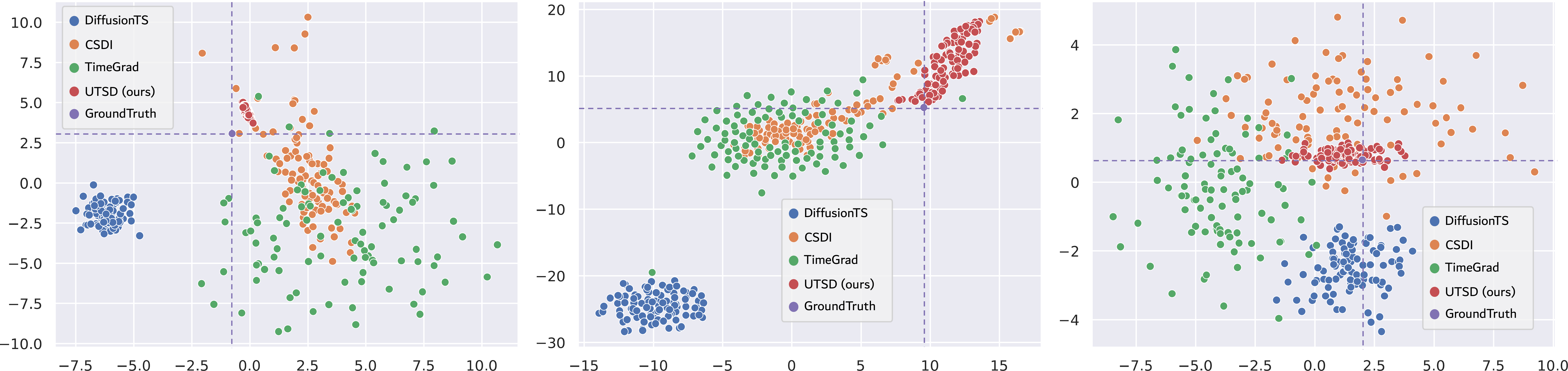}}
	\vspace{10pt}
	\centerline{\includegraphics[width=2.0\columnwidth]{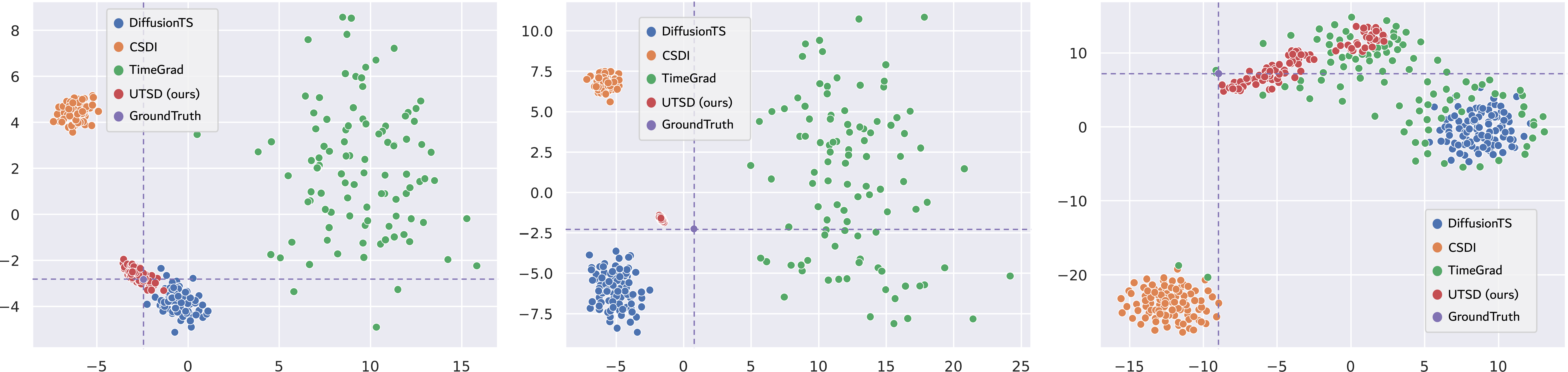}}
	\vspace{10pt}
	\centerline{\includegraphics[width=2.0\columnwidth]{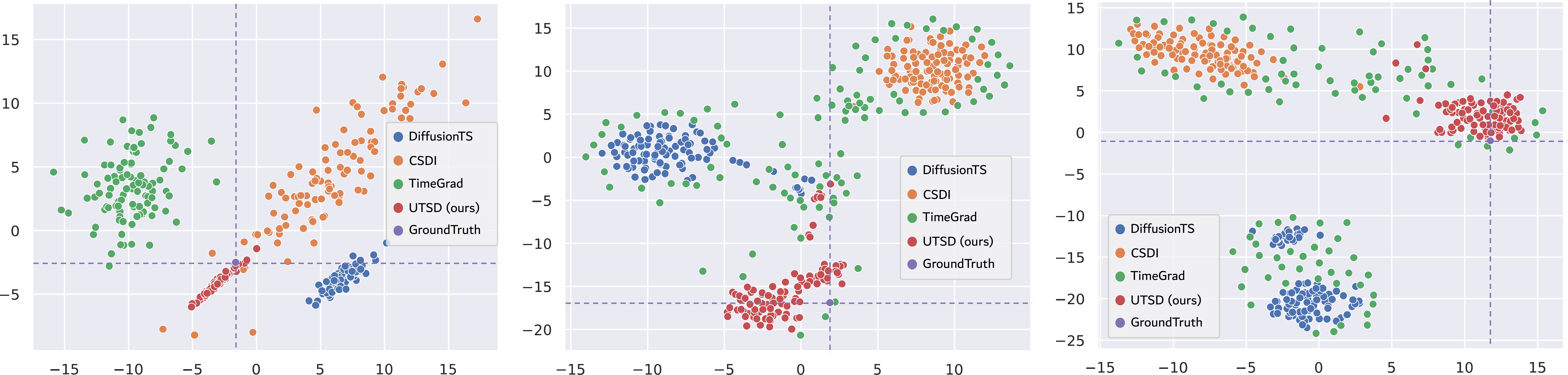}}
	\caption{
Illustration of the dithering issue with popular diffusion models for time series generation tasks, which is one of the challenges in building a unified time series generation model. With the fixed observation sequence and groundtruth, each fully trained diffusion model is repeatedly sampled 50 times. All generated sequences and the groundtruth are projected into a two-dimensional space, by the t-SNE approach. The visualisation results demonstrate the excellent stability and accuracy of the proposed method.
}\label{fig:vis_app_2}
\end{center}
\end{figure*}

The Figure \ref{fig:vis_app_2} illustrates the dithering issue of popular diffusion model on the temporal generation task, as one of the challenges in building a unified times series generation model. 
The existing CSDI~\cite{Tashiro2022csdi} and TimeGrad~\cite{Rasul2021timegrad} try to improve the prediction accuracy by averaging the results of multiple samples, however, this leads to a huge time overhead. In contrast, the proposed \myformer can generate high-quality prediction sequences with only one sampling. 
Visualization of the prediction results obtained by repeated sampling of the four probabilistic models. The same observation sequence is fixed, and each model repeats the inference 50 times. tSNE is utilized to visualize all the predicted sequences. \myformer has the highest stability and accuracy.

\begin{figure*}[t]
\begin{center}
	\centerline{\includegraphics[width=2.0\columnwidth]{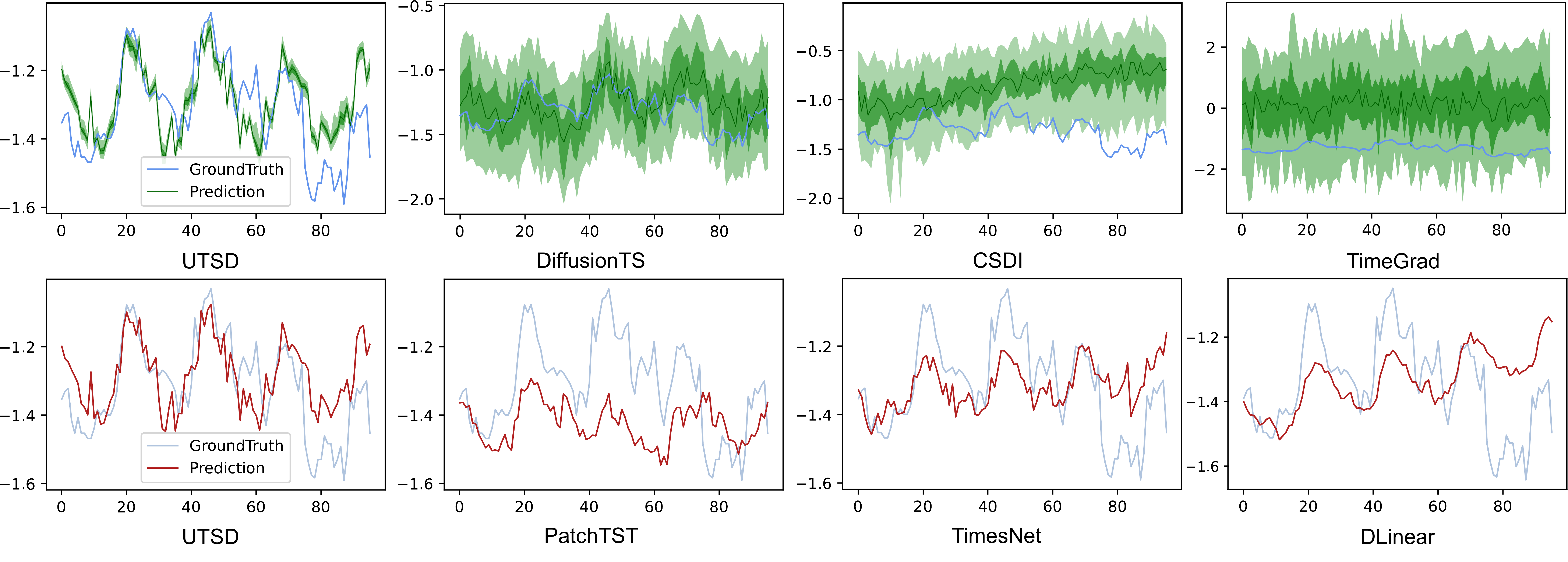}}
    \vspace{20pt}
	\centerline{\includegraphics[width=2.0\columnwidth]{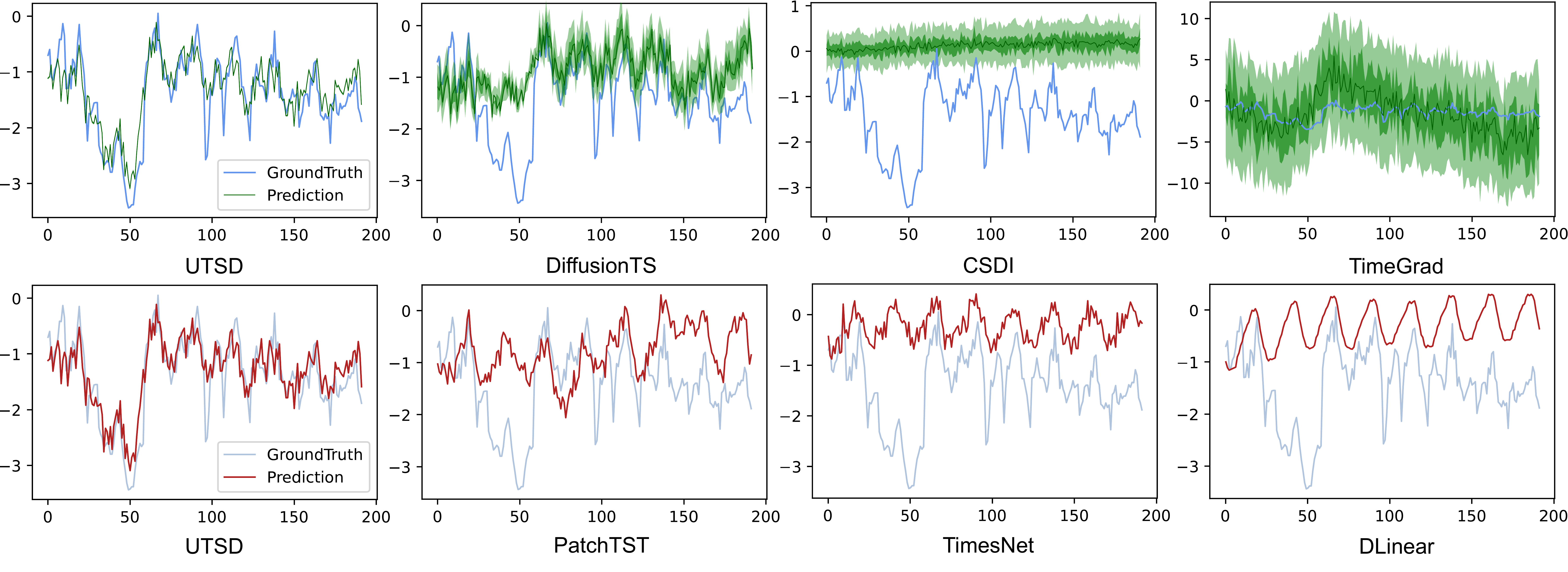}}
	\caption{
Visualization of comparisons between \myformer and exsting probabilistic (upper) and deep model (bottom) baselines on the \textbf{ETTh1} dataset. Where the light blue curve represents the groundtruth, and the green and red curves represent the prediction results of several baselines. The light and dark green staining show the predictions of the probabilistic model at the $10-90\%$ and $25-75\%$ confidence intervals, respectively.
}\label{fig:vis_app_4}
\end{center}
\vspace{-10pt}
\end{figure*}

\begin{figure*}[t]
\begin{center}
	\centerline{\includegraphics[width=2.0\columnwidth]{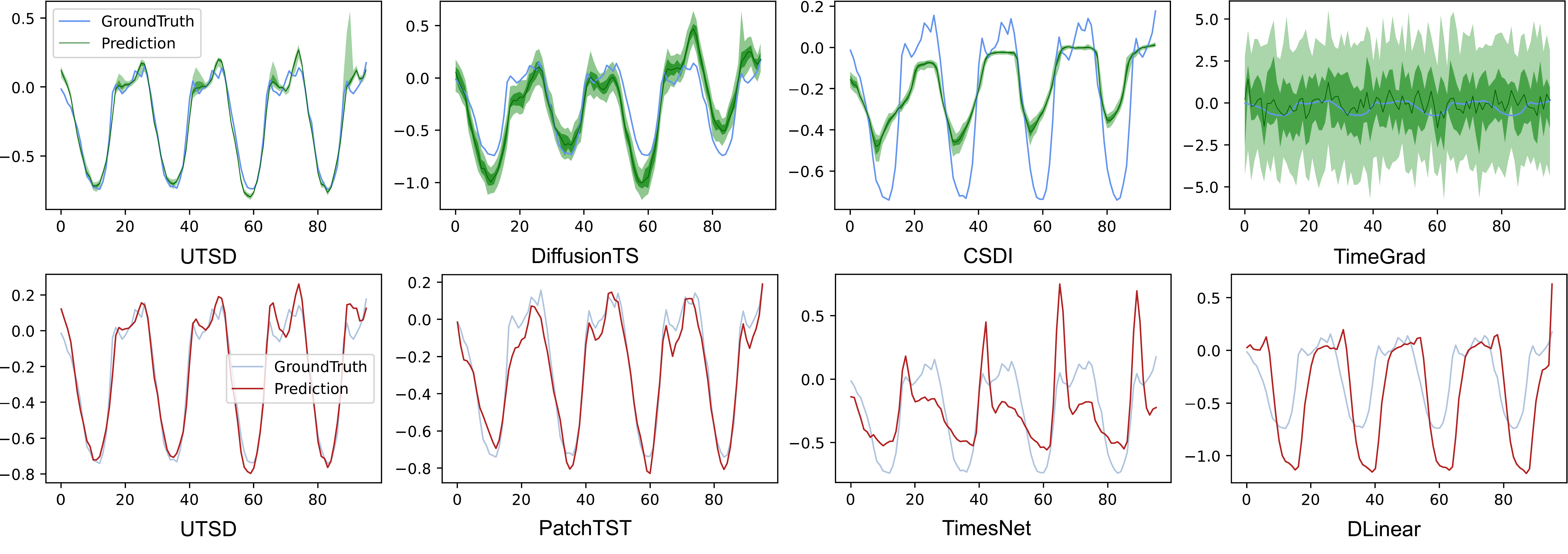}}
	\vspace{-5pt}
	\centerline{\includegraphics[width=2.0\columnwidth]{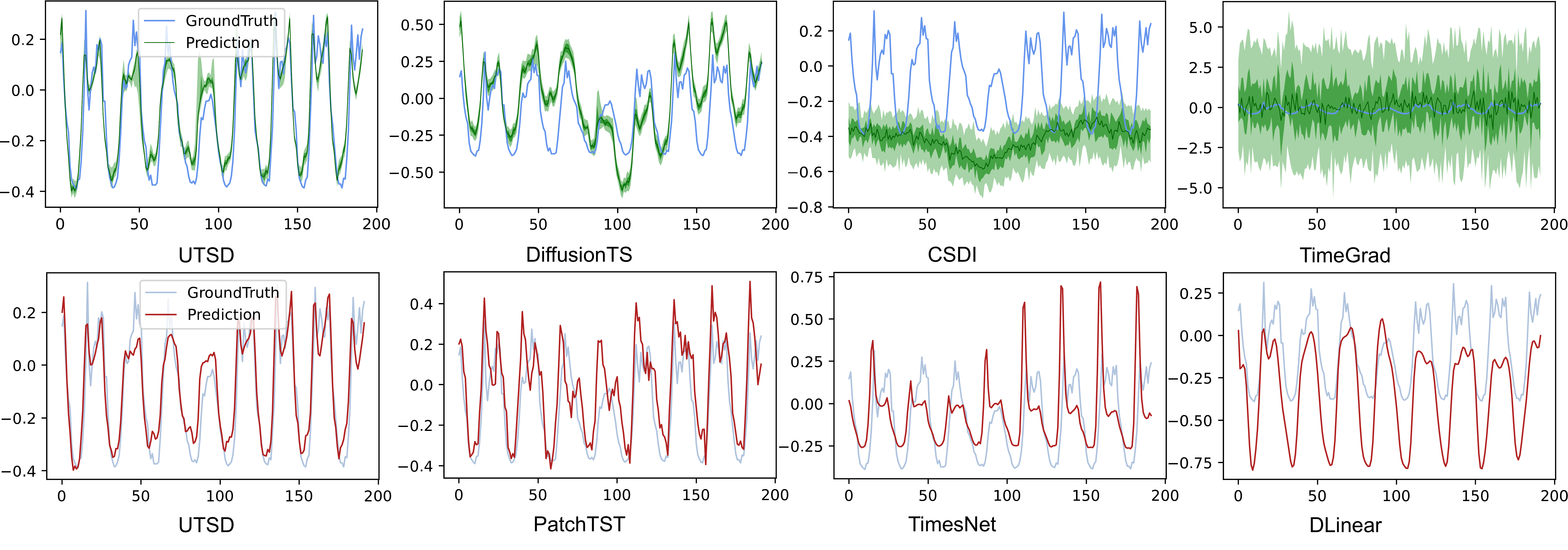}}
	\vspace{-5pt}
	\centerline{\includegraphics[width=2.0\columnwidth]{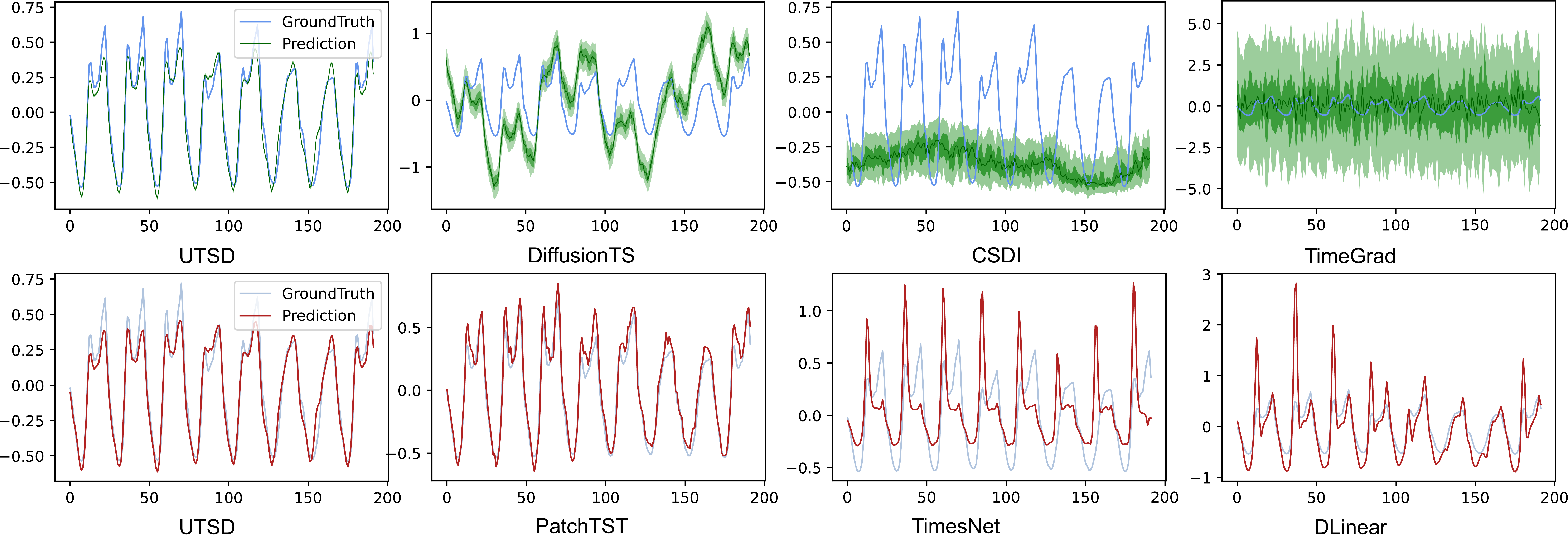}}
	\caption{
Visualization of comparisons between \myformer and exsting probabilistic (upper) and deep model (bottom) baselines on the \textbf{ECL} dataset.
}\label{fig:vis_app_9}
\end{center}
\end{figure*}

To visualise the performance of the model on the Scratch prediction task, Figures \ref{fig:vis_app_4}—\ref{fig:vis_app_9} show the results on several datasets. 
Specifically, the upper half shows the prediction results obtained from the probabilistic model with 50 repetitive samples, where the light and dark green colours denote the prediction results for the $10-90\%$ and $25-75\%$ confidence intervals, respectively (50 repetitive samples for each model), and the blue and green curves denote the true value and median prediction results, respectively. 
The lower half of the display shows shows the prediction results of the deep regression model (\myformer samples only once), where the blue and red curves indicate the ground truth and prediction results, respectively.

\begin{figure*}
\begin{center}
\centerline{\includegraphics[width=2.0\columnwidth]{images/show_truth_predition_tsne_ecl_1.pdf}}
\centerline{\includegraphics[width=2.0\columnwidth]{images/show_truth_predition_tsne_ecl_2.pdf}}
\centerline{\includegraphics[width=2.0\columnwidth]{images/show_truth_predition_tsne_ecl_3.pdf}}
\centerline{\includegraphics[width=2.0\columnwidth]{images/show_truth_predition_tsne_etth1_1.pdf}}
\caption{
Visualization of comparisons between \myformer and baselines on the Electricity (subgraph 1-3) and Traffic (subgraph 4) dataset. The proposed TimeControl achieves the best visualization performance.
}\label{fig:vis_1}
\end{center}
\end{figure*}

\begin{figure*}
\begin{center}
	\centerline{\includegraphics[width=2.0\columnwidth]{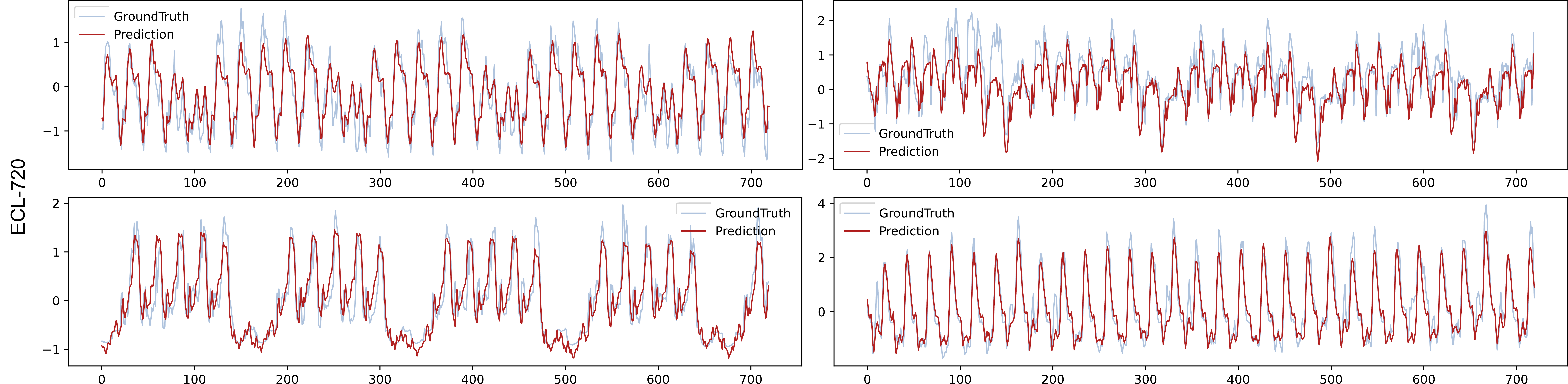}}
    \vspace{20pt}
	\centerline{\includegraphics[width=2.0\columnwidth]{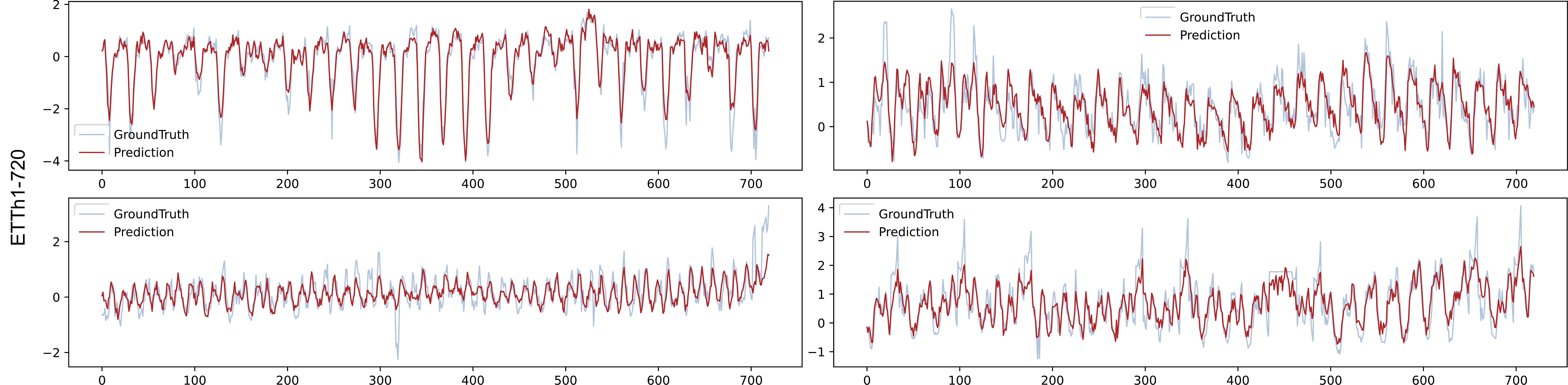}}
	\vspace{20pt}
	\centerline{\includegraphics[width=2.0\columnwidth]{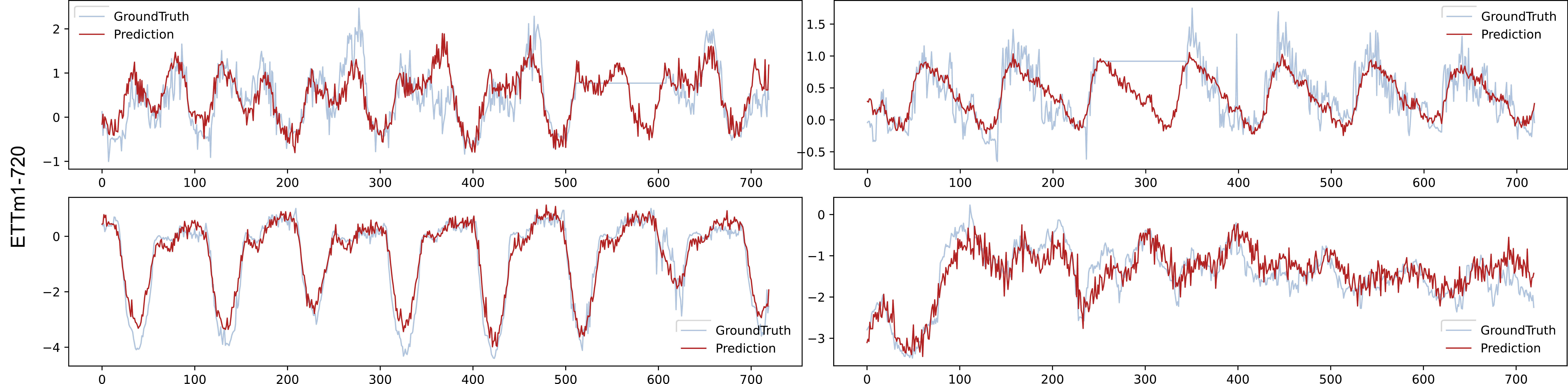}}
	\vspace{20pt}
	\centerline{\includegraphics[width=2.0\columnwidth]{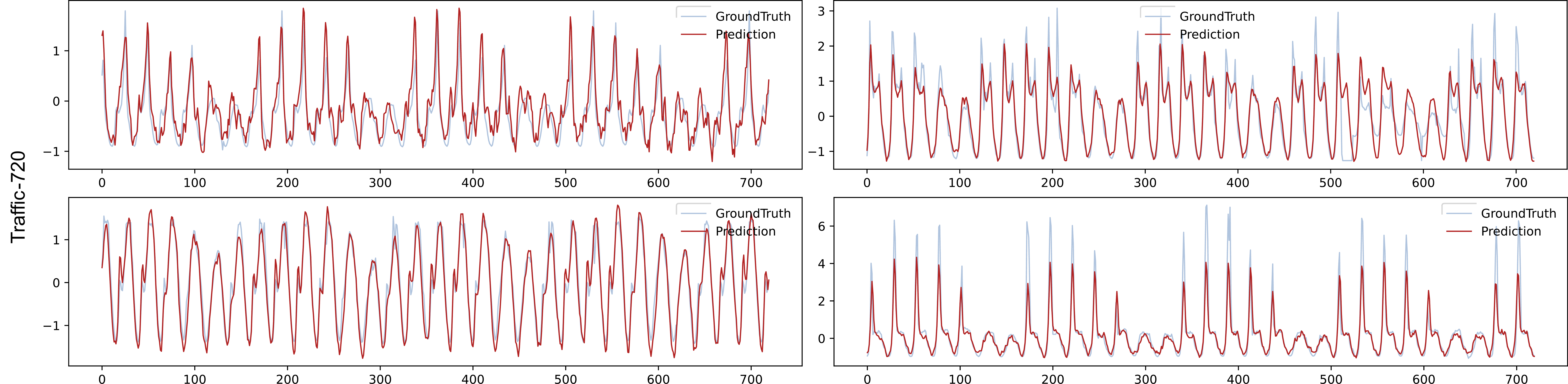}}
	\caption{
Demonstration of \myformer prediction results on real \textbf{long-term multi-periodic} sequences sampled from ETT, ECL and Traffic datasets.
}\label{fig:vis_app_11}
\end{center}
\end{figure*}

\begin{figure*}
\begin{center}
	\centerline{\includegraphics[width=2.0\columnwidth]{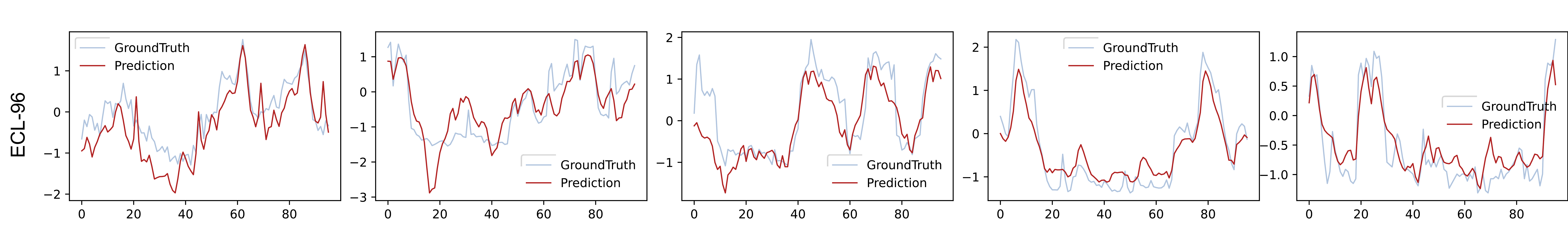}}
	\vspace{0pt}
	\centerline{\includegraphics[width=2.0\columnwidth]{images/show_96_720_etth1_96.pdf}}
	\vspace{0pt}
	\centerline{\includegraphics[width=2.0\columnwidth]{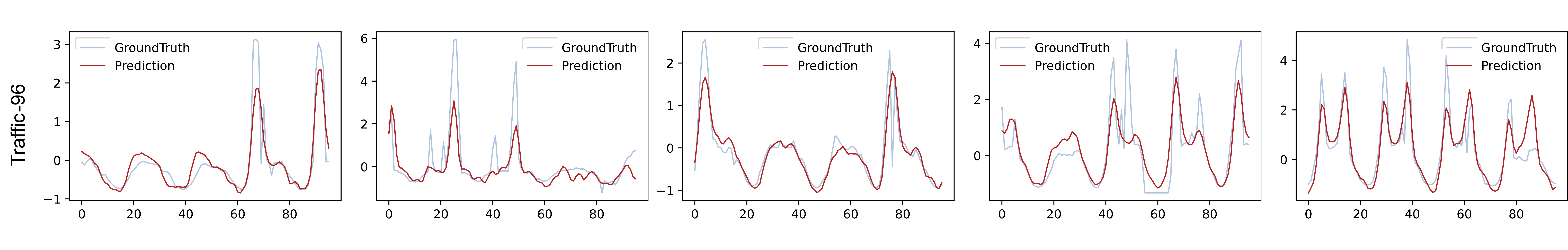}}
	\caption{
Demonstration of \myformer prediction results on real \textbf{short-term non-periodic} sequences sampled from ETT, ECL and Traffic datasets. The blue and red lines respectively represent the actual values and predicted values of the future time series.
}\label{fig:vis_app_10}
\vspace{-20pt}
\end{center}
\end{figure*}

\begin{figure*}
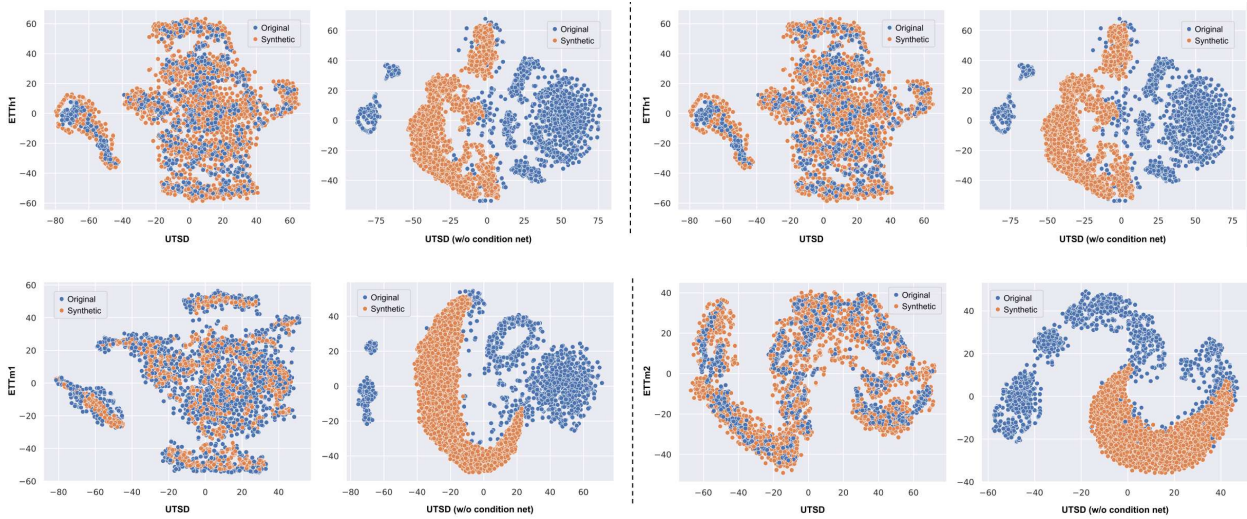

\begin{center}
\centerline{\includegraphics[width=2.0\columnwidth]{images/show_ablation_ETTh.pdf}}
\centerline{\includegraphics[width=2.0\columnwidth]{images/show_ablation_ETTm.pdf}}
\caption{
Visualisation on the validity of the proposed model architecture condition net. Where the upper part shows the visualisation results on the ETTh1 and ETTh2 datasets, and the bottom part shows the visualisation results on the ETTm1 and ETTm2 datasets.
}\label{fig:vis_app_3}
\vspace{-20pt}
\end{center}
\end{figure*}


\begin{figure*}[t]
\begin{center}
	\centerline{\includegraphics[width=2.0\columnwidth]{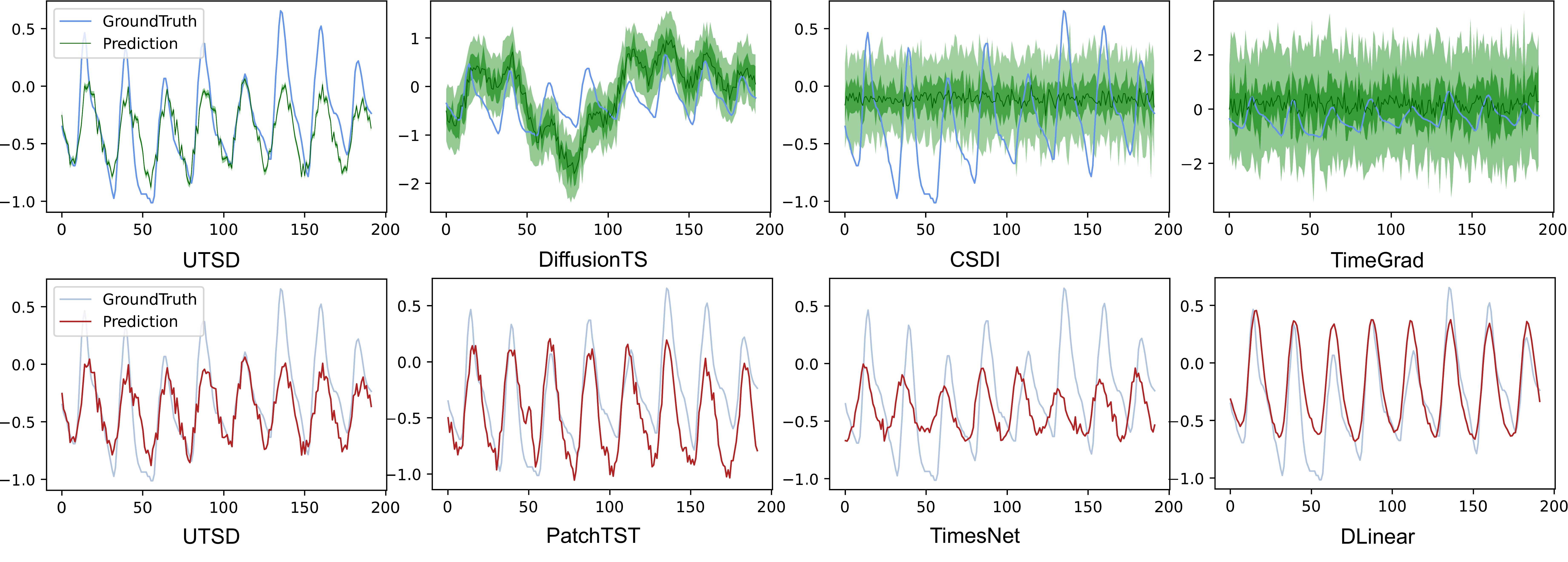}}
    \vspace{20pt}
	\centerline{\includegraphics[width=2.0\columnwidth]{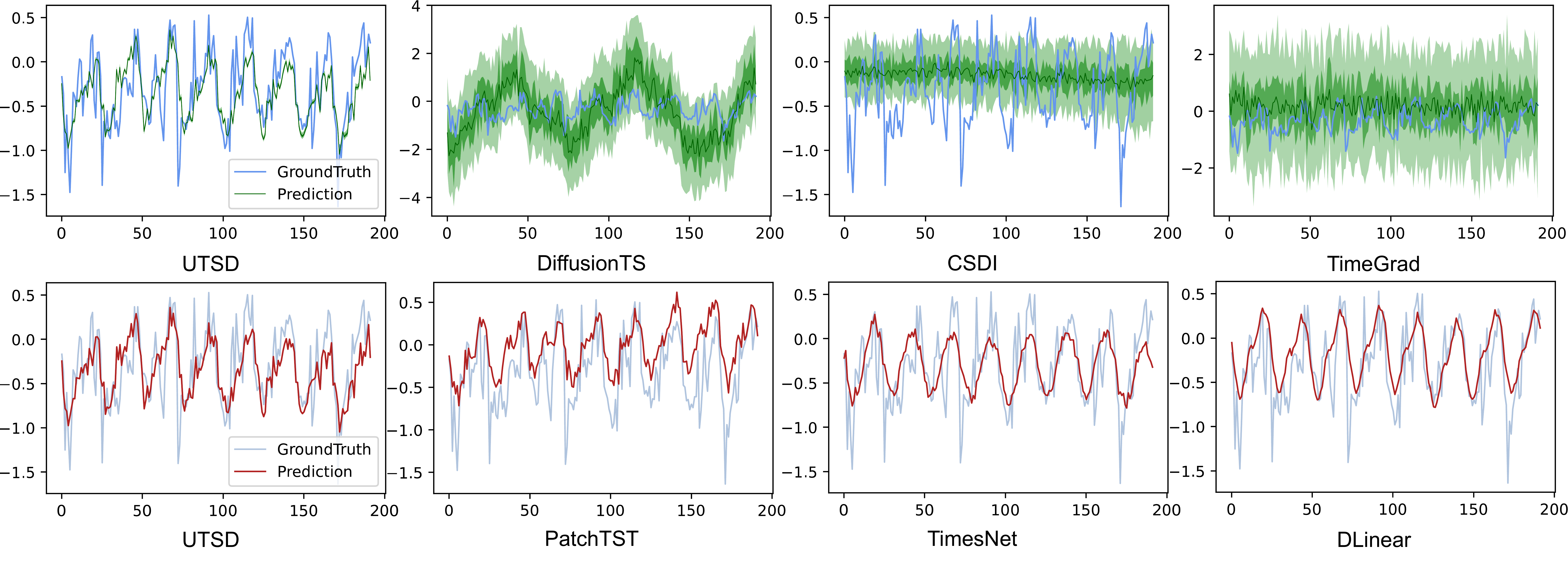}}
	\caption{
Visualization of comparisons between \myformer and exsting probabilistic (upper) and deep model (bottom) baselines on the \textbf{ETTh2} dataset. Where the light blue curve represents the groundtruth, and the green and red curves represent the prediction results of several baselines. The light and dark green staining show the predictions of the probabilistic model at the $10-90\%$ and $25-75\%$ confidence intervals, respectively.
}\label{fig:vis_app_5}
\end{center}
\end{figure*}

\begin{figure*}[t]
\begin{center}
	\centerline{\includegraphics[width=2.0\columnwidth]{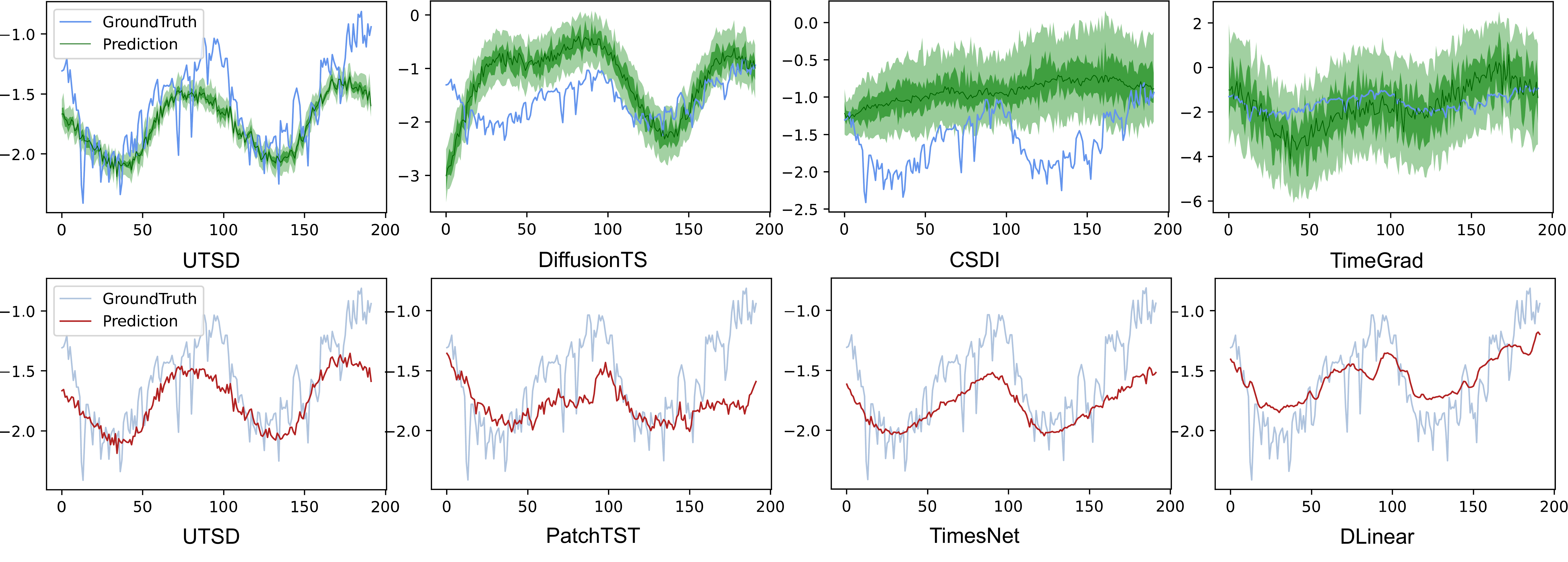}}
    \vspace{20pt}
	\centerline{\includegraphics[width=2.0\columnwidth]{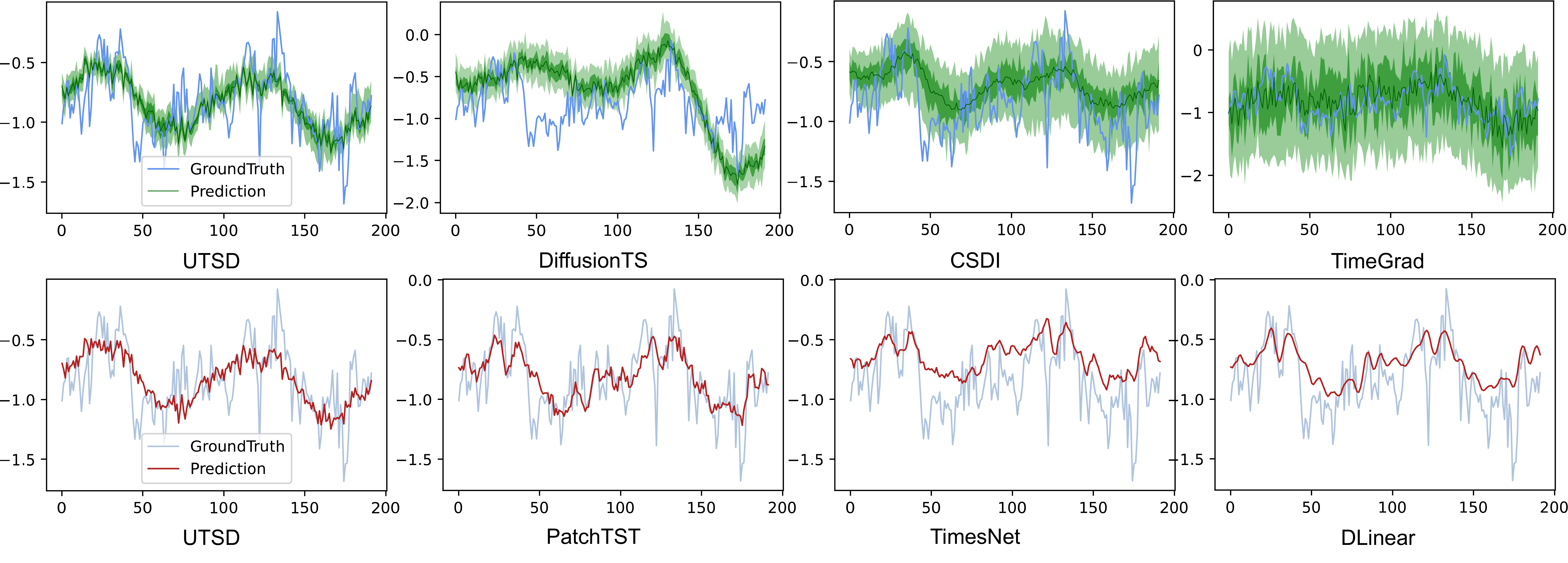}}
    \vspace{20pt}
	\centerline{\includegraphics[width=2.0\columnwidth]{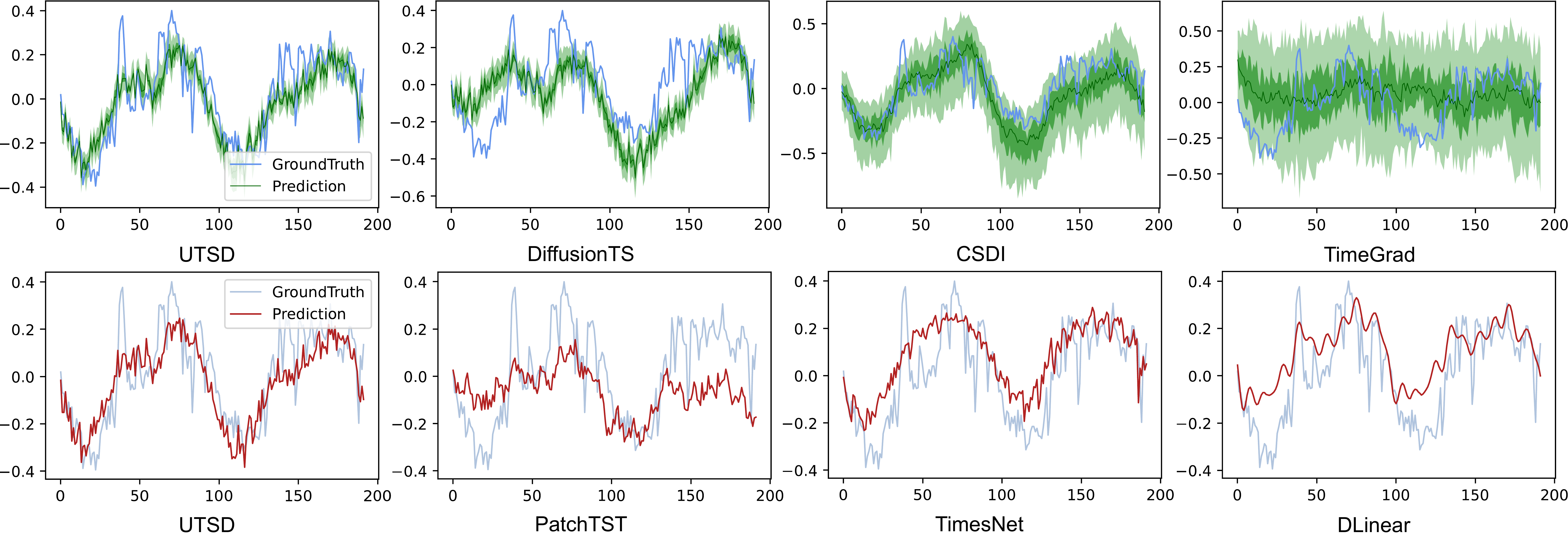}}
	\caption{
Visualization of comparisons between \myformer and exsting probabilistic (upper) and deep model (bottom) baselines on the \textbf{ETTm1} and \textbf{ETTm2} dataset. Where the light blue curve represents the groundtruth, and the green and red curves represent the prediction results of several baselines. The light and dark green staining show the predictions of the probabilistic model at the $10-90\%$ and $25-75\%$ confidence intervals, respectively.
}\label{fig:vis_app_6}
\end{center}
\end{figure*}

\begin{figure*}[t]
\begin{center}
	\centerline{\includegraphics[width=2.0\columnwidth]{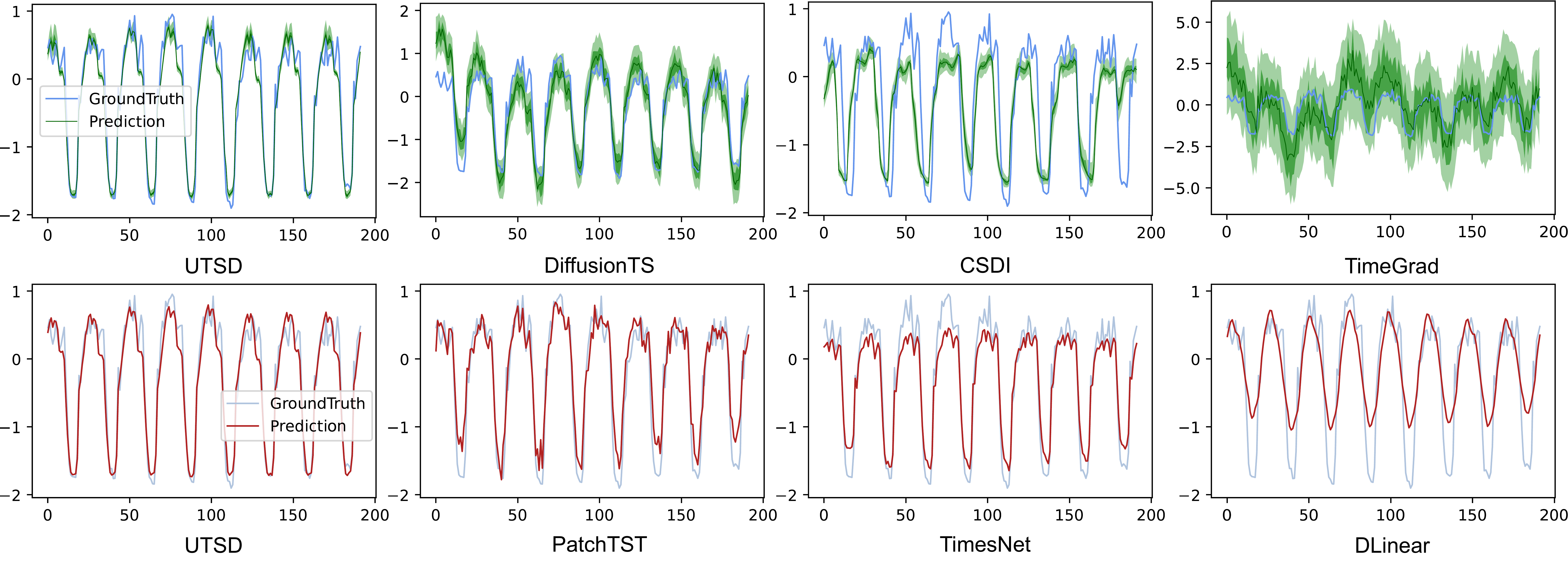}}
    \vspace{5pt}
	\centerline{\includegraphics[width=2.0\columnwidth]{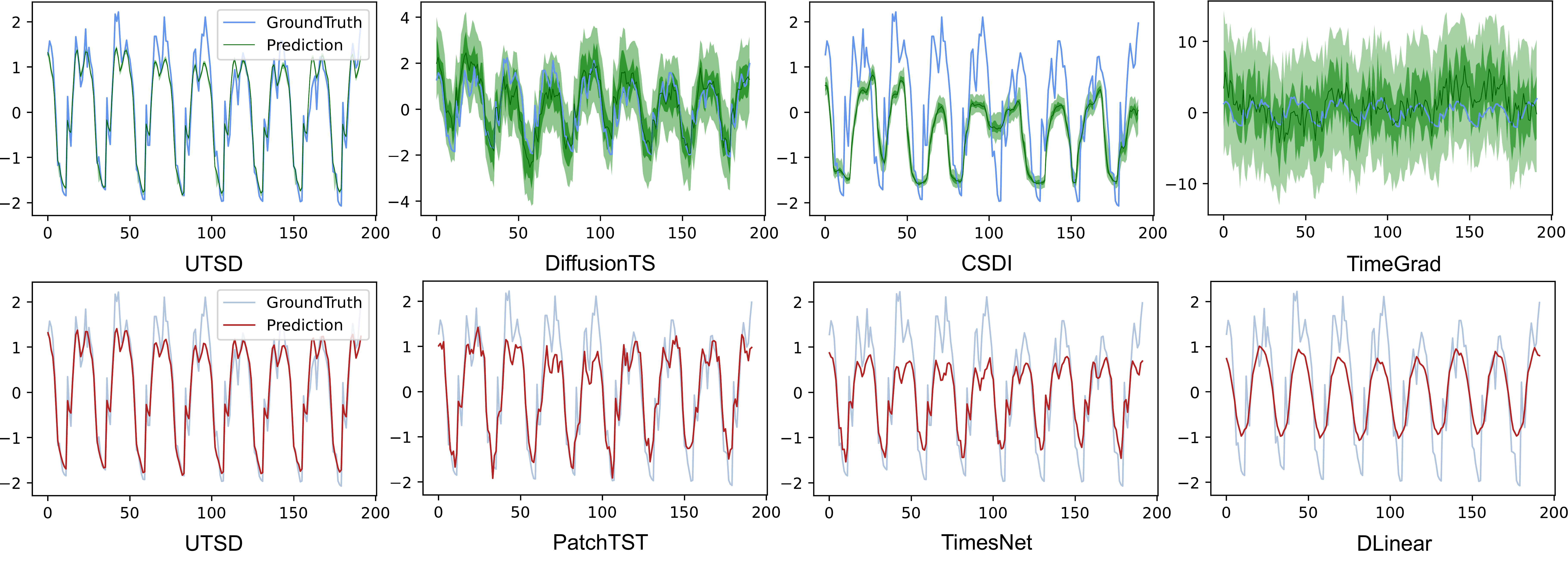}}
    \vspace{5pt}
	\centerline{\includegraphics[width=2.0\columnwidth]{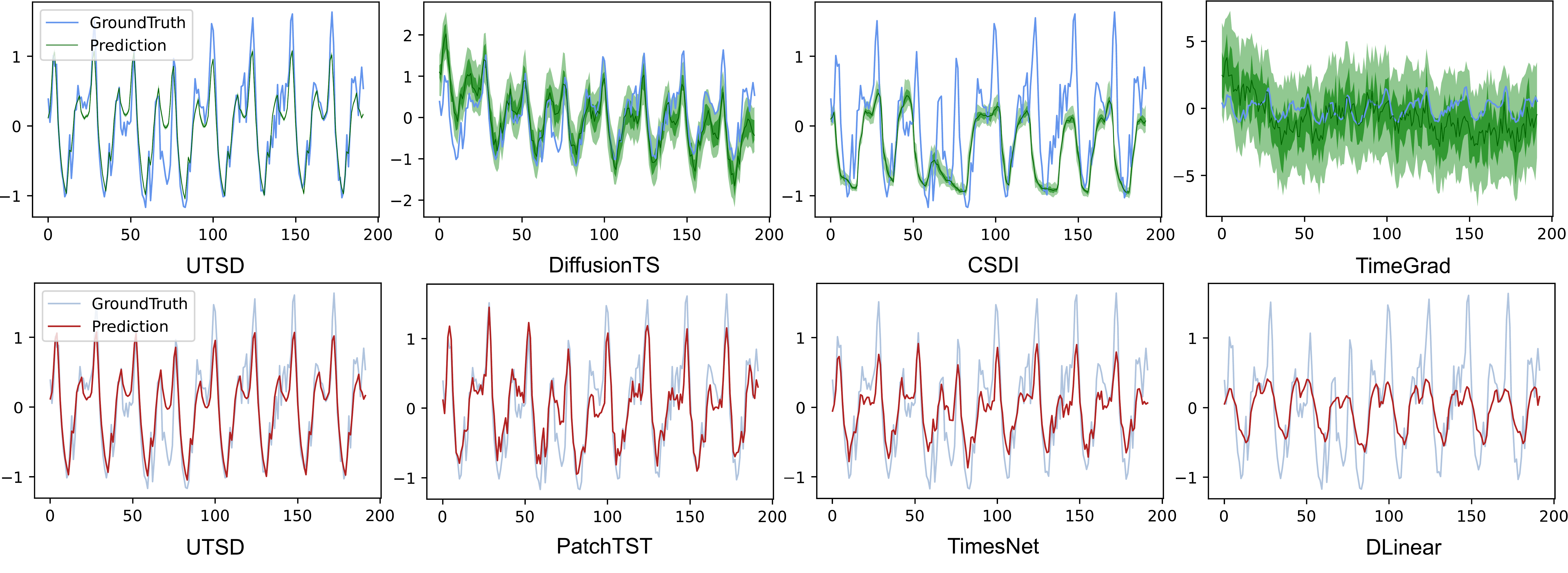}}
	\caption{
Visualization of comparisons between \myformer and exsting probabilistic (upper) and deep model (bottom) baselines on the \textbf{ECL} dataset.
}\label{fig:vis_app_7}
\end{center}
\end{figure*}

\begin{figure*}[t]
\begin{center}
	\centerline{\includegraphics[width=2.0\columnwidth]{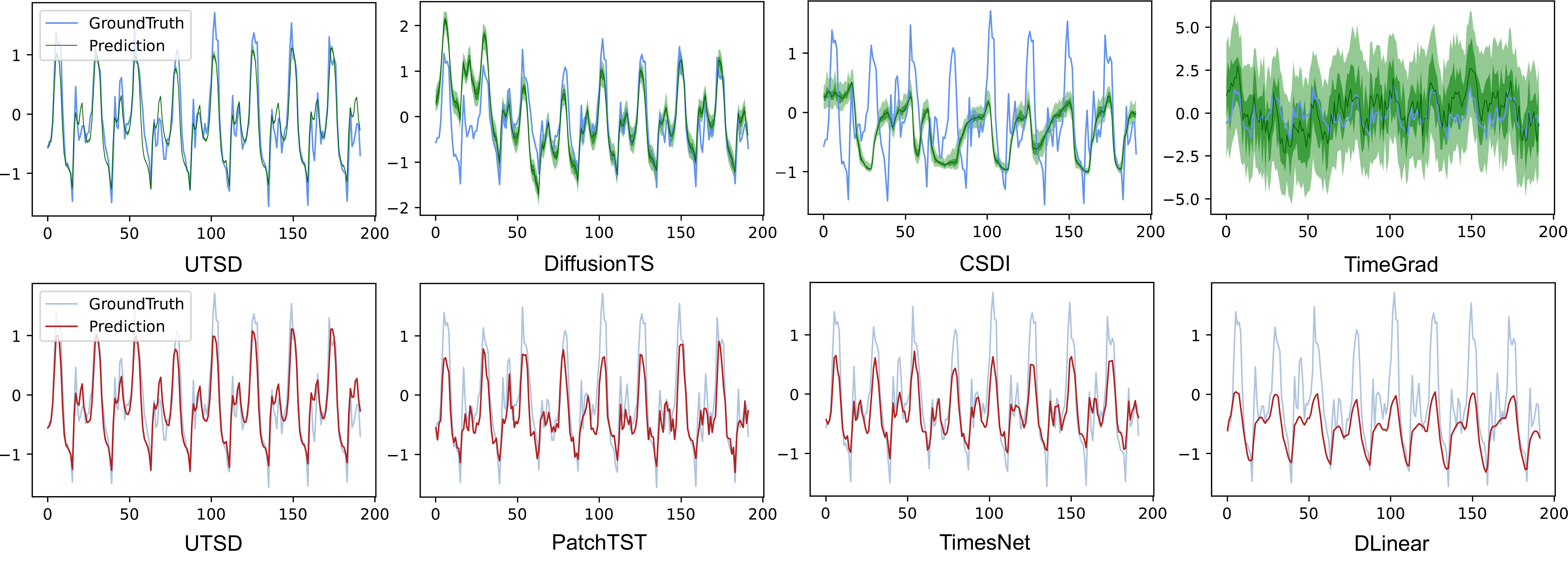}}
    \vspace{5pt}
	\centerline{\includegraphics[width=2.0\columnwidth]{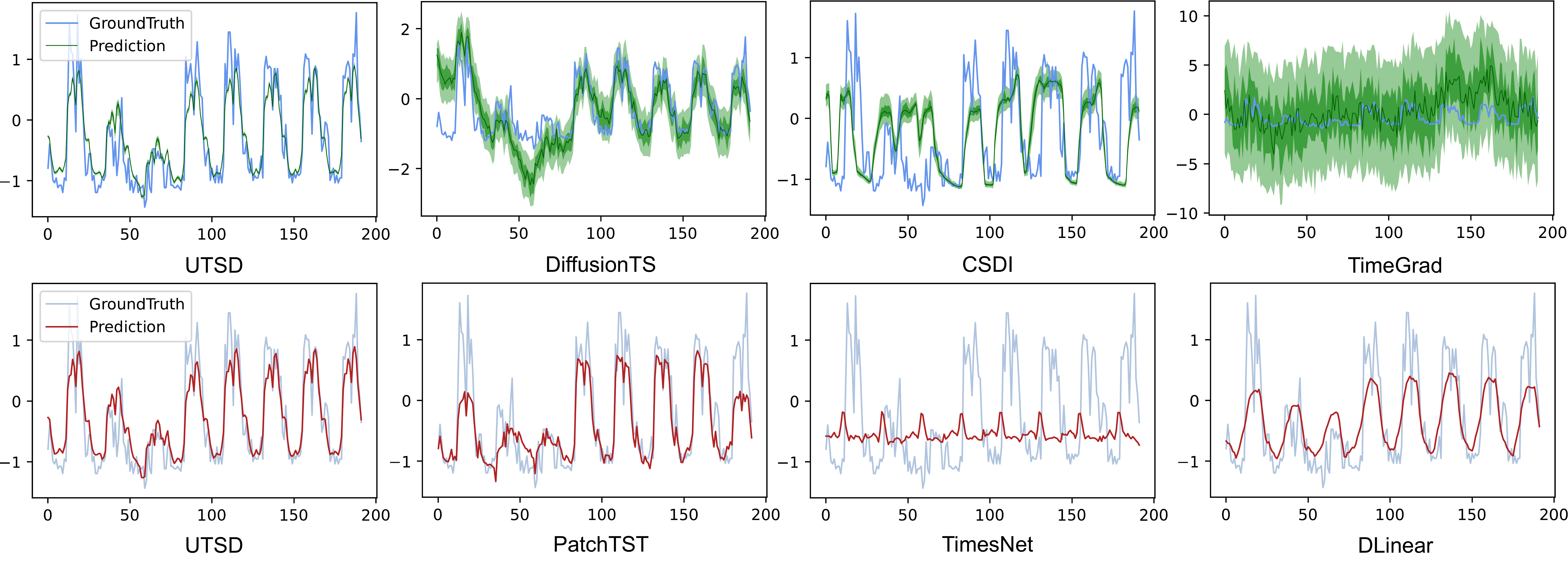}}
    \vspace{5pt}
	\centerline{\includegraphics[width=2.0\columnwidth]{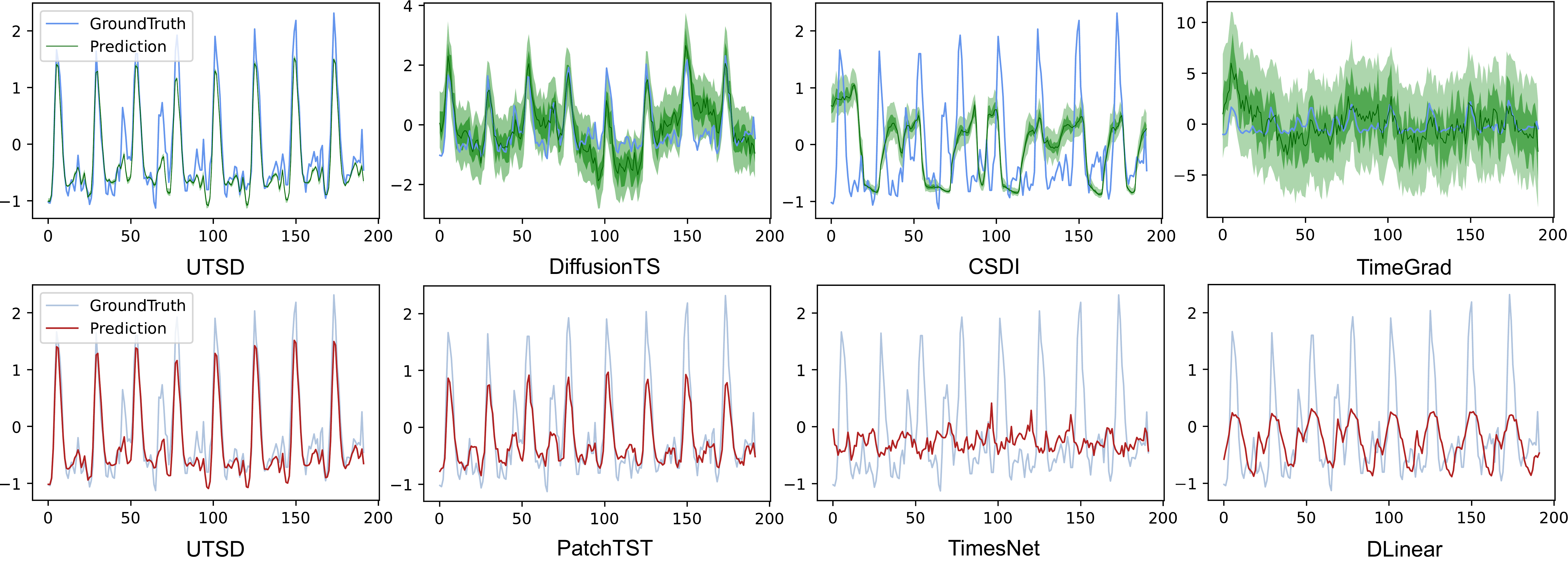}}
	\caption{
Visualization of comparisons between \myformer and exsting probabilistic (upper) and deep model (bottom) baselines on the \textbf{ECL} dataset.
}\label{fig:vis_app_8}
\end{center}
\end{figure*}

%% file: tables_brief/dataset.tex
\begin{table*}[thbp]
\caption{
All the datasets required for time series forecasting. 
We divide these datasets according to the usage during the training and evaluation of \myformer. 
The data of \textbf{Pretraining} is only used for training model. 
In the zero-shot TSF scenario, \textbf{Evaluation}'s test set will be utilize to evaluate the model. In the full-shot TSF, we freeze the dual-autoencoder and quantizer structures and use the \textbf{Evaluation}'s train set to fine-tune the alignment model to align the observed time series to the landscape paintings. 
Finally, the \textbf{Evaluation}'s test set is used to evaluate the model.
}\label{tab:dataset}
\vskip 0.05in
\centering
\resizebox{1.3\columnwidth}{!}{
\begin{threeparttable}
\begin{small}
\renewcommand{\multirowsetup}{\centering}
\setlength{\tabcolsep}{3.8pt}
\begin{tabular}{ccccccc}
\toprule

\multirow{2}{*}{\rotatebox{0}{Dataset}} & 
\multirow{2}{*}{\rotatebox{0}{Domain}} & 
\multirow{2}{*}{\rotatebox{0}{Frequency}} & 
\multirow{2}{*}{\rotatebox{0}{Series Number}} & 
\multicolumn{3}{c}{Series Length} \\
& & & & min & avg & max \\ 
\midrule

Pretraining & & & & & & \\ 
\midrule

Brazilian Cities Temperature & nature & M & 12 & 492 & 757 & 1320 \\ 
Mexico City Bikes & transport & 1H & 494 & 780 & 78313 & 104449 \\ 
Solar (5 Min.) & energy & 5min & 5166 & 105120 & 105120 & 105120 \\ 
Solar (Hourly) & energy & 1H & 5166 & 8760 & 8760 & 8760 \\ 
Spanish Energy and Weather & energy & 1H & 66 & 35064 & 35064 & 35064 \\ 
Taxi (Hourly) & transport & 1H & 2428 & 734 & 739 & 744 \\ 
USHCN & nature & 1D & 6090 & 5906 & 38653 & 59283 \\ 
Weatherbench (Daily) & nature & 1D & 225280 & 14609 & 14609 & 14610 \\ 
Weatherbench (Hourly) & nature & 1H & 225280 & 350633 & 350639 & 350640 \\ 
Weatherbench (Weekly) & nature & 1W & 225280 & 2087 & 2087 & 2087 \\ 
Wiki Daily (100k) & web & 1D & 100000 & 2741 & 2741 & 2741 \\ 
Wind Farms (Daily) & energy & 1D & 337 & 71 & 354 & 366 \\ 
Wind Farms (Hourly) & energy & 1H & 337 & 1715 & 8514 & 8784 \\ 
Electricity (15 Min.) & energy & 15min & 370 & 16032 & 113341 & 140256 \\ 
Electricity (Hourly) & energy & 1H & 321 & 26304 & 26304 & 26304 \\ 
Electricity (Weekly) & energy & 1W & 321 & 156 & 156 & 156 \\ 
KDD Cup 2018 & nature & 1H & 270 & 9504 & 10897 & 10920 \\ 
M4 (Daily) & various & 1D & 4227 & 107 & 2371 & 9933 \\ 
M4 (Hourly) & various & 1H & 414 & 748 & 901 & 1008 \\ 
M4 (Monthly) & various & 1M & 48000 & 60 & 234 & 2812 \\ 
M4 (Weekly) & various & 1W & 359 & 93 & 1035 & 2610 \\ 
Pedestrian Counts & transport & 1H & 66 & 576 & 47459 & 96424 \\ 
Rideshare & transport & 1H & 2340 & 541 & 541 & 541 \\ 
Taxi (30 Min.) & transport & 30min & 2428 & 1469 & 1478 & 1488 \\ 
Temperature-Rain & nature & 1D & 32072 & 725 & 725 & 725 \\ 
Uber TLC (Daily) & transport & 1D & 262 & 181 & 181 & 181 \\ 
Uber TLC (Hourly) & transport & 1H & 262 & 4344 & 4344 & 4344 \\ 

Australian Electricity & energy & 30min & 5 & 230736 & 231052 & 232272 \\ 
CIF 2016 & banking & 1M & 72 & 28 & 98 & 120 \\ 
Car Parts & retail & 1M & 2674 & 51 & 51 & 51 \\ 
Covid Deaths & healthcare & 1D & 266 & 212 & 212 & 212 \\ 
Dominick & retail & 1D & 100014 & 201 & 296 & 399 \\ 
ERCOT Load & energy & 1H & 8 & 154854 & 154854 & 154854 \\ 
FRED-MD & economics & 1M & 107 & 728 & 728 & 728 \\ 
Hospital & healthcare & 1M & 767 & 84 & 84 & 84 \\ 
Tourism (Monthly) & various & 1M & 366 & 91 & 298 & 333 \\ 
Tourism (Quarterly) & various & 1Q & 427 & 30 & 99 & 130 \\ 
Tourism (Yearly) & various & 1Y & 518 & 11 & 24 & 47 \\ 
\midrule

Evaluation & & & & & & \\ 
\midrule
ETTm1 (15 Min.) & energy & 15min & 14 & 69680 & 69680 & 69680 \\ 
ETTm2 (15 Min.) & energy & 15min & 14 & 69680 & 69680 & 69680 \\ 
ETTh1 (Hourly) & energy & 1H & 14 & 17420 & 17420 & 17420 \\ 
ETTh2 (Hourly) & energy & 1H & 14 & 17420 & 17420 & 17420 \\ 
Electricity & Electricity & 1H & 321 & 26304 & 26304 & 26304 \\
Exchange Rate & finance & 1B & 8 & 7588 & 7588 & 7588 \\ 
Traffic & transport & 1H & 862 & 17544 & 17544 & 17544 \\ 
Weather & nature & 1D & 3010 & 1332 & 14296 & 65981 \\ 
Solar & Energy & 10M & 137 & 52560 & 52560 & 52560 \\
ILI & Health & 1W & 7 & 966 & 966 & 966 \\
Energy (London Meters) & energy & 30min & 5560 & 288 & 29951 & 39648 \\ 

\bottomrule
\end{tabular}
\end{small}
\end{threeparttable}
}
\end{table*}

%% file: tables_brief/forecasting_full_all.tex
\begin{table*}[htbp]
\centering
\small
\tabcolsep=0.15cm
\renewcommand\arraystretch{1.3}
\caption{
Comparison of the performance from diverse prediction lengths on \textbf{full-shot and zero-shot time series forecasting}.
Specifically, the left side of double-bar graph shows the zero-shot performance comparison between TimeControl and existing foundation models. The right side of double-bar graph shows the full-shot performance comparison between TimeControl and existing proprietary models. We boldface the best performance on two scenarios, respectively. Among them, the TimeControl demonstrated the best performance in most of the prediction scenarios.
}\label{tab:forecasting_deep_all}
\vskip 0.05in
\resizebox{\textwidth}{!}{
\begin{tabular}{cccc|cc|cc|cc||cc|cc|cc|cc|cc|cc|cc|cc}
\toprule
\hline
\multicolumn{2}{c}{\multirow{5}{*}{Model}} & 
\multicolumn{8}{c||}{\textbf{\textit{Domain-fused pre-training}}} & 
\multicolumn{8}{c|}{\textbf{\textit{Domain-fused pre-training}}} & 
\multicolumn{8}{c}{\textbf{\textit{Models Trained From Scratch}}} \\ 

\multicolumn{2}{c}{} & 
\multicolumn{8}{c||}{\textbf{\textit{(zero-shot without fine-tuning)}}} & 
\multicolumn{8}{c|}{\textbf{\textit{(fine-tuning on target dataset)}}} & 
\multicolumn{8}{c}{\textbf{\textit{(full-shot)}}} \\ 

\cline{3-26} 

\multicolumn{2}{c}{} & 
\multicolumn{2}{c}{\textbf{\myformer}} & 
\multicolumn{2}{c}{Moirai} & 
\multicolumn{2}{c}{UniTime} & 
\multicolumn{2}{c||}{GPT4TS} & 
\multicolumn{2}{c}{\textbf{\myformer}} & 
\multicolumn{2}{c}{TimeLLM} & 
\multicolumn{2}{c}{LLM4TS} & 
\multicolumn{2}{c|}{GPT4TS} & 
\multicolumn{2}{c}{DUET} & 
\multicolumn{2}{c}{PDF} & 
\multicolumn{2}{c}{Pathformer} & 
\multicolumn{2}{c}{TimeMixer} \\ 

\multicolumn{2}{c}{} & 
\multicolumn{2}{c}{\textbf{(Ours)}} & 
\multicolumn{2}{c}{\cite{woo2024moirai}} & 
\multicolumn{2}{c}{\cite{liu2024unitime}} & 
\multicolumn{2}{c||}{\cite{Zhou2023OneFA}} & 
\multicolumn{2}{c}{\textbf{(Ours)}} & 
\multicolumn{2}{c}{\cite{Jin2023TimeLLMTS}} & 
\multicolumn{2}{c}{\cite{Chang2023LLM4TSAP}} & 
\multicolumn{2}{c|}{\cite{Zhou2023OneFA}} & 
\multicolumn{2}{c}{\cite{qiu2025duet}} & 
\multicolumn{2}{c}{\cite{dai2024pdf}} & 
\multicolumn{2}{c}{\cite{chen2024pathformer}} & 
\multicolumn{2}{c}{\cite{wang2024timemixer}} \\
    
    \cline{3-26} 
    
    \multicolumn{2}{c}{} & 
    MSE & MAE & MSE & MAE & MSE & MAE & MSE & MAE & 
    MSE & MAE & MSE & MAE & MSE & MAE & MSE & MAE & MSE & MAE & MSE & MAE & MSE & MAE & MSE & MAE \\
    
    \hline 
    
    \multicolumn{1}{c}{\multirow{5}{*}{\rotatebox{90}{ETTm1}}} & 96 & 
    0.337 & \textbf{0.358} &
    0.404 & 0.383 &
    \textbf{0.322} & 0.363 &
    0.509 & 0.463 &
    0.279 & \textbf{0.313} &
    \textbf{0.272} & 0.334 &
    0.285 & 0.343 &
    0.292 & 0.346 &
    0.279 & 0.333 &
    0.286 & 0.340 &
    0.290 & 0.335 &
    0.293 & 0.345 \\
    
    \multicolumn{1}{c}{} & 192 & 
    \textbf{0.334} & \textbf{0.367} &
    0.435 & 0.402 &
    0.366 & 0.387 &
    0.537 & 0.476 &
    \textbf{0.284} & \textbf{0.338} &
    0.310 & 0.358 &
    0.324 & 0.366 &
    0.332 & 0.372 &
    0.320 & 0.358 &
    0.321 & 0.364 &
    0.337 & 0.363 &
    0.335 & 0.372 \\
    
    \multicolumn{1}{c}{} & 336 & 
    \textbf{0.343} & \textbf{0.358} &
    0.462 & 0.416 &
    0.398 & 0.407 &
    0.564 & 0.488 &
    \textbf{0.292} & \textbf{0.345} &
    0.352 & 0.384 &
    0.353 & 0.385 &
    0.366 & 0.394 &
    0.348 & 0.377 &
    0.354 & 0.383 &
    0.374 & 0.384 &
    0.368 & 0.386 \\
    
    \multicolumn{1}{c}{} & 720 & 
    \textbf{0.350} & \textbf{0.383} &
    0.490 & 0.437 &
    0.454 & 0.440 &
    0.592 & 0.504 &
    \textbf{0.297} & \textbf{0.348} &
    0.383 & 0.411 &
    0.408 & 0.419 &
    0.417 & 0.421 &
    0.405 & 0.408 &
    0.408 & 0.415 &
    0.428 & 0.416 &
    0.426 & 0.417 \\
    
    \cline{2-26} 
    
    \multicolumn{1}{c}{} & Avg. & 
    \textbf{0.341} & \textbf{0.369} &
    0.448 & 0.410 &
    0.385 & 0.399 &
    0.551 & 0.483 &
    \textbf{0.288} & \textbf{0.336} &
    0.329 & 0.372 &
    0.343 & 0.378 &
    0.352 & 0.383 &
    0.338 & 0.369 &
    0.342 & 0.376 &
    0.357 & 0.374 &
    0.355 & 0.380 \\
    
    \hline 
    
    \multicolumn{1}{c}{\multirow{5}{*}{\rotatebox{90}{ETTm2}}} & 96 & 
    0.185 & 0.279 &
    0.205 & 0.282 &
    \textbf{0.183} & \textbf{0.266} &
    0.229 & 0.304 &
    0.181 & 0.274 &
    \textbf{0.161} & 0.253 &
    0.165 & 0.254 &
    0.173 & 0.262 &
    0.161 & \textbf{0.248} &
    0.163 & 0.251 &
    0.164 & 0.250 &
    0.165 & 0.256 \\
    
    \multicolumn{1}{c}{} & 192 & 
    \textbf{0.231} & 0.312 &
    0.261 & 0.318 &
    0.251 & \textbf{0.310} &
    0.287 & 0.338 &
    \textbf{0.211} & 0.296 &
    0.219 & 0.293 &
    0.220 & 0.292 &
    0.229 & 0.301 &
    0.214 & \textbf{0.286} &
    0.219 & 0.290 &
    0.219 & 0.288 &
    0.225 & 0.298 \\
    
    \multicolumn{1}{c}{} & 336 & 
    \textbf{0.276} & \textbf{0.335} &
    0.319 & 0.355 &
    0.319 & 0.351 &
    0.337 & 0.367 &
    \textbf{0.225} & \textbf{0.304} &
    0.271 & 0.329 &
    0.268 & 0.326 &
    0.286 & 0.341 &
    0.267 & 0.321 &
    0.269 & 0.330 &
    0.267 & 0.319 &
    0.277 & 0.332 \\
    
    \multicolumn{1}{c}{} & 720 & 
    \textbf{0.361} & \textbf{0.394} &
    0.415 & 0.410 &
    0.420 & 0.410 &
    0.430 & 0.416 &
    \textbf{0.273} & \textbf{0.343} &
    0.352 & 0.379 &
    0.350 & 0.380 &
    0.378 & 0.401 &
    0.348 & 0.374 &
    0.349 & 0.382 &
    0.361 & 0.377 &
    0.360 & 0.387 \\
    
    \cline{2-26} 
    
    \multicolumn{1}{c}{} & Avg. & 
    \textbf{0.263} & \textbf{0.329} &
    0.300 & 0.341 &
    0.293 & 0.334 &
    0.321 & 0.356 &
    \textbf{0.224} & \textbf{0.303} &
    0.251 & 0.313 &
    0.251 & 0.313 &
    0.267 & 0.326 &
    0.247 & 0.307 &
    0.250 & 0.313 &
    0.253 & 0.308 &
    0.257 & 0.318 \\
    
    \hline 
    
    \multicolumn{1}{c}{\multirow{5}{*}{\rotatebox{90}{ETTh1}}} & 96 & 
    \textbf{0.364} & 0.404 &
    0.375 & \textbf{0.402} &
    0.397 & 0.418 &
    0.449 & 0.424 &
    \textbf{0.274} & \textbf{0.301} &
    0.362 & 0.392 &
    0.371 & 0.394 &
    0.376 & 0.397 &
    0.352 & 0.384 &
    0.360 & 0.391 &
    0.372 & 0.392 &
    0.372 & 0.401 \\
    
    \multicolumn{1}{c}{} & 192 & 
    \textbf{0.384} & \textbf{0.392} &
    0.399 & 0.419 &
    0.434 & 0.439 &
    0.503 & 0.453 &
    \textbf{0.290} & \textbf{0.339} &
    0.398 & 0.418 &
    0.403 & 0.412 &
    0.416 & 0.418 &
    0.398 & 0.409 &
    0.392 & 0.414 &
    0.408 & 0.415 &
    0.413 & 0.430 \\
    
    \multicolumn{1}{c}{} & 336 & 
    \textbf{0.394} & \textbf{0.409} &
    0.412 & 0.429 &
    0.468 & 0.457 &
    0.540 & 0.477 &
    \textbf{0.383} & 0.424 &
    0.430 & 0.427 &
    0.420 & \textbf{0.422} &
    0.442 & 0.433 &
    0.414 & 0.426 &
    0.418 & 0.435 &
    0.438 & 0.434 &
    0.438 & 0.450 \\
    
    \multicolumn{1}{c}{} & 720 & 
    \textbf{0.412} & \textbf{0.415} &
    0.413 & 0.444 &
    0.469 & 0.477 &
    0.515 & 0.489 &
    \textbf{0.387} & \textbf{0.428} &
    0.442 & 0.457 &
    0.422 & 0.444 &
    0.477 & 0.456 &
    0.429 & 0.455 &
    0.456 & 0.462 &
    0.450 & 0.463 &
    0.486 & 0.484 \\
    
    \cline{2-26} 
    
    \multicolumn{1}{c}{} & Avg. & 
    \textbf{0.388} & \textbf{0.405} &
    0.399 & 0.424 &
    0.442 & 0.448 &
    0.502 & 0.461 &
    \textbf{0.334} & \textbf{0.383} &
    0.408 & 0.423 &
    0.404 & 0.418 &
    0.428 & 0.426 &
    0.398 & 0.418 &
    0.406 & 0.425 &
    0.417 & 0.426 &
    0.427 & 0.441 \\
    
    \hline 

    \multicolumn{1}{c}{\multirow{5}{*}{\rotatebox{90}{ETTh2}}} & 96 & 
    0.291 & 0.342 &
    \textbf{0.281} & \textbf{0.334} &
    0.296 & 0.345 &
    0.303 & 0.349 &
    \textbf{0.241} & \textbf{0.301} &
    0.268 & 0.328 &
    0.262 & 0.332 &
    0.285 & 0.342 &
    0.270 & 0.336 &
    0.276 & 0.341 &
    0.279 & 0.336 &
    0.281 & 0.351 \\
    
    \multicolumn{1}{c}{} & 192 & 
    0.387 & 0.405 &
    \textbf{0.340} & \textbf{0.373} &
    0.374 & 0.394 &
    0.391 & 0.399 &
    \textbf{0.275} & 0.375 &
    0.329 & 0.375 &
    0.328 & 0.377 &
    0.354 & 0.389 &
    0.332 & \textbf{0.374} &
    0.339 & 0.382 &
    0.345 & 0.380 &
    0.349 & 0.387 \\
    
    \multicolumn{1}{c}{} & 336 & 
    0.396 & 0.417 &
    \textbf{0.362} & \textbf{0.393} &
    0.415 & 0.427 &
    0.422 & 0.428 &
    \textbf{0.302} & \textbf{0.372} &
    0.368 & 0.409 &
    0.353 & 0.396 &
    0.373 & 0.407 &
    0.353 & 0.397 &
    0.374 & 0.406 &
    0.378 & 0.408 &
    0.366 & 0.413 \\
    
    \multicolumn{1}{c}{} & 720 & 
    0.443 & 0.454 &
    \textbf{0.380} & \textbf{0.416} &
    0.425 & 0.444 &
    0.429 & 0.449 &
    \textbf{0.323} & \textbf{0.386} &
    0.372 & 0.420 &
    0.383 & 0.425 &
    0.406 & 0.441 &
    0.382 & 0.425 &
    0.398 & 0.433 &
    0.437 & 0.455 &
    0.401 & 0.436 \\
    
    \cline{2-26} 
    
    \multicolumn{1}{c}{} & Avg. & 
    0.379 & 0.405 &
    \textbf{0.341} & \textbf{0.379} &
    0.378 & 0.403 &
    0.386 & 0.406 &
    \textbf{0.285} & \textbf{0.358} &
    0.334 & 0.383 &
    0.331 & 0.383 &
    0.355 & 0.395 &
    0.334 & 0.383 &
    0.347 & 0.391 &
    0.360 & 0.395 &
    0.349 & 0.397 \\
    
    \hline 

    \multicolumn{1}{c}{\multirow{5}{*}{\rotatebox{90}{Electricity}}} & 96 & 
    \textbf{0.163} & \textbf{0.278} &
    0.205 & 0.299 &
    0.196 & 0.287 &
    0.232 & 0.321 &
    \textbf{0.128} & 0.221 &
    0.131 & 0.224 &
    0.128 & 0.223 &
    0.139 & 0.238 &
    0.128 & \textbf{0.219} &
    0.128 & 0.222 &
    0.135 & 0.222 &
    0.153 & 0.256 \\
    
    \multicolumn{1}{c}{} & 192 & 
    \textbf{0.172} & \textbf{0.282} &
    0.220 & 0.310 &
    0.199 & 0.291 &
    0.234 & 0.325 &
    0.147 & 0.240 &
    0.152 & 0.241 &
    0.146 & 0.240 &
    0.153 & 0.251 &
    \textbf{0.145} & \textbf{0.235} &
    0.147 & 0.242 &
    0.157 & 0.253 &
    0.168 & 0.269 \\
    
    \multicolumn{1}{c}{} & 336 & 
    \textbf{0.183} & \textbf{0.278} &
    0.236 & 0.323 &
    0.214 & 0.305 &
    0.249 & 0.338 &
    \textbf{0.149} & \textbf{0.244} &
    0.160 & 0.248 &
    0.163 & 0.258 &
    0.169 & 0.266 &
    0.163 & 0.255 &
    0.165 & 0.260 &
    0.170 & 0.267 &
    0.189 & 0.291 \\
    
    \multicolumn{1}{c}{} & 720 & 
    \textbf{0.209} & \textbf{0.312} &
    0.270 & 0.347 &
    0.254 & 0.335 &
    0.289 & 0.366 &
    \textbf{0.172} & \textbf{0.272} &
    0.192 & 0.298 &
    0.200 & 0.292 &
    0.206 & 0.297 &
    0.193 & 0.281 &
    0.199 & 0.289 &
    0.211 & 0.302 &
    0.228 & 0.320 \\
    
    \cline{2-26} 
    
    \multicolumn{1}{c}{} & Avg. & 
    \textbf{0.183} & \textbf{0.289} &
    0.233 & 0.320 &
    0.216 & 0.305 &
    0.251 & 0.338 &
    \textbf{0.149} & \textbf{0.244} &
    0.158 & 0.252 &
    0.159 & 0.253 &
    0.167 & 0.263 &
    0.157 & 0.247 &
    0.160 & 0.253 &
    0.168 & 0.261 &
    0.184 & 0.284 \\
    
    \hline 

    \multicolumn{1}{c}{\multirow{5}{*}{\rotatebox{90}{Traffic}}} & 96 & 
    0.329 & \textbf{0.214} &
    0.343 & 0.263 &
    \textbf{0.328} & 0.252 &
    0.388 & 0.282 &
    \textbf{0.284} & \textbf{0.203} &
    0.362 & 0.248 &
    0.372 & 0.259 &
    0.388 & 0.282 &
    0.360 & 0.238 &
    0.368 & 0.252 &
    0.384 & 0.250 &
    0.369 & 0.257 \\
    
    \multicolumn{1}{c}{} & 192 & 
    \textbf{0.340} & \textbf{0.240} &
    0.383 & 0.277 &
    0.346 & 0.261 &
    0.407 & 0.290 &
    \textbf{0.293} & \textbf{0.211} &
    0.374 & 0.247 &
    0.391 & 0.265 &
    0.407 & 0.290 &
    0.383 & 0.249 &
    0.382 & 0.261 &
    0.405 & 0.257 &
    0.400 & 0.272 \\
    
    \multicolumn{1}{c}{} & 336 & 
    \textbf{0.348} & \textbf{0.241} &
    0.390 & 0.281 &
    0.354 & 0.265 &
    0.412 & 0.294 &
    \textbf{0.308} & \textbf{0.215} &
    0.385 & 0.271 &
    0.405 & 0.275 &
    0.412 & 0.294 &
    0.395 & 0.259 &
    0.393 & 0.268 &
    0.424 & 0.265 &
    0.407 & 0.272 \\
    
    \multicolumn{1}{c}{} & 720 & 
    \textbf{0.351} & \textbf{0.260} &
    0.420 & 0.296 &
    0.396 & 0.286 &
    0.450 & 0.312 &
    \textbf{0.319} & \textbf{0.223} &
    0.430 & 0.288 &
    0.437 & 0.292 &
    0.450 & 0.312 &
    0.435 & 0.278 &
    0.438 & 0.297 &
    0.452 & 0.283 &
    0.461 & 0.316 \\
    
    \cline{2-26} 
    
    \multicolumn{1}{c}{} & Avg. & 
    \textbf{0.342} & \textbf{0.239} &
    0.384 & 0.279 &
    0.356 & 0.266 &
    0.414 & 0.295 &
    \textbf{0.301} & \textbf{0.213} &
    0.388 & 0.264 &
    0.401 & 0.273 &
    0.414 & 0.295 &
    0.393 & 0.256 &
    0.395 & 0.270 &
    0.416 & 0.264 &
    0.409 & 0.279 \\
    
    \hline
    
    \multicolumn{1}{c}{\multirow{5}{*}{\rotatebox{90}{Weather}}} & 96 & 
    \textbf{0.157} & \textbf{0.206} &
    0.173 & 0.212 &
    0.171 & 0.214 &
    0.212 & 0.251 &
    \textbf{0.133} & 0.195 &
    0.147 & 0.201 &
    0.147 & 0.196 &
    0.162 & 0.212 &
    0.146 & \textbf{0.191} &
    0.147 & 0.196 &
    0.148 & 0.195 &
    0.147 & 0.198 \\
    
    \multicolumn{1}{c}{} & 192 & 
    \textbf{0.204} & \textbf{0.250} & 
    0.216 & 0.250 &
    0.217 & 0.254 &
    0.261 & 0.288 &
    \textbf{0.184} & 0.237 &
    0.189 & 0.234 &
    0.191 & 0.238 &
    0.204 & 0.248 &
    0.188 & \textbf{0.231} &
    0.193 & 0.240 &
    0.191 & 0.235 &
    0.192 & 0.243 \\
    
    \multicolumn{1}{c}{} & 336 & 
    \textbf{0.251} & \textbf{0.279} &
    0.260 & 0.282 &
    0.274 & 0.293 &
    0.313 & 0.324 &
    \textbf{0.207} & \textbf{0.258} &
    0.262 & 0.279 &
    0.241 & 0.277 &
    0.254 & 0.286 &
    0.234 & 0.268 &
    0.245 & 0.280 &
    0.243 & 0.274 &
    0.247 & 0.284 \\
    
    \multicolumn{1}{c}{} & 720 & 
    \textbf{0.309} & \textbf{0.317} &
    0.320 & 0.322 &
    0.351 & 0.343 &
    0.386 & 0.372 &
    \textbf{0.264} & \textbf{0.313} &
    0.304 & 0.316 &
    0.313 & 0.329 &
    0.326 & 0.337 &
    0.305 & 0.319 &
    0.323 & 0.334 &
    0.318 & 0.326 &
    0.318 & 0.330 \\
    
    \cline{2-26}
    
    \multicolumn{1}{c}{} & Avg. & 
    \textbf{0.230} & \textbf{0.263} &
    0.242 & 0.267 &
    0.253 & 0.276 &
    0.293 & 0.309 &
    \textbf{0.202} & \textbf{0.246} &
    0.225 & 0.257 &
    0.223 & 0.260 &
    0.237 & 0.271 &
    0.218 & 0.252 &
    0.227 & 0.263 &
    0.225 & 0.258 &
    0.226 & 0.264 \\
    
    \hline 

    \multicolumn{1}{c}{\multirow{5}{*}{\rotatebox{90}{Exchange}}} & 96 & 
    0.198 & 0.338 &
    0.094 & 0.223 &
    \textbf{0.086} & \textbf{0.209} &
    0.142 & 0.261 &
    0.085 & 0.236 &
    0.113 & 0.249 &
    0.107 & 0.221 &
    0.124 & 0.254 &
    \textbf{0.080} & \textbf{0.198} &
    0.083 & 0.200 &
    0.088 & 0.208 &
    0.084 & 0.207 \\
    
    \multicolumn{1}{c}{} & 192 & 
    0.215 & 0.328 &
    0.189 & 0.316 &
    \textbf{0.174} & \textbf{0.299} &
    0.224 & 0.339 &
    \textbf{0.141} & \textbf{0.251} &
    0.197 & 0.316 &
    0.193 & 0.315 &
    0.219 & 0.348 &
    0.162 & 0.288 &
    0.172 & 0.294 &
    0.183 & 0.304 &
    0.178 & 0.300 \\
    
    \multicolumn{1}{c}{} & 336 & 
    \textbf{0.270} & \textbf{0.377} &
    0.334 & 0.421 &
    0.319 & 0.408 &
    0.377 & 0.448 &
    \textbf{0.247} & \textbf{0.291} &
    0.298 & 0.425 &
    0.351 & 0.423 &
    0.357 & 0.432 &
    0.294 & 0.392 &
    0.323 & 0.411 &
    0.354 & 0.429 &
    0.376 & 0.451 \\
    
    \multicolumn{1}{c}{} & 720 & 
    0.882 & \textbf{0.546} &
    1.064 & 0.852 &
    \textbf{0.875} & 0.701 &
    0.939 & 0.736 &
    0.760 & 0.601 &
    0.914 & 0.729 &
    0.956 & 0.748 &
    0.993 & 0.821 &
    \textbf{0.583} & \textbf{0.580} &
    0.820 & 0.682 &
    0.909 & 0.716 &
    0.884 & 0.707 \\
    
    \cline{2-26} 
    
    \multicolumn{1}{c}{} & Avg. & 
    0.391 & \textbf{0.397} &
    0.420 & 0.453 &
    \textbf{0.364} & 0.404 &
    0.421 & 0.446 &
    0.308 & \textbf{0.345} &
    0.381 & 0.429 &
    0.402 & 0.426 &
    0.423 & 0.464 &
    \textbf{0.280} & 0.364 &
    0.350 & 0.397 &
    0.384 & 0.414 &
    0.381 & 0.416 \\
    
    \hline
    
    \multicolumn{1}{c}{\multirow{5}{*}{\rotatebox{90}{ILI}}} & 24 & 
    \textbf{2.018} & \textbf{0.938} &
    2.157 & 0.962 &
    2.460 & 0.954 &
    3.322 & 1.278 &
    1.863 & 0.923 &
    \textbf{1.285} & \textbf{0.727} &
    1.894 & 0.832 &
    2.063 & 0.881 &
    1.577 & 0.760 &
    1.801 & 0.874 &
    2.086 & 0.922 &
    1.804 & 0.820 \\
    
    \multicolumn{1}{c}{} & 36 & 
    \textbf{1.782} & \textbf{0.861} &
    1.858 & 0.879 &
    1.998 & 0.912 &
    3.696 & 1.374 &
    1.639 & 0.851 &
    \textbf{1.404} & 0.814 &
    1.725 & 0.863 &
    1.868 & 0.892 &
    1.596 & \textbf{0.794} &
    1.743 & 0.867 &
    1.912 & 0.882 &
    1.891 & 0.926 \\
    
    \multicolumn{1}{c}{} & 48 & 
    \textbf{1.824} & \textbf{0.859} &
    1.923 & 0.873 &
    1.979 & 0.912 &
    3.765 & 1.402 &
    1.745 & 0.825 &
    \textbf{1.523} & \textbf{0.807} &
    1.681 & 0.855 &
    1.790 & 0.884 &
    1.632 & 0.810 &
    1.843 & 0.926 &
    1.985 & 0.905 &
    1.752 & 0.866 \\
    
    \multicolumn{1}{c}{} & 60 & 
    \textbf{1.892} & \textbf{0.873} &
    2.282 & 0.915 &
    2.109 & 0.938 &
    3.928 & 1.432 &
    1.721 & 0.817 &
    \textbf{1.531} & 0.854 &
    1.728 & 0.903 &
    1.979 & 0.957 &
    1.660 & \textbf{0.815} &
    1.845 & 0.925 &
    1.999 & 0.929 &
    1.831 & 0.930 \\
    
    \cline{2-26} 
    
    \multicolumn{1}{c}{} & Avg. & 
    \textbf{1.872} & \textbf{0.883} &
    2.055 & 0.907 &
    2.137 & 0.929 &
    3.678 & 1.372 &
    1.742 & 0.854 &
    \textbf{1.435} & 0.801 &
    1.757 & 0.863 &
    1.925 & 0.903 &
    1.616 & \textbf{0.795} &
    1.808 & 0.898 &
    1.995 & 0.909 &
    1.820 & 0.886 \\
    
    \hline

    \multicolumn{1}{c}{\multirow{5}{*}{\rotatebox{90}{Solar}}} & 96 & 
    \textbf{0.162} & \textbf{0.210} &
    0.180 & 0.233 &
    0.264 & 0.245 &
    0.383 & 0.325 &
    \textbf{0.158} & \textbf{0.195} &
    0.241 & 0.279 &
    0.219 & 0.239 &
    0.188 & 0.216 &
    0.169 & 0.218 &
    0.181 & 0.247 &
    0.218 & 0.235 &
    0.179 & 0.232 \\
    
    \multicolumn{1}{c}{} & 192 & 
    \textbf{0.178} & 0.238 &
    0.194 & 0.252 &
    0.234 & \textbf{0.238} &
    0.312 & 0.288 &
    \textbf{0.162} & \textbf{0.207} &
    0.222 & 0.273 &
    0.197 & 0.223 &
    0.203 & 0.222 &
    0.187 & 0.254 &
    0.200 & 0.259 &
    0.196 & 0.220 &
    0.201 & 0.259 \\
    
    \multicolumn{1}{c}{} & 336 & 
    0.184 & 0.242 &
    \textbf{0.179} & 0.244 &
    0.241 & \textbf{0.240} &
    0.294 & 0.280 &
    0.168 & 0.246 &
    \textbf{0.151} & \textbf{0.213} &
    0.196 & 0.225 &
    0.216 & 0.227 &
    0.199 & 0.241 &
    0.208 & 0.269 &
    0.195 & 0.220 &
    0.190 & 0.256 \\
    
    \multicolumn{1}{c}{} & 720 & 
    \textbf{0.170} & 0.241 &
    0.173 & \textbf{0.239} &
    0.257 & 0.249 &
    0.288 & 0.287 &
    \textbf{0.172} & \textbf{0.216} &
    0.210 & 0.269 &
    0.202 & 0.239 &
    0.216 & 0.228 &
    0.202 & 0.245 &
    0.212 & 0.275 &
    0.208 & 0.237 &
    0.203 & 0.261 \\
    
    \cline{2-26} 
    
    \multicolumn{1}{c}{} & Avg. & 
    \textbf{0.173} & \textbf{0.233} &
    0.180 & 0.242 &
    0.250 & 0.243 &
    0.315 & 0.294 &
    \textbf{0.164} & \textbf{0.208} &
    0.193 & 0.260 &
    0.202 & 0.232 &
    0.205 & 0.224 &
    0.189 & 0.241 &
    0.200 & 0.263 &
    0.204 & 0.228 &
    0.193 & 0.252 \\
    
    \hline
    
    \multicolumn{2}{c}{$1^{\text{st}}$ Count} & 
    \multicolumn{2}{c|}{\textbf{74}} & 
    \multicolumn{2}{c|}{13} & 
    \multicolumn{2}{c|}{13} & 
    \multicolumn{2}{c||}{0} & 
    \multicolumn{2}{c|}{\textbf{72}} & 
    \multicolumn{2}{c|}{11} & 
    \multicolumn{2}{c|}{1} & 
    \multicolumn{2}{c|}{0} & 
    \multicolumn{2}{c|}{16} & 
    \multicolumn{2}{c|}{0} & 
    \multicolumn{2}{c|}{0} & 
    \multicolumn{2}{c}{0} \\
    \hline
    \bottomrule
    \end{tabular}
    } 
    \vspace{-0.5cm}
\end{table*}

%% file: tables_brief/forecasting_zero_all.tex
\begin{table*}[htpb]
\caption{
Comparison of the performance on \textbf{domain-transfer forecasting} task. 
We utilize the notation ``source\small{$\rightarrow$}target'' to represent the scenario of domain transfer. Specifically, the left side of the double vertical line illustrates performance comparisons between TimeControl and existing foundational models. These methods are first pre-trained on textual (e.g., TimeLLM, LLMTime, GPT4TS), visual (TimeVLM), or time-series (TimeControl) domains, followed by continuing to train on the source domain, and ultimately evaluated on the target domain. The right side of the double vertical line demonstrates performance comparisons of existing proprietary models, which are trained from scratch on the source domain and subsequently evaluated on the target domain.
}\label{tab:brief_forecasting_zero_all}
\vskip 0.05in
\centering
\resizebox{2.0\columnwidth}{!}{
\begin{small}
\renewcommand{\multirowsetup}{\centering}
\tabcolsep=0.15cm
\renewcommand\arraystretch{1.3}
\begin{tabular}{cccc|cc|cc|cc|cc||cc|cc|cc|cc|cc|cc}
\toprule
\hline

\multicolumn{2}{c}{\multirow{5}{*}{Model}} & 
\multicolumn{10}{c||}{\textbf{\textit{Time Series Foundation Models}}} & 
\multicolumn{12}{c}{\textbf{\textit{Time Series Proprietary Models}}} \\ 

\multicolumn{2}{c}{} & 
\multicolumn{10}{c||}{\textbf{\textit{(Pre-trained on multi-domain, continue trained on the source domain)}}} & 
\multicolumn{12}{c}{\textbf{\textit{(Trained from scratch on the source domain)}}} \\ 

\cline{3-24} 

\multicolumn{2}{c}{\multirow{3}{*}{Model}} &
\multicolumn{2}{c}{\textbf{\myformer}} & 
\multicolumn{2}{c}{TimeVLM} &
\multicolumn{2}{c}{TimeLLM} &
\multicolumn{2}{c}{LLMTime} &
\multicolumn{2}{c||}{GPT4TS} &

\multicolumn{2}{c}{DUET} &
\multicolumn{2}{c}{PDF} &
\multicolumn{2}{c}{Pathformer} &
\multicolumn{2}{c}{TimeMixer} &
\multicolumn{2}{c}{PatchTST} &
\multicolumn{2}{c}{DLinear} \\

& &
\multicolumn{2}{c}{\textbf{(Ours))}} & 
\multicolumn{2}{c}{\cite{zhong2025timevlm}} &
\multicolumn{2}{c}{\cite{Jin2023TimeLLMTS}} &
\multicolumn{2}{c}{\cite{llmtime}} &
\multicolumn{2}{c||}{\cite{Zhou2023OneFA}} &

\multicolumn{2}{c}{\cite{qiu2025duet}} &
\multicolumn{2}{c}{\cite{dai2024pdf}} &
\multicolumn{2}{c}{\cite{chen2024pathformer}} &
\multicolumn{2}{c}{\cite{wang2024timemixer}} &
\multicolumn{2}{c}{\cite{Nie2023PatchTST}} &
\multicolumn{2}{c}{\cite{zeng2023dlinear}} \\

\cline{3-24} & &
MSE & MAE & MSE & MAE & MSE & MAE & MSE & MAE & MSE & MAE & 
MSE & MAE & MSE & MAE & MSE & MAE & MSE & MAE & MSE & MAE & MSE & MAE \\

\hline
\multirow{5}{*}{\shortstack{ETTh1\\ $\downarrow$ \\ETTh2}}
& 96 & \textbf{0.273} & \textbf{0.328} & 0.277 & 0.338 & 0.279 & 0.337 & 0.510 & 0.576 & 0.335 & 0.374 & 0.347 & 0.400 & 0.304 & 0.350 & 0.358 & 0.387 & 0.479 & 0.489 & 0.469 & 0.486 & 0.445 & 0.603 \\
& 192 & \textbf{0.309} & \textbf{0.356} & 0.333 & 0.378 & 0.351 & 0.374 & 0.523 & 0.586 & 0.412 & 0.417 & 0.447 & 0.460 & 0.386 & 0.400 & 0.427 & 0.429 & 0.590 & 0.545 & 0.634 & 0.567 & 0.457 & 0.613 \\
& 336 & \textbf{0.335} & \textbf{0.378} & 0.360 & 0.399 & 0.388 & 0.415 & 0.640 & 0.637 & 0.441 & 0.444 & 0.515 & 0.505 & 0.414 & 0.428 & 0.449 & 0.451 & 0.631 & 0.581 & 0.655 & 0.588 & 0.559 & 0.667 \\
& 720 & \textbf{0.350} & \textbf{0.379} & 0.383 & 0.425 & 0.391 & 0.420 & 2.296 & 1.034 & 0.438 & 0.452 & 0.665 & 0.589 & 0.419 & 0.443 & 0.448 & 0.458 & 0.627 & 0.591 & 0.570 & 0.549 & 2.004 & 1.082 \\
\cline{2-24} 
& Avg. & \textbf{0.317} & \textbf{0.360} & 0.338 & 0.385 & 0.353 & 0.387 & 0.992 & 0.708 & 0.406 & 0.422 & 0.493 & 0.488 & 0.380 & 0.405 & 0.421 & 0.431 & 0.582 & 0.552 & 0.582 & 0.548 & 0.866 & 0.741 \\

\hline
\multirow{5}{*}{\shortstack{ETTh1\\ $\downarrow$ \\ETTm2}}
& 96 & 0.204 & 0.305 & 0.207 & 0.297 & \textbf{0.189} & \textbf{0.293} & 0.646 & 0.563 & 0.236 & 0.315 & 0.255 & 0.357 & 0.215 & 0.304 & 0.239 & 0.313 & 0.321 & 0.413 & 0.352 & 0.432 & 0.511 & 0.579 \\
& 192 & 0.259 & 0.338 & 0.258 & 0.329 & \textbf{0.237} & \textbf{0.312} & 0.934 & 0.654 & 0.287 & 0.342 & 0.338 & 0.413 & 0.275 & 0.339 & 0.291 & 0.342 & 0.401 & 0.458 & 0.413 & 0.460 & 0.654 & 0.646 \\
& 336 & 0.309 & 0.366 & 0.310 & \textbf{0.360} & \textbf{0.291} & 0.365 & 1.157 & 0.728 & 0.341 & 0.374 & 0.425 & 0.465 & 0.334 & 0.373 & 0.342 & 0.371 & 0.481 & 0.501 & 0.465 & 0.489 & 0.795 & 0.711 \\
& 720 & 0.393 & 0.415 & 0.398 & 0.412 & \textbf{0.372} & \textbf{0.390} & 4.730 & 1.531 & 0.435 & 0.422 & 0.640 & 0.573 & 0.431 & 0.424 & 0.434 & 0.419 & 0.618 & 0.573 & 0.599 & 0.551 & 1.025 & 0.808 \\
\cline{2-24} 
& Avg. & 0.291 & 0.356 & 0.293 & 0.350 & \textbf{0.273} & \textbf{0.340} & 1.867 & 0.869 & 0.325 & 0.363 & 0.415 & 0.452 & 0.314 & 0.360 & 0.327 & 0.361 & 0.455 & 0.487 & 0.457 & 0.483 & 0.747 & 0.686 \\

\hline
\multirow{5}{*}{\shortstack{ETTh2\\ $\downarrow$ \\ETTh1}}
& 96 & \textbf{0.373} & \textbf{0.404} & 0.434 & 0.441 & 0.450 & 0.452 & 1.130 & 0.777 & 0.732 & 0.577 & 0.689 & 0.555 & 0.485 & 0.465 & 0.848 & 0.601 & 0.753 & 0.593 & 0.693 & 0.569 & 1.070 & 0.787 \\
& 192 & \textbf{0.405} & \textbf{0.422} & 0.464 & 0.454 & 0.465 & 0.461 & 1.242 & 0.820 & 0.758 & 0.559 & 0.707 & 0.568 & 0.565 & 0.509 & 0.860 & 0.610 & 0.764 & 0.602 & 0.760 & 0.601 & 1.144 & 0.811 \\
& 336 & \textbf{0.434} & \textbf{0.442} & 0.489 & 0.481 & 0.501 & 0.482 & 1.328 & 0.864 & 0.759 & 0.578 & 0.710 & 0.577 & 0.581 & 0.515 & 0.867 & 0.626 & 0.770 & 0.618 & 0.781 & 0.619 & 1.206 & 0.859 \\
& 720 & \textbf{0.489} & \textbf{0.488} & 0.595 & 0.543 & 0.501 & 0.502 & 4.145 & 1.461 & 0.781 & 0.597 & 0.704 & 0.596 & 0.628 & 0.561 & 0.887 & 0.648 & 0.788 & 0.640 & 0.796 & 0.644 & 1.467 & 0.969 \\
\cline{2-24} 
& Avg. & \textbf{0.425} & \textbf{0.439} & 0.496 & 0.480 & 0.479 & 0.474 & 1.961 & 0.981 & 0.757 & 0.578 & 0.703 & 0.574 & 0.565 & 0.513 & 0.865 & 0.621 & 0.768 & 0.613 & 0.757 & 0.608 & 1.223 & 0.857 \\

\hline
\multirow{5}{*}{\shortstack{ETTh2\\ $\downarrow$ \\ETTm2}}
& 96 & 0.179 & \textbf{0.263} & 0.204 & 0.297 & \textbf{0.174} & 0.276 & 0.646 & 0.563 & 0.253 & 0.329 & 0.240 & 0.336 & 0.226 & 0.309 & 0.248 & 0.324 & 0.126 & 0.265 & 0.263 & 0.352 & 0.477 & 0.493 \\
& 192 & \textbf{0.213} & \textbf{0.296} & 0.255 & 0.328 & 0.233 & 0.315 & 0.934 & 0.654 & 0.293 & 0.346 & 0.295 & 0.369 & 0.289 & 0.345 & 0.296 & 0.352 & 0.182 & 0.308 & 0.326 & 0.389 & 0.569 & 0.535 \\
& 336 & \textbf{0.254} & \textbf{0.327} & 0.311 & 0.362 & 0.291 & 0.337 & 1.157 & 0.728 & 0.347 & 0.376 & 0.345 & 0.397 & 0.348 & 0.379 & 0.353 & 0.383 & 0.225 & 0.343 & 0.387 & 0.426 & 0.679 & 0.583 \\
& 720 & \textbf{0.378} & \textbf{0.384} & 0.420 & 0.425 & 0.392 & 0.417 & 4.730 & 1.531 & 0.446 & 0.429 & 0.432 & 0.442 & 0.439 & 0.427 & 0.471 & 0.446 & 0.920 & 0.721 & 0.487 & 0.478 & 0.906 & 0.678 \\
\cline{2-24} 
& Avg. & \textbf{0.256} & \textbf{0.318} & 0.297 & 0.353 & 0.272 & 0.341 & 1.867 & 0.869 & 0.335 & 0.370 & 0.328 & 0.386 & 0.325 & 0.365 & 0.342 & 0.376 & 0.363 & 0.409 & 0.366 & 0.411 & 0.658 & 0.572 \\

\hline
\multirow{5}{*}{\shortstack{ETTm1\\ $\downarrow$ \\ETTh2}}
& 96 & \textbf{0.277} & \textbf{0.317} & 0.297 & 0.356 & 0.321 & 0.369 & 0.510 & 0.576 & 0.353 & 0.392 & 0.365 & 0.415 & 0.354 & 0.385 & 0.377 & 0.407 & 0.368 & 0.422 & 0.435 & 0.470 & 0.647 & 0.609 \\
& 192 & \textbf{0.311} & \textbf{0.344} & 0.349 & 0.388 & 0.389 & 0.410 & 0.523 & 0.586 & 0.443 & 0.437 & 0.454 & 0.462 & 0.447 & 0.434 & 0.471 & 0.453 & 0.458 & 0.470 & 0.495 & 0.489 & 0.784 & 0.677 \\
& 336 & \textbf{0.329} & \textbf{0.362} & 0.374 & 0.409 & 0.408 & 0.433 & 0.640 & 0.637 & 0.469 & 0.461 & 0.496 & 0.494 & 0.481 & 0.463 & 0.472 & 0.484 & 0.500 & 0.502 & 0.470 & 0.472 & 0.822 & 0.715 \\
& 720 & 0.401 & \textbf{0.408} & \textbf{0.396} & 0.433 & 0.406 & 0.436 & 2.296 & 1.034 & 0.466 & 0.468 & 0.541 & 0.529 & 0.474 & 0.471 & 0.495 & 0.482 & 0.546 & 0.538 & 0.480 & 0.485 & 0.818 & 0.720 \\
\cline{2-24} 
& Avg. & \textbf{0.329} & \textbf{0.358} & 0.354 & 0.397 & 0.381 & 0.412 & 0.992 & 0.708 & 0.433 & 0.439 & 0.464 & 0.475 & 0.439 & 0.438 & 0.457 & 0.454 & 0.468 & 0.483 & 0.470 & 0.479 & 0.768 & 0.680 \\

\hline
\multirow{5}{*}{\shortstack{ETTm1\\ $\downarrow$ \\ETTm2}}
& 96 & \textbf{0.156} & \textbf{0.252} & 0.178 & 0.264 & 0.169 & 0.257 & 0.646 & 0.563 & 0.217 & 0.294 & 0.221 & 0.314 & 0.195 & 0.271 & 0.222 & 0.295 & 0.293 & 0.385 & 0.385 & 0.457 & 0.455 & 0.538 \\
& 192 & \textbf{0.208} & \textbf{0.285} & 0.226 & 0.298 & 0.227 & 0.318 & 0.934 & 0.654 & 0.277 & 0.327 & 0.286 & 0.359 & 0.258 & 0.311 & 0.288 & 0.337 & 0.394 & 0.476 & 0.433 & 0.469 & 0.581 & 0.599 \\
& 336 & \textbf{0.265} & \textbf{0.324} & 0.279 & 0.329 & 0.290 & 0.338 & 1.157 & 0.728 & 0.331 & 0.360 & 0.357 & 0.406 & 0.317 & 0.348 & 0.341 & 0.367 & 0.503 & 0.506 & 0.476 & 0.477 & 0.695 & 0.659 \\
& 720 & \textbf{0.344} & \textbf{0.373} & 0.373 & 0.385 & 0.375 & 0.367 & 4.730 & 1.531 & 0.429 & 0.413 & 0.476 & 0.476 & 0.416 & 0.404 & 0.436 & 0.418 & 0.651 & 0.549 & 0.582 & 0.535 & 0.900 & 0.756 \\
\cline{2-24} 
& Avg. & \textbf{0.243} & \textbf{0.309} & 0.264 & 0.319 & 0.268 & 0.320 & 1.867 & 0.869 & 0.313 & 0.348 & 0.335 & 0.389 & 0.296 & 0.334 & 0.322 & 0.354 & 0.460 & 0.479 & 0.469 & 0.484 & 0.657 & 0.637 \\

\hline 
\multirow{5}{*}{\shortstack{ETTm2\\ $\downarrow$ \\ETTh2}}
& 96 & \textbf{0.239} & \textbf{0.301} & 0.285 & 0.347 & 0.298 & 0.356 & 0.510 & 0.576 & 0.360 & 0.401 & 0.333 & 0.391 & 0.327 & 0.367 & 0.360 & 0.401 & 0.337 & 0.351 & 0.353 & 0.393 & 0.448 & 0.623 \\
& 192 & \textbf{0.275} & \textbf{0.329} & 0.347 & 0.383 & 0.358 & 0.398 & 0.522 & 0.585 & 0.435 & 0.435 & 0.439 & 0.456 & 0.411 & 0.416 & 0.435 & 0.435 & 0.423 & 0.397 & 0.434 & 0.437 & 0.590 & 0.726 \\
& 336 & \textbf{0.302} & \textbf{0.351} & 0.380 & 0.415 & 0.367 & 0.412 & 0.640 & 0.637 & 0.460 & 0.459 & 0.505 & 0.503 & 0.439 & 0.447 & 0.460 & 0.459 & 0.452 & 0.427 & 0.452 & 0.459 & 0.679 & 0.801 \\
& 720 & \textbf{0.316} & \textbf{0.412} & 0.424 & 0.451 & 0.393 & 0.434 & 2.296 & 1.034 & 0.485 & 0.477 & 0.543 & 0.534 & 0.459 & 0.470 & 0.485 & 0.477 & 0.472 & 0.449 & 0.453 & 0.467 & 0.730 & 0.850 \\
\cline{2-24} 
& Avg. & \textbf{0.283} & \textbf{0.348} & 0.359 & 0.399 & 0.354 & 0.400 & 0.992 & 0.708 & 0.435 & 0.443 & 0.455 & 0.471 & 0.409 & 0.425 & 0.435 & 0.443 & 0.421 & 0.406 & 0.423 & 0.439 & 0.612 & 0.750 \\

\hline
\multirow{5}{*}{\shortstack{ETTm2\\ $\downarrow$ \\ETTm1}}
& 96 & \textbf{0.335} & \textbf{0.357} & 0.370 & 0.390 & 0.359 & 0.397 & 1.179 & 0.781 & 0.747 & 0.558 & 0.570 & 0.490 & 0.491 & 0.437 & 0.747 & 0.558 & 0.730 & 0.580 & 0.735 & 0.576 & 0.740 & 0.658 \\
& 192 & \textbf{0.354} & \textbf{0.371} & 0.400 & 0.409 & 0.390 & 0.420 & 1.327 & 0.846 & 0.781 & 0.560 & 0.590 & 0.506 & 0.530 & 0.470 & 0.781 & 0.560 & 0.763 & 0.582 & 0.753 & 0.586 & 0.833 & 0.713 \\
& 336 & \textbf{0.380} & \textbf{0.384} & 0.426 & 0.420 & 0.421 & 0.445 & 1.478 & 0.902 & 0.778 & 0.578 & 0.706 & 0.567 & 0.565 & 0.497 & 0.778 & 0.578 & 0.760 & 0.600 & 0.750 & 0.593 & 0.927 & 0.760 \\
& 720 & \textbf{0.427} & \textbf{0.412} & 0.531 & 0.487 & 0.487 & 0.488 & 3.749 & 1.408 & 0.769 & 0.573 & 0.731 & 0.584 & 0.686 & 0.565 & 0.769 & 0.573 & 0.751 & 0.595 & 0.782 & 0.609 & 2.353 & 1.186 \\
\cline{2-24} 
& Avg. & \textbf{0.374} & \textbf{0.382} & 0.432 & 0.426 & 0.414 & 0.438 & 1.933 & 0.984 & 0.769 & 0.567 & 0.649 & 0.537 & 0.568 & 0.492 & 0.769 & 0.567 & 0.751 & 0.589 & 0.755 & 0.591 & 1.213 & 0.829 \\

\hline
\multicolumn{2}{c}{$1^{\text{st}}$ Count} & 
\multicolumn{2}{c|}{\textbf{68}} & 
\multicolumn{2}{c|}{2} & 
\multicolumn{2}{c|}{10} & 
\multicolumn{2}{c|}{0} & 
\multicolumn{2}{c|}{0} & 
\multicolumn{2}{c|}{0} & 
\multicolumn{2}{c|}{0} & 
\multicolumn{2}{c|}{0} & 
\multicolumn{2}{c|}{0} & 
\multicolumn{2}{c|}{0} & 
\multicolumn{2}{c}{0} \\
\hline
\bottomrule
\end{tabular}
\end{small}
}
\end{table*}

%% file: tables_brief/generation_full.tex
\begin{table*}[htbp]
\caption{
To further verify the comprehensive performance of the proposed \myformer in Long-term Time-series Generation, we introduce additional evaluation metrics: Context-FID Score, Correlational Score, Discriminative Score, Predictive Score. We boldface the best performance on all metrics and datasets, respectively. For the diffusion-based generative models, we conducted three repeated experiments and report the mean values and variances. We boldface the best performance on all benchmarks, and TimeControl achieved optimal performance across most metrics.
}\label{tab:generation_full}
\vskip 0.05in
\centering
\resizebox{2.0\columnwidth}{!}{
\begin{threeparttable}
\begin{small}
\renewcommand{\multirowsetup}{\centering}
\tabcolsep=0.15cm
\renewcommand\arraystretch{1.3}
\begin{tabular}{cc|c|ccccccc|c}
\toprule
\hline

\multicolumn{2}{c|}{\multirow{2}{*}{\scalebox{1.0}{Dataset}}} &
\multicolumn{1}{c|}{\multirow{2}{*}{\scalebox{1.0}{Length}}} &
\multicolumn{1}{c}{\textbf{\myformer}} & 
\multicolumn{1}{c}{Diffusion-TS} & 
\multicolumn{1}{c}{TimeGAN} & 
\multicolumn{1}{c}{TimeVAE} & 
\multicolumn{1}{c}{TimeGrad} & 
\multicolumn{1}{c}{TimeDiff} &
\multicolumn{1}{c|}{Cot-GAN} &
\multicolumn{1}{c}{\multirow{2}{*}{{Improve(\%)}}} \\

& & &
\multicolumn{1}{c}{\textbf{(Ours)}} & 
\multicolumn{1}{c}{\cite{Yuan2024DiffusionTSID}} & 
\multicolumn{1}{c}{\cite{yoon2021timegan}} & 
\multicolumn{1}{c}{\cite{desai2023timevae}} & 
\multicolumn{1}{c}{\cite{Rasul2021timegrad}} & 
\multicolumn{1}{c}{\cite{shen2024timediff}} & 
\multicolumn{1}{c|}{\cite{xu2022cotgan}} & 
\\

\hline
\hline
\multirow{12}[0]{*}{ETTh} 

& Context-FID & 64 & 
\textbf{0.522±.031} & 0.631±.058 & 1.130±.102 & 0.827±.146 & 1.543±.153 & 1.279±.083 & 3.008±.277 &  \\
& Score & 128 & 
\textbf{0.633±.029} & 0.787±.062 & 1.553±.169 & 1.062±.134 & 2.354±.170 & 2.554±.318 & 2.639±.427 & {18.2} \\
& (Lower the Better) & 256 & 
\textbf{0.347±.010} & 0.423±.038 & 5.872±.208 & 0.826±.093 & 2.899±.289 & 3.524±.830 & 4.075±.894 &  \\

  \cline{2-11}

  & Correlational & 64 & 
  \textbf{0.070±.002} & 0.082±.005 & 0.483±.019 & 0.067±.006 & 0.186±.008 & 0.094±.010 & 0.271±.007 &  \\
  & Score & 128 & 
  \textbf{0.072±.002} & 0.088±.005 & 0.188±.006 & 0.054±.007 & 0.203±.006 & 0.113±.012 & 0.176±.006 & {16.2} \\
  & (Lower the Better) & 256 & 
  \textbf{0.054±.003} & 0.064±.007 & 0.522±.013 & 0.046±.007 & 0.199±.003 & 0.135±.006 & 0.222±.010 &  \\
  
  \cline{2-11}
  
  & Discriminative & 64 & 
  \textbf{0.087±.017} & 0.106±.048 & 0.227±.078 & 0.171±.142 & 0.254±.074 & 0.150±.003 & 0.296±.348 &  \\
  & Score & 128 & 
  \textbf{0.120±.023} & 0.144±.060 & 0.188±.074 & 0.154±.087 & 0.274±.047 & 0.176±.015 & 0.451±.080 & {18.5} \\
  & (Lower the Better) & 256 & 
  \textbf{0.048±.011} & 0.060±.030 & 0.442±.056 & 0.178±.076 & 0.304±.068 & 0.243±.005 & 0.461±.010 &  \\
  
  \cline{2-11}

  & Predictive & 64 & 
  \textbf{0.098±.003} & 0.116±.000 & 0.132±.008 & 0.118±.004 & 0.133±.008 & 0.118±.004 & 0.135±.003 &  \\
  & Score & 128 & 
  \textbf{0.087±.003} & 0.110±.003 & 0.153±.014 & 0.113±.005 & 0.129±.003 & 0.120±.008 & 0.126±.001 & {17.8} \\
  & (Lower the Better) & 256 & 
  \textbf{0.090±.006} & 0.109±.013 & 0.220±.008 & 0.110±.027 & 0.132±.001 & 0.118±.003 & 0.129±.000 &  \\

  \hline
  \hline
  \multirow{12}[0]{*}{Energy} 
  
  & Context-FID & 64 & 
  0.136±.014 & \textbf{0.135±.017} & 1.230±.070 & 2.662±.087 & 2.697±.418 & 0.762±.157 & 1.824±.144 &  \\
  & Score & 128 & 
  \textbf{0.084±.015} & 0.087±.019 & 2.535±.372 & 3.125±.106 & 5.552±.528 & 1.344±.131 & 1.822±.271 & 2.2 \\
  & (Lower the Better) & 256 & 
  \textbf{0.122±.019} & 0.126±.024 & 5.032±.831 & 3.768±.998 & 5.572±.584 & 4.735±.729 & 2.533±.467 &  \\
  
  \cline{2-11} 
  
  & Correlational & 64 & 
  \textbf{0.618±.012} & 0.672±.035 & 3.668±.106 & 1.653±.208 & 6.847±.083 & 1.281±.218 & 3.319±.062 &  \\
  & Score & 128 & 
  \textbf{0.426±.031} & 0.451±.079 & 4.790±.116 & 1.820±.329 & 6.663±.112 & 1.376±.201 & 3.713±.055 & 6.4 \\
  & (Lower the Better) & 256 & 
  \textbf{0.341±.039} & 0.361±.092 & 4.487±.214 & 1.279±.114 & 5.690±.102 & 1.800±.138 & 3.739±.089 &  \\
  
  \cline{2-11}
  
  & Discriminative & 64 & 
  \textbf{0.066±.005} & 0.078±.021 & 0.498±.001 & 0.499±.000 & 0.497±.004 & 0.328±.031 & 0.499±.001 &  \\
  & Score & 128 & 
  \textbf{0.127±.038} & 0.143±.075 & 0.499±.001 & 0.499±.000 & 0.499±.001 & 0.396±.024 & 0.499±.001 & 13.2 \\
  & (Lower the Better) & 256 & 
  \textbf{0.252±.047} & 0.290±.123 & 0.499±.000 & 0.499±.000 & 0.499±.000 & 0.437±.095 & 0.498±.004 &  \\
  
  \cline{2-11}
  & Predictive & 64 & 
  \textbf{0.225±.001} & 0.249±.000 & 0.291±.003 & 0.302±.001 & 0.252±.001 & 0.252±.000 & 0.262±.002 &  \\
  & Score & 128 & 
  \textbf{0.221±.001} & 0.247±.001 & 0.303±.002 & 0.318±.000 & 0.252±.000 & 0.251±.000 & 0.269±.002 & 11.0 \\
  & (Lower the Better) & 256 & 
  \textbf{0.214±.001} & 0.245±.001 & 0.351±.004 & 0.353±.003 & 0.251±.000 & 0.251±.000 & 0.275±.004 &  \\

  \hline
  \bottomrule
  \end{tabular}
  \end{small}
  \end{threeparttable}
}
\end{table*}

%% file: tables_brief/imputation_full.tex
\begin{table*}[htbp]
\caption{Comparison of the complete performance with diverse mask ratios on \textbf{full-data imputation} task.}\label{tab:imputation_full}
\vskip 0.05in
\centering
\resizebox{2.0\columnwidth}{!}{
\begin{threeparttable}
\begin{small}
\renewcommand{\multirowsetup}{\centering}
\tabcolsep=0.15cm
\renewcommand\arraystretch{1.3}
\begin{tabular}{cccc|cc|cc|cc|cc|cc|cc|cc|cc|cc}
\toprule
\hline

\multicolumn{2}{c}{\multirow{1}{*}{Method}} & 
\multicolumn{2}{c}{\rotatebox{0}{\scalebox{0.8}{\textbf{\myformer}}}} &
\multicolumn{2}{c}{\rotatebox{0}{\scalebox{0.8}{TimeLLM}}} & 
\multicolumn{2}{c}{\rotatebox{0}{\scalebox{0.8}{GPT4TS}}} &
\multicolumn{2}{c}{\rotatebox{0}{\scalebox{0.8}{TimesNet}}} & 
\multicolumn{2}{c}{\rotatebox{0}{\scalebox{0.8}{LLMTime}}} &
\multicolumn{2}{c}{\rotatebox{0}{\scalebox{0.8}{PatchTST}}} & 
\multicolumn{2}{c}{\rotatebox{0}{\scalebox{0.8}{DLinear}}} & 
\multicolumn{2}{c}{\rotatebox{0}{\scalebox{0.8}{FEDformer}}} & 
\multicolumn{2}{c}{\rotatebox{0}{\scalebox{0.8}{Stationary}}} & 
\multicolumn{2}{c}{\rotatebox{0}{\scalebox{0.8}{Autoformer}}} \\

\cline{3-22}

\multicolumn{2}{c}{{MaskRatio}} & 
MSE & MAE & MSE & MAE & MSE & MAE & MSE & MAE & MSE & MAE & MSE & MAE & MSE & MAE & MSE & MAE & MSE & MAE & MSE & MAE \\

\hline

\multirow{5}{*}{\rotatebox{90}{{ETTm1}}}
& 12.5\% & 0.019 & \textbf{0.077} & \textbf{0.017} & 0.085 & 0.023 & 0.101 & 0.041 & 0.130 & 0.096 & 0.229 & 0.093 & 0.206 & 0.080 & 0.193 & 0.052 & 0.166 & 0.032 & 0.119 & 0.046 & 0.144 \\
& 25\% & 0.022 & \textbf{0.092} & \textbf{0.022} & 0.096 & 0.025 & 0.104 & 0.047 & 0.139 & 0.100 & 0.234 & 0.098 & 0.212 & 0.086 & 0.201 & 0.059 & 0.174 & 0.040 & 0.128 & 0.056 & 0.156 \\
& 37.5\% & \textbf{0.028} & \textbf{0.110} & 0.029 & 0.111 & 0.029 & 0.111 & 0.049 & 0.143 & 0.133 & 0.271 & 0.113 & 0.231 & 0.103 & 0.219 & 0.069 & 0.191 & 0.039 & 0.131 & 0.057 & 0.161 \\
& 50\% & \textbf{0.035} & \textbf{0.117} & 0.040 & 0.128 & 0.036 & 0.124 & 0.055 & 0.151 & 0.186 & 0.323 & 0.134 & 0.255 & 0.132 & 0.248 & 0.089 & 0.218 & 0.047 & 0.145 & 0.067 & 0.174 \\
\cline{2-22}
& Avg. & \textbf{0.026} & \textbf{0.099} & 0.028 & 0.105 & 0.027 & 0.107 & 0.047 & 0.140 & 0.120 & 0.253 & 0.104 & 0.218 & 0.093 & 0.206 & 0.062 & 0.177 & 0.036 & 0.126 & 0.051 & 0.150 \\
\hline

\multirow{5}{*}{\rotatebox{90}{{ETTm2}}}
& 12.5\% & 0.018 & 0.079 & \textbf{0.017} & \textbf{0.076} & 0.018 & 0.080 & 0.026 & 0.094 & 0.108 & 0.239 & 0.034 & 0.127 & 0.062 & 0.166 & 0.056 & 0.159 & 0.021 & 0.088 & 0.023 & 0.092 \\
& 25\% & \textbf{0.019} & 0.082 & 0.020 & \textbf{0.080} & 0.020 & 0.085 & 0.028 & 0.099 & 0.164 & 0.294 & 0.042 & 0.143 & 0.085 & 0.196 & 0.080 & 0.195 & 0.024 & 0.096 & 0.026 & 0.101 \\
& 37.5\% & \textbf{0.021} & \textbf{0.085} & 0.022 & 0.087 & 0.023 & 0.091 & 0.030 & 0.104 & 0.237 & 0.356 & 0.051 & 0.159 & 0.106 & 0.222 & 0.110 & 0.231 & 0.027 & 0.103 & 0.030 & 0.108 \\
& 50\% & \textbf{0.024} & \textbf{0.094} & 0.025 & 0.095 & 0.026 & 0.098 & 0.034 & 0.110 & 0.323 & 0.421 & 0.059 & 0.174 & 0.131 & 0.247 & 0.156 & 0.276 & 0.030 & 0.108 & 0.035 & 0.119 \\
\cline{2-22}
& Avg. & \textbf{0.020} & 0.085 & 0.021 & \textbf{0.084} & 0.022 & 0.088 & 0.029 & 0.102 & 0.208 & 0.327 & 0.046 & 0.151 & 0.096 & 0.208 & 0.101 & 0.215 & 0.026 & 0.099 & 0.029 & 0.105 \\
\hline

\multirow{5}{*}{\rotatebox{90}{{ETTh1}}}
& 12.5\% & \textbf{0.040} & \textbf{0.137} & 0.043 & 0.140 & 0.057 & 0.159 & 0.093 & 0.201 & 0.126 & 0.263 & 0.240 & 0.345 & 0.151 & 0.267 & 0.070 & 0.190 & 0.060 & 0.165 & 0.074 & 0.182 \\
& 25\% & \textbf{0.053} & \textbf{0.155} & 0.054 & 0.156 & 0.069 & 0.178 & 0.107 & 0.217 & 0.169 & 0.304 & 0.265 & 0.364 & 0.180 & 0.292 & 0.106 & 0.236 & 0.080 & 0.189 & 0.090 & 0.203 \\
& 37.5\% & \textbf{0.070} & \textbf{0.175} & 0.072 & 0.180 & 0.084 & 0.196 & 0.120 & 0.230 & 0.220 & 0.347 & 0.296 & 0.382 & 0.215 & 0.318 & 0.124 & 0.258 & 0.102 & 0.212 & 0.109 & 0.222 \\
& 50\% & \textbf{0.093} & \textbf{0.202} & 0.107 & 0.216 & 0.102 & 0.215 & 0.141 & 0.248 & 0.293 & 0.402 & 0.334 & 0.404 & 0.257 & 0.347 & 0.165 & 0.299 & 0.133 & 0.240 & 0.137 & 0.248 \\
\cline{2-22}
& Avg. & \textbf{0.064} & \textbf{0.167} & 0.069 & 0.173 & 0.078 & 0.187 & 0.115 & 0.224 & 0.202 & 0.329 & 0.284 & 0.373 & 0.201 & 0.306 & 0.117 & 0.246 & 0.094 & 0.201 & 0.103 & 0.214 \\
\hline

\multirow{5}{*}{\rotatebox{90}{{ETTh2}}}
& 12.5\% & 0.040 & \textbf{0.124} & \textbf{0.039} & 0.125 & 0.040 & 0.130 & 0.057 & 0.152 & 0.187 & 0.319 & 0.101 & 0.231 & 0.100 & 0.216 & 0.095 & 0.212 & 0.042 & 0.133 & 0.044 & 0.138 \\
& 25\% & \textbf{0.043} & \textbf{0.131} & 0.044 & 0.135 & 0.046 & 0.141 & 0.061 & 0.158 & 0.279 & 0.390 & 0.115 & 0.246 & 0.127 & 0.247 & 0.137 & 0.258 & 0.049 & 0.147 & 0.050 & 0.149 \\
& 37.5\% & \textbf{0.049} & \textbf{0.143} & 0.051 & 0.147 & 0.052 & 0.151 & 0.067 & 0.166 & 0.400 & 0.465 & 0.126 & 0.257 & 0.158 & 0.276 & 0.187 & 0.304 & 0.056 & 0.158 & 0.060 & 0.163 \\
& 50\% & \textbf{0.053} & \textbf{0.155} & 0.059 & 0.158 & 0.060 & 0.162 & 0.073 & 0.174 & 0.602 & 0.572 & 0.136 & 0.268 & 0.183 & 0.299 & 0.232 & 0.341 & 0.065 & 0.170 & 0.068 & 0.173 \\
\cline{2-22}
& Avg. & \textbf{0.047} & \textbf{0.138} & 0.048 & 0.141 & 0.049 & 0.146 & 0.065 & 0.163 & 0.367 & 0.436 & 0.119 & 0.250 & 0.142 & 0.259 & 0.163 & 0.279 & 0.053 & 0.152 & 0.055 & 0.156 \\
\hline

\multirow{5}{*}{\rotatebox{90}{{Electricity}}}
& 12.5\% & \textbf{0.043} & \textbf{0.129} & 0.080 & 0.194 & 0.085 & 0.202 & 0.055 & 0.160 & 0.196 & 0.321 & 0.102 & 0.229 & 0.092 & 0.214 & 0.107 & 0.237 & 0.093 & 0.210 & 0.089 & 0.210 \\
& 25\% & \textbf{0.049} & \textbf{0.142} & 0.087 & 0.203 & 0.089 & 0.206 & 0.065 & 0.175 & 0.207 & 0.332 & 0.121 & 0.252 & 0.118 & 0.247 & 0.120 & 0.251 & 0.097 & 0.214 & 0.096 & 0.220 \\
& 37.5\% & \textbf{0.056} & \textbf{0.151} & 0.094 & 0.211 & 0.094 & 0.213 & 0.076 & 0.189 & 0.219 & 0.344 & 0.141 & 0.273 & 0.144 & 0.276 & 0.136 & 0.266 & 0.102 & 0.220 & 0.104 & 0.229 \\
& 50\% & \textbf{0.065} & \textbf{0.165} & 0.101 & 0.220 & 0.100 & 0.221 & 0.091 & 0.208 & 0.235 & 0.357 & 0.160 & 0.293 & 0.175 & 0.305 & 0.158 & 0.284 & 0.108 & 0.228 & 0.113 & 0.239 \\
\cline{2-22}
& Avg. & \textbf{0.053} & \textbf{0.147} & 0.090 & 0.207 & 0.092 & 0.210 & 0.072 & 0.183 & 0.214 & 0.339 & 0.131 & 0.262 & 0.132 & 0.260 & 0.130 & 0.259 & 0.100 & 0.218 & 0.101 & 0.225 \\
\hline

\multirow{5}{*}{\rotatebox{90}{{Weather}}}
& 12.5\% & \textbf{0.024} & \textbf{0.040} & 0.026 & 0.049 & 0.025 & 0.045 & 0.029 & 0.049 & 0.057 & 0.141 & 0.047 & 0.101 & 0.039 & 0.084 & 0.041 & 0.107 & 0.027 & 0.051 & 0.026 & 0.047 \\
& 25\% & \textbf{0.026} & \textbf{0.043} & 0.028 & 0.052 & 0.029 & 0.052 & 0.031 & 0.053 & 0.065 & 0.155 & 0.052 & 0.111 & 0.048 & 0.103 & 0.064 & 0.163 & 0.029 & 0.056 & 0.030 & 0.054 \\
& 37.5\% & \textbf{0.030} & \textbf{0.047} & 0.033 & 0.060 & 0.031 & 0.057 & 0.035 & 0.058 & 0.081 & 0.180 & 0.058 & 0.121 & 0.057 & 0.117 & 0.107 & 0.229 & 0.033 & 0.062 & 0.032 & 0.060 \\
& 50\% & \textbf{0.033} & \textbf{0.052} & 0.037 & 0.065 & 0.034 & 0.062 & 0.038 & 0.063 & 0.102 & 0.207 & 0.065 & 0.133 & 0.066 & 0.134 & 0.183 & 0.312 & 0.037 & 0.068 & 0.037 & 0.067 \\
\cline{2-22}
& Avg. & \textbf{0.028} & \textbf{0.046} & 0.031 & 0.056 & 0.030 & 0.054 & 0.060 & 0.144 & 0.076 & 0.171 & 0.055 & 0.117 & 0.052 & 0.110 & 0.099 & 0.203 & 0.032 & 0.059 & 0.031 & 0.057 \\
\hline

\multicolumn{2}{c}{{$1^{\text{st}}$ Count}} & 
\multicolumn{2}{c}{\textbf{53}} & 
\multicolumn{2}{c}{7} & 
\multicolumn{2}{c}{0} & 
\multicolumn{2}{c}{0} & 
\multicolumn{2}{c}{0} & 
\multicolumn{2}{c}{0} & 
\multicolumn{2}{c}{0} & 
\multicolumn{2}{c}{0} & 
\multicolumn{2}{c}{0} & 
\multicolumn{2}{c}{0} \\
    
\hline
\bottomrule
\end{tabular}
\end{small}
\end{threeparttable}
}
\end{table*}